\documentclass[journal]{IEEEtran}

\makeatletter
\let\NAT@parse\undefined
\makeatother
\usepackage{cite}
\usepackage{amsmath,amssymb,amsfonts}
\usepackage{amsthm}

\newtheorem{theorem}{Theorem}
\newtheorem*{theoremS1}{Theorem 1}
\usepackage{algorithmic}
\usepackage{algorithm}
\usepackage{graphicx}
\usepackage{textcomp}
\usepackage[table]{xcolor}
\usepackage{array}
\newcolumntype{C}[1]{>{\centering\arraybackslash}p{#1}}
\usepackage{booktabs}
\usepackage{multirow}
\usepackage{adjustbox}  
\usepackage{url}
\usepackage{float}
\usepackage{placeins}
\usepackage[T1]{fontenc}
\usepackage[bookmarksnumbered=true,colorlinks=true,allcolors=blue]{hyperref}
\begin{document}
\bstctlcite{IEEEexample:BSTcontrol}

\title{IKOD: Image-Attention-Guided Key-Value Merging against Object Hallucination in the Decoding of Large Vision-Language Models}

\author{Jiabing~Yang$^{\ast}$,
        Chenhang~Cui$^{\ast}$,
        Yixiang~Chen,
        Tao~Yu,
        Yan~Huang$^{\dagger}$,
        and~Liang~Wang,~\IEEEmembership{Fellow,~IEEE}%
\thanks{$^{\ast}$Equal contribution.}%
\thanks{$^{\dagger}$Corresponding author.}%
\thanks{Jiabing Yang, Yixiang Chen, Tao Yu, Yan Huang, and Liang Wang are with the School of Artificial Intelligence, University of Chinese Academy of Sciences (UCAS), Beijing 100864, China, and also with the New Laboratory of Pattern Recognition (NLPR), Institute of Automation, Chinese Academy of Sciences (CASIA), Beijing 100190, China (e-mail: yangjiabing2025@ia.ac.cn; yixiang.chen@cripac.ia.ac.cn; yutao2025@ia.ac.cn; yhuang@nlpr.ia.ac.cn; wangliang@nlpr.ia.ac.cn).}%
\thanks{Chenhang Cui is with the School of Computing, National University of Singapore, Singapore 117417 (e-mail: chenhangcui@gmail.com).}}

\markboth{IEEE Transactions on Image Processing}%
{Yang \MakeLowercase{\textit{et al.}}: Image-Attention-Guided Key-Value Merging against Object Hallucination}

\maketitle


\begin{abstract}
Large vision-language models (LVLMs) exhibit strong visual understanding and language generation capabilities yet remain prone to object hallucination, producing plausible descriptions unsupported by the input image. Analyzing generation dynamics across eight diverse LVLMs, we observe that cross-modal visual attention steadily decays as generation unfolds because appended text tokens continuously expand the softmax denominator; consequently, hallucinated objects consistently appear later and under lower image attention than grounded mentions. Unlike existing training-free remedies that rely on input perturbations or external models, we propose IKOD (Image-attention-guided Key-value merging against Object hallucination in Decoding), a collaborative decoder that directly compresses the competing text context within the native key-value (KV) cache. At each step, IKOD keeps visual tokens intact and merges low-attention text entries into a compact auxiliary cache, fusing the resulting image-focused distribution with the base branch to correct ungrounded predictions while preserving linguistic fluency. An adaptive variant (IKOD-A) further schedules collaboration strength online from the measured attention decay under a single universal configuration. Extensive experiments across hallucination and general multimodal benchmarks on eight diverse LVLMs demonstrate that IKOD and IKOD-A consistently mitigate object hallucination while preserving object recall and general multimodal capabilities at minimal inference cost. Code is available at \url{https://github.com/jiabingyang01/IKOD}.
\end{abstract}

\begin{IEEEkeywords}
Large vision-language models, object hallucination, training-free decoding, visual attention, key-value cache.
\end{IEEEkeywords}

\section{Introduction}
\label{sec:intro}
\IEEEPARstart{L}{arge} vision-language models (LVLMs) extend autoregressive language models with visual perception and now support a wide range of multimodal tasks~\cite{LLaVA-1.5,llavanext2024,llavaonevision2024,qwen2vl2024,qwen25vl2025,qwen3vl2025,internvl3_2025,InstructBLIP}. Despite this progress, they frequently suffer from object hallucination, describing nonexistent objects unsupported by the input image~\cite{li2023evaluating,rohrbach2018object}. Because such errors undermine reliability in high-stakes domains including medical diagnosis~\cite{chen2024detecting}, finance~\cite{kang2023deficiency}, autonomous driving~\cite{Xiong_2026_CVPR}, and robotics~\cite{ding2024open6dor}, reducing hallucination is essential for trustworthy deployment~\cite{vcd,huang2024opera,agla}.

Existing remedies fall into three main categories. Instruction and preference tuning require costly annotated data and retraining~\cite{liu2023aligning,zhao2023svit,rlhfv}, while post-hoc correction relies on external detectors or verification models~\cite{zhou2023analyzing,yin2024woodpecker}. Training-free decoding methods adjust the output distribution directly at inference time~\cite{zhang2024debiasing}, yet most rely on perturbed inputs or external views~\cite{vcd,agla}, attention-pattern calibration~\cite{huang2024opera}, or internal representation intervention~\cite{revisit2025}. This raises a natural question: \textit{Can we restore visual grounding at inference time simply by reorganizing the model's own context, without extra training, external models, or input perturbations?}

Our approach starts from a consistent observation~\cite{favero2024multi,vista2025,tarac2025}: as an LVLM generates a response, the attention it allocates to image tokens steadily declines, as documented across eight diverse LVLMs in Fig.~\ref{fig:fading}. Moreover, as shown in Section~\ref{sec:analysis}, hallucinated object mentions consistently fall later and under lower image attention than grounded mentions in the same caption. This decline follows directly from softmax normalization in self-attention: for a fixed query, each newly appended text token adds a positive exponential term to the softmax denominator, gradually diluting the share of attention placed on visual tokens. Because the growing text context competes with the visual prefix for attention, visual grounding can be restored by directly compressing this competing text context within the native KV cache.

Guided by this finding, we propose \textbf{I}mage-attention-guided \textbf{K}ey-value merging against \textbf{O}bject hallucination in \textbf{D}ecoding (IKOD), a training-free collaborative decoding method. At each decoding step, IKOD builds a compact, image-focused auxiliary branch from the model's own KV cache by keeping the visual prefix intact and merging the text entries that attend least to the image. A single forward pass over this merged cache produces an image-focused distribution that collaborates with the base branch to improve visual grounding while preserving linguistic fluency. Beyond a static fusion weight, we also introduce an adaptive variant (IKOD-A) that schedules the collaboration strength online from the measured attention decay. Across eight 7B--8B LVLMs, both variants consistently suppress hallucination on POPE~\cite{li2023evaluating}, CHAIR~\cite{rohrbach2018object}, and AMBER~\cite{wang2023amber} at minimal decoding overhead without compromising object recall or general multimodal capabilities, with gains transferring across diverse decoding strategies, prompt formulations, and 13B--32B scales.

Our main contributions are threefold:
\begin{itemize}
\item We show across eight diverse LVLMs that cross-modal visual attention steadily decays over generation, and reveal that object hallucinations consistently appear later and under lower image attention than grounded mentions, motivating decoding-time context compression.
\item We propose IKOD, a training-free collaborative decoder that merges low-attention text KV entries into an image-focused auxiliary branch to provably restore visual attention, fusing its distribution with the base branch without input perturbations or external models.
\item We design a decay-governed adaptive variant (IKOD-A) that operates under a single universal configuration, and demonstrate through extensive experiments that both variants consistently mitigate hallucination while preserving object recall and general capabilities at minimal overhead.
\end{itemize}

\section{Related Work}
\label{sec:related}

\noindent\textbf{Visual grounding in long-form generation.}
As LVLMs generate longer responses, model predictions often drift toward language priors over visual evidence and become more prone to hallucination~\cite{favero2024multi,zhou2023analyzing,vista2025}. Existing methods counter this drift from different angles, including amplifying visual signals or steering hidden states~\cite{favero2024multi,vista2025,rudder2026}, accumulating historical attention across decoding steps~\cite{tarac2025,afip2026}, summarizing earlier text context~\cite{sumgd2025}, or rolling back generation upon heuristic triggers~\cite{halback2026,vgd2026}, despite the high variance of per-step attention as a token-level signal~\cite{saliencydrops2026}. In contrast, IKOD counteracts softmax attention dilution at its source by compressing low-image-attention text entries into an auxiliary KV cache, restoring the visual attention share without modifying base activations or rolling back generation.

\noindent\textbf{Training-free decoding.}
Training-free decoders adjust next-token logits at inference time without parameter updates. A major family combines the base prediction with a contrastive or cooperative auxiliary distribution~\cite{vcd,icd2024,agla} derived from perturbed or omitted visual inputs~\cite{vcd,favero2024multi,sid2025}, modified prompts or self-descriptions~\cite{icd2024,code2024}, augmented image views~\cite{agla}, or intermediate layer representations~\cite{chuang2023dola,revisit2025}. Other methods calibrate attention via beam-search penalties~\cite{huang2024opera} or attention rescaling~\cite{pai2024}, or steer internal hidden states~\cite{vista2025,rudder2026}. Among dual-branch methods, AGLA~\cite{agla} similarly fuses a positive auxiliary distribution over plausible candidates, but relies on an external augmented view; in contrast, IKOD forms its image-focused branch internally by reorganizing the native KV cache, provably increasing the visual attention share for any fixed query.

\noindent\textbf{Key-value cache compression and modification.}
KV cache pruning, merging, and smoothing are widely used for long-context acceleration~\cite{cam2024,meda2025,liu2024efficient} and have recently been adapted to mitigate LVLM hallucination~\cite{vasparse2025,kvsmooth2026,lime2026}. Specifically, CaM~\cite{cam2024}, MEDA~\cite{meda2025}, and Elastic Cache~\cite{liu2024efficient} merge redundant entries around anchor tokens to save memory, whereas VASparse~\cite{vasparse2025} retains visual-aware tokens, KVSmooth~\cite{kvsmooth2026} smooths adjacent keys and values, and LIME~\cite{lime2026} optimizes additive cache perturbations via backpropagation. Unlike methods that permanently alter the base cache or require gradients, IKOD keeps the base cache and visual prefix intact, applying closed-form bucket averaging to low-attention text tokens to form a transient image-focused branch for collaborative decoding.

\section{Observations on Visual Attention}
\label{sec:analysis}
\begin{figure}[!t]
    \centering
    \includegraphics[width=\linewidth]{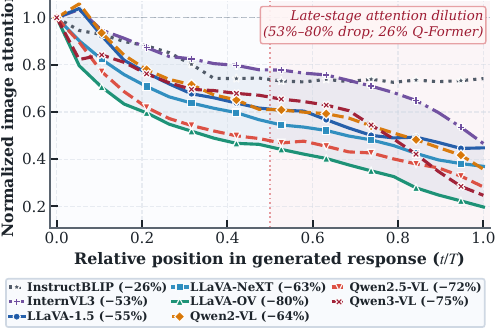}
    \caption{Visual attention trajectories. Normalized visual attention share against relative response position across eight LVLMs, averaged over 500 MSCOCO validation images. Each curve is scaled to its first bin, with the first-to-last-bin change shown in the legend.}
    \label{fig:fading}
\end{figure}
While visual attention decay over generation has been noted on individual models~\cite{favero2024multi,itad2025,vista2025,tarac2025,mcothall2026}, we first verify whether this pattern holds across modern architectures by analyzing eight open-source LVLMs from the LLaVA, Qwen-VL, InternVL, and InstructBLIP families.

\subsection{Measuring Visual Attention}
At decoding step $t$, let $S_t^{i,h} = \sum_{j \in \mathcal{I}} \mathrm{Att}_{t,j}^{i,h}$ denote the attention share allocated to visual tokens $\mathcal{I}$ by query head $h \in \mathcal{H}_q$ in layer $i \in \{1,\dots,I\}$. Averaging over all layers and heads gives the step-$t$ image attention:
\begin{equation}
S_t = \frac{1}{I |\mathcal{H}_q|} \sum_{i=1}^I \sum_{h \in \mathcal{H}_q} S_t^{i,h} = \frac{1}{I |\mathcal{H}_q|} \sum_{i=1}^I \sum_{h \in \mathcal{H}_q} \sum_{j \in \mathcal{I}} \mathrm{Att}_{t,j}^{i,h}.
\label{eq:att_traj}
\end{equation}
This defines an attention trajectory $\boldsymbol{S} = (S_1, \dots, S_T)$ over sequence length $T$. We collect trajectories under greedy decoding (512-token cap) on the 500 CHAIR MSCOCO images~\cite{MSCOCO}, partition each trajectory into 20 uniform relative-position bins, and label object mentions as grounded or hallucinated via CHAIR~\cite{rohrbach2018object}.

\subsection{Visual Attention Degradation across Architectures}
\begin{figure}[!t]
    \centering
    \includegraphics[width=\linewidth]{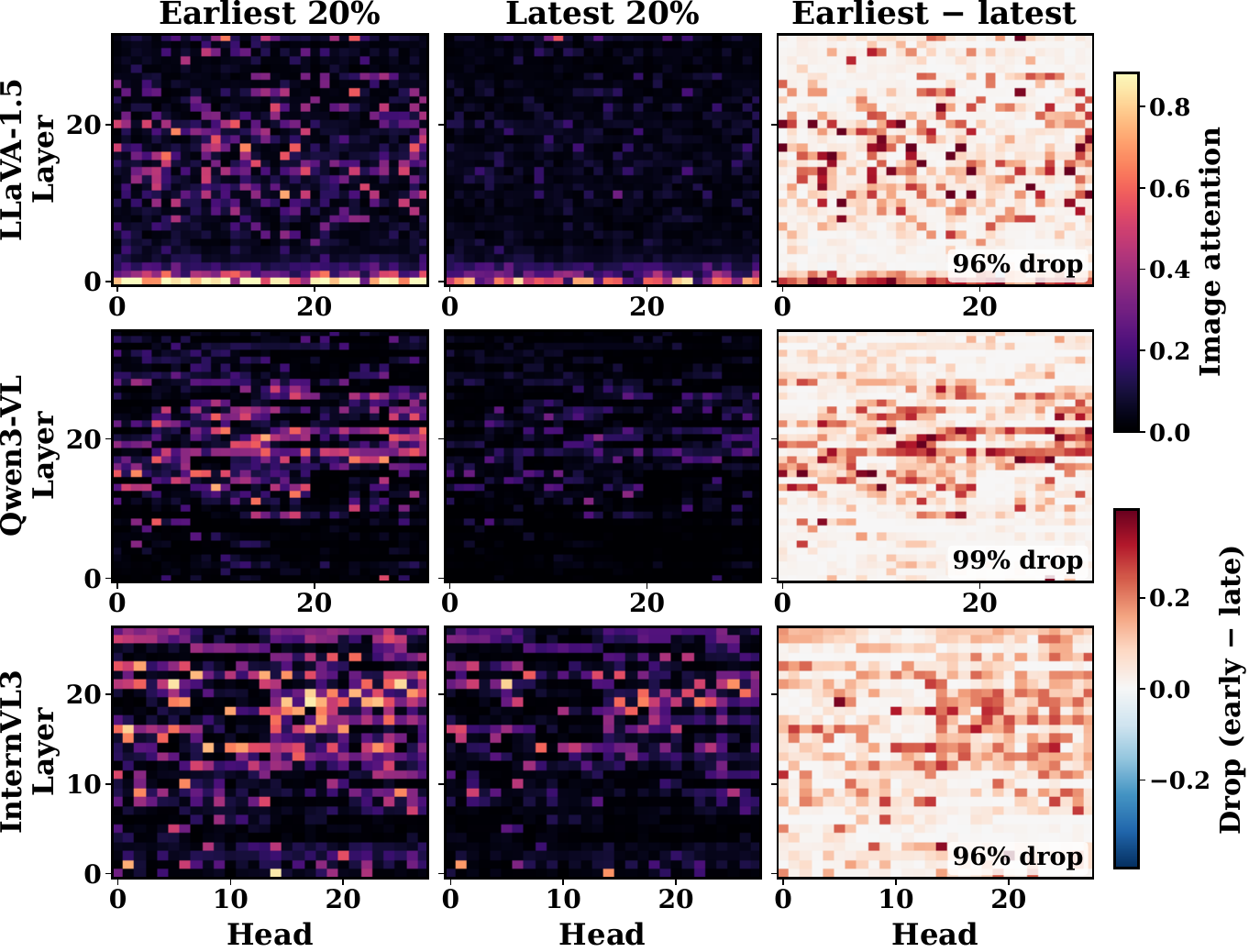}
    \caption{Layer- and head-wise visual attention. Visual attention share for the first 20\% (left) and last 20\% (middle) of generated tokens across three representative backbones, alongside their difference $S_{\mathrm{last}}^{i,h} - S_{\mathrm{first}}^{i,h}$ (right, where red marks attention decay).}
    \label{fig:layerhead}
\end{figure}

\begin{figure}[!t]
    \centering
    \includegraphics[width=\linewidth]{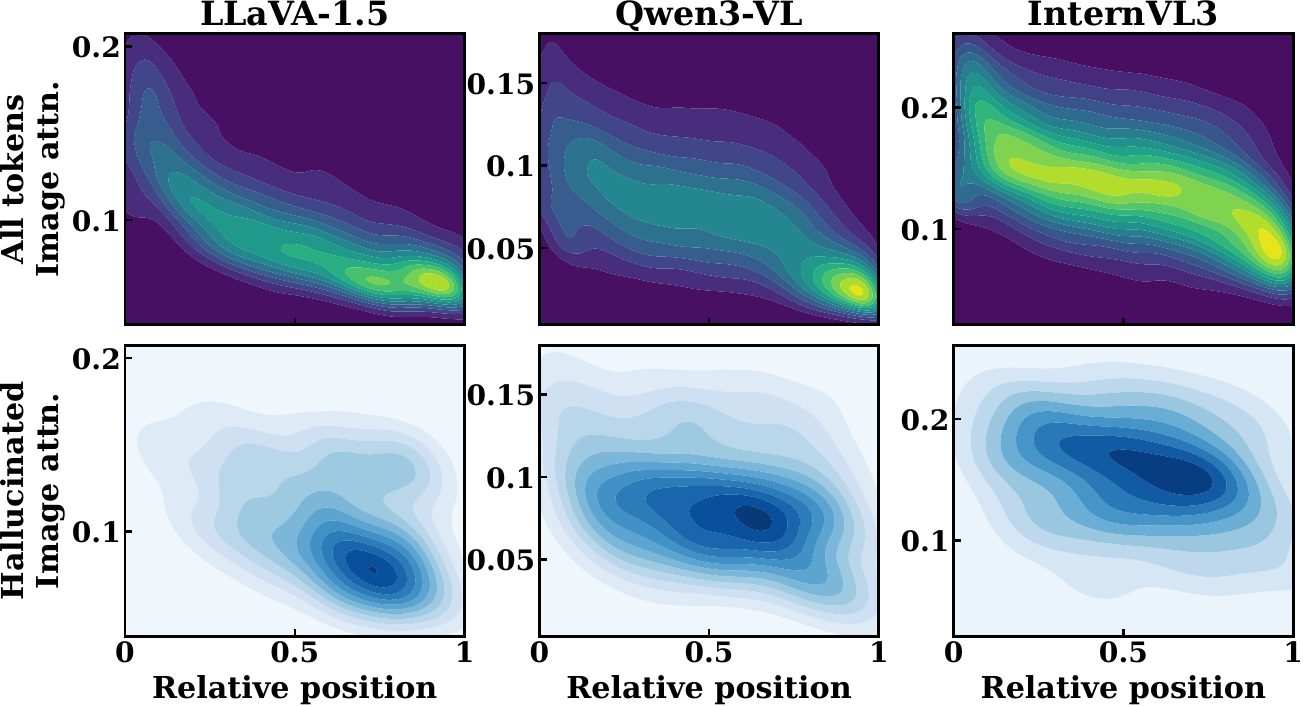}
    \caption{Joint position--attention density of generated and hallucinated tokens. Kernel density over relative token position and visual attention share for all generated tokens (top) and hallucinated object mentions (bottom) across three representative backbones on 500 MSCOCO captions.}
    \label{fig:density}
\end{figure}

In Fig.~\ref{fig:fading}, layer-and-head averaged image attention $S_t$ declines smoothly across generation on all eight backbones: from the first to the last bin, the relative drop spans 53\% on InternVL3 to 80\% on LLaVA-OneVision, and 26\% on Q-Former-based InstructBLIP (32 query tokens). To check whether this decay is driven by a few heads or holds network-wide, Fig.~\ref{fig:layerhead} inspects the unaveraged per-layer, per-head attention $S_t^{i,h}$ on LLaVA-1.5, Qwen3-VL, and InternVL3 (remaining five models in the Supplementary Material): whereas middle and deep heads attend strongly to the image in the first 20\% of tokens (left), attention drops across 91.1\%--99.5\% of layer-head pairs by the last 20\% (middle and right, red).

\subsection{Association with Hallucinated Mentions}
Prior work notes that object hallucinations occur more frequently later in generation~\cite{zhou2023analyzing,favero2024multi,pas2026}. Fig.~\ref{fig:density} examines this relationship at the token level by comparing the joint position--attention density of all generated tokens (top row) against CHAIR-hallucinated object mentions (bottom row) across three representative backbones on 500 MSCOCO captions (with all eight backbones reported in the Supplementary Material). While early tokens span higher image attention before settling into a narrow low-attention band, hallucinated mentions shift markedly toward middle-to-late sequence positions and cluster tightly within this low-attention band: across all eight backbones, hallucinated object mentions appear 4.8 to 21.8 percentage points later than grounded mentions within the same caption, accompanied by 0.1\% to 9.7\% lower mean image attention. Progressive attention dilution by accumulating post-visual text tokens thus directly weakens visual grounding over long generation, motivating our decoding-time compression of low-attention text context to restore image focus.

\section{Method}
\label{sec:method}
\begin{figure*}[!t]
    \centering
    \includegraphics[width=\textwidth]{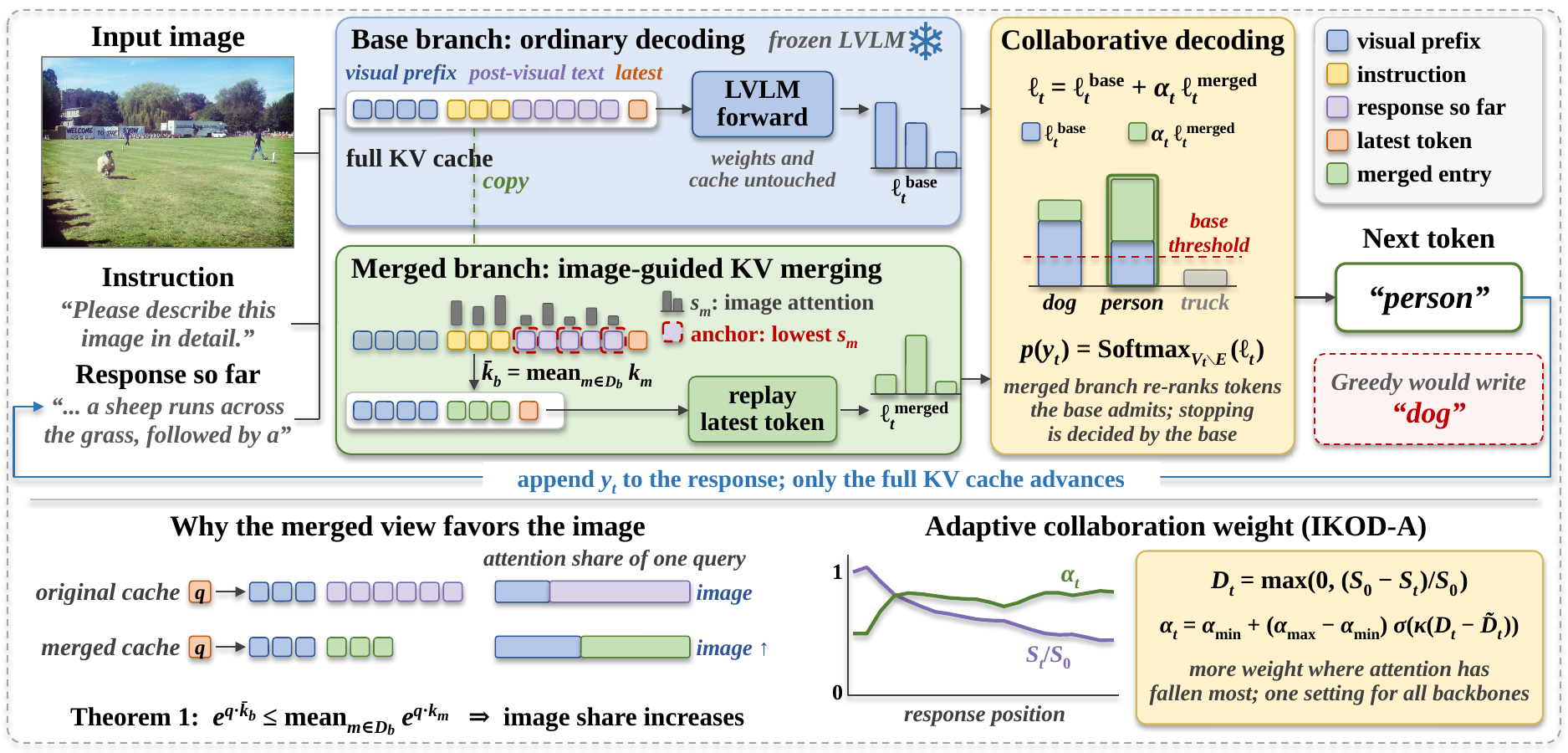}
    \caption{Overview of the proposed IKOD and IKOD-A framework. (a) At each decoding step $t$, the base branch evaluates over the full KV cache $\mathcal{C}_t$ while the merged branch selects the lowest-$s^i_m$ post-visual text tokens as anchors and averages each contiguous bucket into a centroid $(\tilde{\boldsymbol{k}}^i_b,\tilde{\boldsymbol{v}}^i_b)$ for additive collaborative decoding. (b) Bucket centroid merging provably increases visual attention share ($S' > S$, Theorem~\ref{thm:share}). (c) IKOD-A schedules the collaboration weight $\alpha_t$ online from real-time visual attention decay $D_t$.}
    \label{fig:pipeline}
\end{figure*}

To counter visual attention dilution from accumulating text tokens, we propose IKOD to restore visual grounding at decoding time. As shown in Fig.~\ref{fig:pipeline}, IKOD operates in two stages at each step: (1) \emph{image-attention-guided KV merging}, which compresses low-attention text entries to form a compact, image-focused branch alongside the base branch; and (2) \emph{collaborative decoding}, which additively fuses both branches' logits under a plausibility constraint. Beyond a fixed fusion weight (IKOD), our adaptive variant \emph{IKOD-A} schedules collaboration strength online from the measured attention decay under a single universal configuration.

\subsection{Image-Attention-Guided Key-Value Merging}
\label{sec:kv}
Consider a multimodal transformer with layers $i=1,\dots,I$, query heads $\mathcal{H}_q$, and key-value (KV) heads $\mathcal{H}_{kv}$, supporting both multi-head and grouped-query attention. At step $t$, the layer-$i$ KV cache holds the visual prefix (system prompt and visual tokens $\mathcal{I}$) followed by the post-visual text tokens (post-image instruction tokens and tokens generated up to step $t-1$). For each cached post-visual text position $m$, we measure its layer-$i$ image attention by summing its attention over $\mathcal{I}$ and averaging across query heads:
\begin{equation}
s^i_m=\frac{1}{|\mathcal{H}_q|}\sum_{h\in\mathcal{H}_q}\sum_{j\in\mathcal{I}}\mathrm{Att}^{i,h}_{m,j},
\label{eq:score}
\end{equation}
where $\mathrm{Att}^{i,h}_{m,j}$ is the attention weight that query head $h$ of layer $i$ places on visual token $j$ when position $m$ is evaluated. Because $s^i_m$ is already computed when position $m$ is processed during prefill or prior decoding steps, recording these scores adds no extra attention overhead.

\noindent\textbf{Mergeable text region.}
At step $t$, the candidate set of mergeable text positions $\mathcal{P}_t$ consists of all cached post-visual text positions prior to the current query token $t$. Including both post-visual instruction tokens and previously generated tokens in $\mathcal{P}_t$ ensures that the operator applies uniformly to short-answer tasks (such as POPE, where no response tokens exist at the first step) and long-form captioning.

\noindent\textbf{Anchor selection and bucket merging.}
As generation proceeds, an increasing share of text tokens attend weakly to the image and expand the softmax denominator. IKOD selects these low-attention tokens as anchors to partition and compress the text cache while keeping query token $t$ unmerged for local syntactic continuity. Given anchor ratio $\lambda \in (0, 1)$, layer $i$ selects the $n_t = \operatorname{round}(\lambda|\mathcal{P}_t|)$ lowest-$s^i_m$ positions in $\mathcal{P}_t$ as sorted anchors $\mathcal{A}^i = \{a^i_1 < \dots < a^i_{n_t}\}$ (leaving the cache unmerged if $n_t = 0$), and assigns each $m \in \mathcal{P}_t$ to its nearest anchor (breaking ties leftward to partition $\mathcal{P}_t$ at midpoints $\lfloor(a^i_b + a^i_{b+1})/2\rfloor$ into $n_t$ contiguous intervals):
\begin{equation}
D^i_b = \big\{m \in \mathcal{P}_t:\ \arg\min\nolimits_{a \in \mathcal{A}^i}|m - a| = a^i_b\big\}.
\label{eq:bucket}
\end{equation}
Letting $\boldsymbol{k}^{i,g}_m$ and $\boldsymbol{v}^{i,g}_m$ denote the key and value vectors at position $m$ in KV head $g \in \mathcal{H}_{kv}$ of layer $i$, we average the keys and values within each bucket $D^i_b$:
\begin{equation}
\bar{\boldsymbol{k}}^{i,g}_{b} = \frac{1}{|D^i_b|}\sum_{m \in D^i_b}\boldsymbol{k}^{i,g}_m,\qquad
\bar{\boldsymbol{v}}^{i,g}_{b} = \frac{1}{|D^i_b|}\sum_{m \in D^i_b}\boldsymbol{v}^{i,g}_m.
\label{eq:merge}
\end{equation}
Concatenating the intact visual prefix, the merged centroids $\{\bar{\boldsymbol{k}}^{i,g}_b, \bar{\boldsymbol{v}}^{i,g}_b\}_{b=1}^{n_t}$, and the current token forms the merged KV cache. Unlike token pruning, which drops tokens and hurts recall, bucket averaging folds surrounding semantic cues into each centroid while shrinking the competing text sequence from $|\mathcal{P}_t|$ to $n_t$ entries. To see why this restores visual focus, consider a single attention head with query $\boldsymbol{q}\in\mathbb{R}^{d_h}$ and scaled scores $z_m = \boldsymbol{q}^\top\boldsymbol{k}_m/\sqrt{d_h}$: in the full cache, the softmax denominator sums the exponential scores of the visual tokens $A_{\mathcal{I}}=\sum_{j\in\mathcal{I}}e^{z_j}>0$, the mergeable text tokens $A_{\mathcal{P}}=\sum_{m\in\mathcal{P}_t}e^{z_m}$, and the unmerged text tokens $B \ge 0$ (system prefix and token $t$), giving pre-merge visual attention share $S = A_{\mathcal{I}} / (A_{\mathcal{I}} + B + A_{\mathcal{P}})$ and the following guarantee:

\begin{theorem}[Visual attention recovery under bucket merging]
\label{thm:share}
Fix any query $\boldsymbol{q}$ and visual prefix keys $\{\boldsymbol{k}_j\}_{j \in \mathcal{I}}$. Partitioning $\mathcal{P}_t$ into nonempty buckets $\{D_b\}_{b=1}^{n_t}$ and replacing each bucket with its centroid key $\bar{\boldsymbol{k}}_b = \frac{1}{|D_b|}\sum_{m \in D_b}\boldsymbol{k}_m$ yields visual attention share $S' \ge S$, with strict inequality $S' > S$ whenever any $|D_b| \ge 2$. If $\min_b |D_b| \ge b_0$, then $S' \ge A_{\mathcal{I}} / (A_{\mathcal{I}} + B + A_{\mathcal{P}}/b_0)$.
\end{theorem}

By linearity of the inner product, each centroid has score $\bar{z}_b = \boldsymbol{q}^\top\bar{\boldsymbol{k}}_b/\sqrt{d_h} = \frac{1}{|D_b|}\sum_{m\in D_b}z_m$. Applying Jensen's inequality to the convex exponential function gives
\begin{equation}
\exp(\bar{z}_b) \le \frac{1}{|D_b|} \sum_{m \in D_b} \exp(z_m) \le \sum_{m \in D_b} \exp(z_m),
\end{equation}
with strict second inequality whenever $|D_b| \ge 2$ since $\exp(z_m) > 0$. Summing over $b=1,\dots,n_t$ gives $A'_{\mathcal{P}} = \sum_{b=1}^{n_t}\exp(\bar{z}_b) \le \sum_{b=1}^{n_t}\frac{1}{|D_b|}\sum_{m\in D_b}\exp(z_m) \le A_{\mathcal{P}}/b_0$ when $\min_b|D_b|\ge b_0$. Since $A_{\mathcal{I}}$ and $B$ are unchanged, replacing $A_{\mathcal{P}}$ with $A'_{\mathcal{P}}$ yields the lower bound and $S' > S$ whenever any $|D_b| \ge 2$ (proof in the Supplementary Material), combining Jensen convexity with $1/|D_b|$ count compression so smaller $\lambda \approx 1/b_0$ yields stronger attention recovery. Evaluating token $t$ on this merged cache gives logits $\ell^{\mathrm{merged}}_t$ alongside base logits $\ell^{\mathrm{base}}_t$.

\subsection{Collaborative Decoding}
\label{sec:co}
While the merged cache strengthens visual grounding, compressing text history coarsens fine-grained phrasing details needed for fluency. IKOD thus additively fuses the base logits $\ell^{\mathrm{base}}_t$ and merged logits $\ell^{\mathrm{merged}}_t$:
\begin{equation}
\ell_t = \ell^{\mathrm{base}}_t + \alpha\,\ell^{\mathrm{merged}}_t,
\label{eq:collab}
\end{equation}
where $\alpha \ge 0$ sets the collaboration weight. Following prior decoding works~\cite{cd,vcd,agla}, we restrict candidate tokens to the plausible set from the base branch:
\begin{equation}
\mathcal{V}_t = \big\{y \in \mathcal{V}:\ \ell^{\mathrm{base}}_t(y) \ge \log\beta + \max\nolimits_w\ell^{\mathrm{base}}_t(w)\big\},
\label{eq:plaus}
\end{equation}
where $\ell^{\mathrm{base}}_t(y)$ is the logit of token $y \in \mathcal{V}$ and $\beta \in (0,1]$ is a plausibility threshold. Stopping is determined by the base branch: if $\arg\max_w \ell^{\mathrm{base}}_t(w) \in \mathcal{E}$ (for end-of-sequence set $\mathcal{E}$), generation stops; otherwise, tokens outside $\mathcal{V}_t \setminus \mathcal{E}$ are masked out ($p(y)=0$) to form the next-token distribution:
\begin{equation}
p(y_t) = \operatorname{Softmax}_{\mathcal{V}_t \setminus \mathcal{E}}(\ell_t),
\label{eq:select}
\end{equation}
from which $y_t$ is decoded (via $\arg\max$ under greedy decoding or sampled from $p(y_t)$ under stochastic decoding) and appended to the KV cache for the next step.

\subsection{Adaptive Variant: IKOD-A}
\label{sec:adaptive}
\begin{algorithm}[t]
\caption{Single Decoding Step of IKOD and IKOD-A}
\label{alg:ikod}
\small
\begin{algorithmic}[1]
\REQUIRE Full KV cache $\{(\boldsymbol{K}^i,\boldsymbol{V}^i)\}_{i=1}^I$, visual prefix $\mathcal{I}$, cached post-visual text scores $\{s^i_m\}_{m\in\mathcal{P}_t}$, query token $t$, ratio $\lambda$, threshold $\beta$, weight $\alpha$ (IKOD) or $(\alpha_{\min},\alpha_{\max},\kappa,S_0)$ (IKOD-A)
\ENSURE Next token $y_t$ (or EOS) and updated KV cache
\STATE Evaluate token $t$ on the full KV cache to obtain base logits $\ell_t^{\mathrm{base}}$ and current image-attention scores $\{s^i_t\}_{i=1}^I$ via Eq.~\eqref{eq:score}
\IF{$\arg\max_w\ell_t^{\mathrm{base}}(w)\in\mathcal{E}$}
  \RETURN EOS
\ENDIF
\STATE Form plausible candidate set $\mathcal{V}_t$ from $\ell_t^{\mathrm{base}}$ via Eq.~\eqref{eq:plaus}
\STATE Set fixed weight $\alpha_t \leftarrow \alpha$ (IKOD), or compute $S_t = \frac{1}{I}\sum_{i=1}^I s^i_t$, decay $D_t$, and adaptive weight $\alpha_t$ via Eqs.~\eqref{eq:decay}--\eqref{eq:adaptive} (IKOD-A)
\IF{$|\mathcal{V}_t \setminus \mathcal{E}| = 1$ \OR $\alpha_t = 0$}
  \STATE Select $y_t \leftarrow \arg\max_{y \in \mathcal{V}_t \setminus \mathcal{E}} \ell_t^{\mathrm{base}}(y)$
\ELSE
  \FOR{each layer $i=1,\dots,I$ (in parallel across layers)}
    \STATE Select $n_t = \operatorname{round}(\lambda|\mathcal{P}_t|)$ lowest-score positions in $\mathcal{P}_t$ as anchors $\mathcal{A}^i$ and partition $\mathcal{P}_t$ into buckets $\{D^i_b\}_{b=1}^{n_t}$ via Eq.~\eqref{eq:bucket}
    \STATE Concatenate intact visual prefix, bucket means $\{\bar{\boldsymbol{k}}^{i,g}_b,\bar{\boldsymbol{v}}^{i,g}_b\}$ via Eq.~\eqref{eq:merge}, and token $t$ to form the merged KV cache
  \ENDFOR
  \STATE Evaluate token $t$ on the merged KV cache to obtain merged logits $\ell_t^{\mathrm{merged}}$
  \STATE Decode $y_t$ from $p(y_t)$ in Eq.~\eqref{eq:select} with $\ell_t = \ell_t^{\mathrm{base}} + \alpha_t\,\ell_t^{\mathrm{merged}}$
\ENDIF
\STATE Append $y_t$ to the full KV cache and store $\{s^i_t\}_{i=1}^I$ for step $t+1$
\RETURN $y_t$
\end{algorithmic}
\end{algorithm}
Because visual attention decays progressively over generation and varies in scale across architectures, applying a constant $\alpha$ at every step can over-intervene early on. We therefore introduce IKOD-A (summarized with IKOD in Algorithm~\mbox{\ref{alg:ikod}}), which replaces static $\alpha$ in Eqs.~\eqref{eq:collab} and~\eqref{eq:select} with an online weight $\alpha_t$ driven by real-time attention decay, applying minimal intervention early and stronger collaboration as attention drops. Letting $S_t = \frac{1}{I}\sum_{i=1}^I s^i_t$ (Eq.~\eqref{eq:att_traj}) and $\bar{S}_0 = \max(S_0, 10^{-6})$ at the first generated token, we define relative decay $D_t$ and weight $\alpha_t$ as
\begin{equation}
D_t = \max\Big(0,\ \frac{\bar{S}_0 - S_t}{\bar{S}_0}\Big),
\label{eq:decay}
\end{equation}
\begin{equation}
\alpha_t = \alpha_{\min} + (\alpha_{\max} - \alpha_{\min})\,\sigma\big(\kappa\,(D_t - \tilde{D}_t)\big),
\label{eq:adaptive}
\end{equation}
where $[\alpha_{\min}, \alpha_{\max}]$ is the weight range, $\sigma(\cdot)$ is the logistic sigmoid with slope $\kappa > 0$, and $\tilde{D}_t = \operatorname{median}_{k \le t}(D_k)$ is the running median decay. Thus $\alpha_t$ rises toward $\alpha_{\max}$ when $D_t > \tilde{D}_t$ and relaxes toward $\alpha_{\min}$ otherwise. Normalizing $D_t$ by $\bar{S}_0$ makes it scale-invariant across models, enabling a single universal setting $(\alpha_{\min}, \alpha_{\max}, \kappa) = (0, 1, 10)$ across all backbones and benchmarks.


\begin{table*}[!t]
\centering
\caption{POPE F1 scores (\%) on the MSCOCO validation set across eight LVLMs. {\normalfont\itshape Average} denotes the three-split mean F1. Bold and underline mark the best and second-best results in each row. $^{\ast}$: five-beam search.}
\label{tab:pope}
\small
\setlength{\tabcolsep}{1.798pt}
\begin{tabular}{ll cccccccccc}
\toprule
\multirow{2}{*}{Model} & \multirow{2}{*}{Split} & Greedy & VCD & M3ID & OPERA$^{\ast}$ & ICD & SID & AGLA & ReVisiT & \textbf{IKOD} & \textbf{IKOD-A} \\
  &  & \textit{(Baseline)} & \textit{(CVPR'24)} & \textit{(CVPR'24)} & \textit{(CVPR'24)} & \textit{(ACL'24)} & \textit{(ICLR'25)} & \textit{(CVPR'25)} & \textit{(ACL'26)} & \textit{(Ours)} & \textit{(Ours)} \\
\midrule
\multirow{4}{*}{LLaVA-1.5~7B} & \textit{Random} & 86.95 & \textbf{89.06} & 85.10 & 87.21 & 85.53 & 85.83 & 86.99 & 85.50 & \underline{88.90} & 87.70 \\
  & \textit{Popular} & 85.96 & \underline{86.86} & 84.21 & 86.08 & 84.55 & 84.52 & 85.92 & 84.78 & \textbf{87.26} & 86.54 \\
  & \textit{Adversarial} & 84.02 & 83.41 & 82.21 & 84.03 & 83.01 & 82.54 & 84.08 & 82.79 & \underline{84.32} & \textbf{84.35} \\
\cmidrule(l){2-12}
  & \textit{Average} & 85.64 & \underline{86.45} & 83.84 & 85.77 & 84.36 & 84.30 & 85.67 & 84.36 & \textbf{86.82} & 86.20 \\
\midrule
\multirow{4}{*}{LLaVA-NeXT~7B} & \textit{Random} & 85.57 & \underline{89.01} & 86.25 & 86.28 & 83.19 & 87.17 & 86.40 & 85.60 & \textbf{89.50} & 87.68 \\
  & \textit{Popular} & 84.64 & \underline{87.11} & 85.26 & 85.26 & 82.42 & 86.08 & 85.47 & 84.54 & \textbf{88.42} & 86.59 \\
  & \textit{Adversarial} & 83.84 & \underline{85.72} & 84.38 & 84.41 & 81.78 & 84.93 & 84.71 & 83.68 & \textbf{86.67} & 85.44 \\
\cmidrule(l){2-12}
  & \textit{Average} & 84.68 & \underline{87.28} & 85.30 & 85.32 & 82.46 & 86.06 & 85.53 & 84.61 & \textbf{88.20} & 86.57 \\
\midrule
\multirow{4}{*}{LLaVA-OV~7B} & \textit{Random} & 89.32 & \underline{92.68} & 90.05 & 89.76 & 89.28 & 89.78 & 89.58 & 88.04 & \textbf{93.23} & 91.68 \\
  & \textit{Popular} & 88.13 & \textbf{90.15} & 88.70 & 88.51 & 88.09 & 88.59 & 88.52 & 87.11 & 89.68 & \underline{89.74} \\
  & \textit{Adversarial} & 86.99 & \underline{87.79} & 87.50 & 87.32 & 86.94 & 87.58 & 87.27 & 85.88 & 86.78 & \textbf{88.11} \\
\cmidrule(l){2-12}
  & \textit{Average} & 88.15 & \textbf{90.20} & 88.75 & 88.53 & 88.10 & 88.65 & 88.46 & 87.01 & \underline{89.90} & 89.84 \\
\midrule
\multirow{4}{*}{Qwen2-VL~7B} & \textit{Random} & 89.04 & 89.17 & \textbf{89.29} & \underline{89.20} & 88.54 & 89.04 & 88.84 & 87.62 & 88.96 & 89.09 \\
  & \textit{Popular} & 87.72 & 87.70 & \textbf{87.97} & \underline{87.86} & 87.14 & 87.51 & 87.43 & 86.24 & 87.30 & 87.65 \\
  & \textit{Adversarial} & 86.43 & \underline{86.56} & \textbf{86.58} & 86.48 & 85.58 & 86.23 & 86.13 & 85.15 & 86.10 & 86.31 \\
\cmidrule(l){2-12}
  & \textit{Average} & 87.73 & 87.81 & \textbf{87.95} & \underline{87.85} & 87.09 & 87.59 & 87.47 & 86.34 & 87.45 & 87.68 \\
\midrule
\multirow{4}{*}{Qwen2.5-VL~7B} & \textit{Random} & 86.74 & 87.36 & \underline{88.66} & 86.88 & 85.33 & \textbf{89.01} & 86.80 & 84.48 & 87.35 & 87.34 \\
  & \textit{Popular} & 85.94 & 86.32 & \underline{87.73} & 86.04 & 84.56 & \textbf{88.14} & 85.97 & 83.68 & 86.10 & 86.31 \\
  & \textit{Adversarial} & 85.22 & 85.49 & \textbf{86.71} & 85.25 & 83.70 & \underline{86.59} & 85.22 & 83.28 & 85.15 & 85.51 \\
\cmidrule(l){2-12}
  & \textit{Average} & 85.97 & 86.39 & \underline{87.70} & 86.06 & 84.53 & \textbf{87.92} & 86.00 & 83.81 & 86.20 & 86.39 \\
\midrule
\multirow{4}{*}{Qwen3-VL~8B} & \textit{Random} & 90.46 & \underline{91.57} & 90.12 & 90.58 & 90.58 & 91.45 & 90.50 & 90.42 & \textbf{92.35} & \underline{91.57} \\
  & \textit{Popular} & 88.24 & 88.90 & 88.33 & 88.27 & 88.62 & 88.88 & 88.13 & 88.05 & \textbf{89.43} & \underline{89.00} \\
  & \textit{Adversarial} & 86.45 & 86.79 & 86.77 & 86.48 & 86.55 & \underline{86.97} & 86.40 & 86.32 & \textbf{87.16} & 86.95 \\
\cmidrule(l){2-12}
  & \textit{Average} & 88.39 & 89.09 & 88.40 & 88.44 & 88.58 & 89.10 & 88.34 & 88.27 & \textbf{89.65} & \underline{89.17} \\
\midrule
\multirow{4}{*}{InternVL3~8B} & \textit{Random} & 93.92 & 93.74 & 93.85 & 93.99 & 93.88 & 94.15 & 94.00 & 93.99 & \underline{94.22} & \textbf{94.61} \\
  & \textit{Popular} & 90.47 & 90.44 & \textbf{90.69} & 90.57 & 90.32 & 90.25 & 90.61 & \textbf{90.69} & \underline{90.63} & 89.69 \\
  & \textit{Adversarial} & 87.41 & 87.59 & \textbf{88.17} & 87.53 & 87.06 & 87.02 & 87.43 & \underline{87.89} & 87.40 & 86.49 \\
\cmidrule(l){2-12}
  & \textit{Average} & 90.60 & 90.59 & \textbf{90.90} & 90.69 & 90.42 & 90.47 & 90.68 & \underline{90.86} & 90.75 & 90.26 \\
\midrule
\multirow{4}{*}{InstructBLIP~7B} & \textit{Random} & 86.98 & 85.28 & 84.71 & 86.99 & 84.37 & 83.86 & \underline{87.07} & 83.02 & \textbf{87.88} & \textbf{87.88} \\
  & \textit{Popular} & 84.31 & 83.01 & 82.49 & 84.32 & 82.29 & 81.97 & \underline{84.49} & 76.94 & \textbf{84.73} & \textbf{84.73} \\
  & \textit{Adversarial} & \underline{82.13} & 80.80 & 81.27 & \textbf{82.17} & 81.31 & 80.74 & 82.05 & 75.77 & 81.85 & 81.85 \\
\cmidrule(l){2-12}
  & \textit{Average} & 84.47 & 83.03 & 82.82 & 84.49 & 82.66 & 82.19 & \underline{84.53} & 78.58 & \textbf{84.82} & \textbf{84.82} \\
\bottomrule
\end{tabular}

\end{table*}

\begin{table*}[!t]
\centering
\caption{CHAIR results on 500 MSCOCO validation images across eight LVLMs. $C_S$ and $C_I$ denote sentence- and instance-level hallucination rates (\%), Recall and F1 denote object recall and F1 (\%), and Words is mean caption length. Bold and underline mark the best and second-best results per row. $^{\ast}$: five-beam search.}
\label{tab:chair}
\small
\setlength{\tabcolsep}{2.414pt}
\begin{tabular}{ll c c c c c c c c c c}
\toprule
\multirow{2}{*}{Model} & \multirow{2}{*}{Metric} & Greedy & VCD & M3ID & OPERA$^{\ast}$ & ICD & SID & AGLA & ReVisiT & \textbf{IKOD} & \textbf{IKOD-A} \\
  &  & \textit{(Baseline)} & \textit{(CVPR'24)} & \textit{(CVPR'24)} & \textit{(CVPR'24)} & \textit{(ACL'24)} & \textit{(ICLR'25)} & \textit{(CVPR'25)} & \textit{(ACL'26)} & \textit{(Ours)} & \textit{(Ours)} \\
\midrule
\multirow{5}{*}{LLaVA-1.5~7B} & $C_S\downarrow$ & 51.4 & 56.4 & 49.4 & \textbf{37.4} & 54.0 & 51.6 & 50.2 & 53.6 & \underline{40.2} & 47.6 \\
  & $C_I\downarrow$ & 15.9 & 17.0 & 15.5 & \textbf{12.3} & 16.8 & 15.0 & 15.6 & 16.1 & \underline{12.5} & 14.3 \\
  & Recall$\uparrow$ & 77.8 & 77.2 & \underline{77.9} & 74.3 & 76.6 & 75.7 & 77.1 & 76.6 & 76.1 & \textbf{79.3} \\
  & F1$\uparrow$ & 80.8 & 80.0 & 81.1 & 80.4 & 79.8 & 80.1 & 80.6 & 80.1 & \underline{81.4} & \textbf{82.4} \\
  & Words & 84.4 & 86.0 & 79.1 & 70.8 & 81.0 & 80.6 & 84.2 & 90.7 & 88.9 & 85.8 \\
\midrule
\multirow{5}{*}{LLaVA-NeXT~7B} & $C_S\downarrow$ & 30.4 & 37.4 & 33.2 & \underline{26.0} & 28.4 & 37.4 & 29.2 & 29.0 & \textbf{24.8} & 27.8 \\
  & $C_I\downarrow$ & 9.2 & 11.7 & 8.9 & 8.4 & 8.8 & 10.2 & 7.9 & \underline{7.0} & \textbf{6.1} & 7.3 \\
  & Recall$\uparrow$ & 61.1 & 63.9 & 65.3 & 59.5 & 60.4 & \underline{65.6} & 59.8 & \textbf{66.0} & 60.9 & 60.8 \\
  & F1$\uparrow$ & 73.0 & 74.1 & \underline{76.1} & 72.1 & 72.7 & 75.8 & 72.5 & \textbf{77.2} & 73.9 & 73.4 \\
  & Words & 159.8 & 158.0 & 153.7 & 148.9 & 152.2 & 152.5 & 157.8 & 172.5 & 168.2 & 157.2 \\
\midrule
\multirow{5}{*}{LLaVA-OV~7B} & $C_S\downarrow$ & 33.6 & 45.0 & 39.2 & 33.6 & 37.4 & 36.8 & 35.4 & \underline{29.8} & \textbf{25.8} & 33.2 \\
  & $C_I\downarrow$ & 8.2 & 12.2 & 8.2 & 7.9 & 8.8 & 9.4 & 8.0 & 7.4 & \underline{6.4} & \textbf{6.2} \\
  & Recall$\uparrow$ & 65.7 & \underline{66.2} & \textbf{66.9} & 62.3 & 64.5 & 64.8 & 65.4 & 64.0 & 59.7 & 64.6 \\
  & F1$\uparrow$ & \underline{76.6} & 75.5 & \textbf{77.4} & 74.3 & 75.6 & 75.6 & 76.5 & 75.7 & 72.9 & 76.5 \\
  & Words & 213.9 & 172.3 & 230.8 & 185.3 & 204.1 & 180.0 & 227.0 & 222.2 & 327.2 & 252.7 \\
\midrule
\multirow{5}{*}{Qwen2-VL~7B} & $C_S\downarrow$ & 25.2 & \textbf{19.6} & 27.2 & \underline{20.6} & 20.8 & 25.4 & 22.6 & 23.0 & 21.8 & 23.8 \\
  & $C_I\downarrow$ & 7.4 & 7.9 & 9.4 & 8.6 & \underline{7.0} & 8.1 & 7.3 & 11.9 & \textbf{4.0} & 8.4 \\
  & Recall$\uparrow$ & 66.9 & 54.6 & \underline{67.2} & 63.6 & 59.4 & 61.7 & 65.9 & 66.5 & 66.3 & \textbf{68.6} \\
  & F1$\uparrow$ & \underline{77.7} & 68.6 & 77.2 & 75.0 & 72.5 & 73.8 & 77.0 & 75.8 & \textbf{78.4} & \textbf{78.4} \\
  & Words & 84.2 & 65.1 & 79.8 & 78.7 & 90.2 & 72.3 & 88.2 & 89.4 & 103.8 & 86.1 \\
\midrule
\multirow{5}{*}{Qwen2.5-VL~7B} & $C_S\downarrow$ & 23.8 & 30.0 & 29.0 & \textbf{17.0} & 24.6 & 29.0 & 23.6 & \underline{21.2} & 23.2 & 25.4 \\
  & $C_I\downarrow$ & \underline{6.8} & 8.9 & 8.9 & 7.8 & 9.2 & 8.6 & 8.2 & 8.2 & \textbf{6.4} & 7.9 \\
  & Recall$\uparrow$ & 57.5 & \underline{58.0} & 57.7 & 47.9 & 55.5 & \textbf{58.6} & 54.9 & 52.4 & 56.9 & 56.5 \\
  & F1$\uparrow$ & \underline{71.1} & 70.9 & 70.7 & 63.0 & 68.9 & \textbf{71.4} & 68.7 & 66.7 & 70.8 & 70.0 \\
  & Words & 164.2 & 154.9 & 135.8 & 107.0 & 177.2 & 147.4 & 158.5 & 176.5 & 162.9 & 154.0 \\
\midrule
\multirow{5}{*}{Qwen3-VL~8B} & $C_S\downarrow$ & 53.6 & 58.8 & 54.6 & \textbf{48.8} & 55.8 & 55.0 & 51.8 & 51.4 & 52.0 & \underline{49.0} \\
  & $C_I\downarrow$ & 10.9 & 12.1 & 11.0 & 10.2 & 10.8 & 11.1 & 10.5 & 10.3 & \textbf{9.2} & \underline{9.7} \\
  & Recall$\uparrow$ & 74.5 & 75.4 & 74.5 & 74.1 & \underline{75.6} & \textbf{76.2} & 74.7 & \underline{75.6} & 75.0 & 74.4 \\
  & F1$\uparrow$ & 81.1 & 81.2 & 81.1 & 81.2 & 81.8 & \textbf{82.1} & 81.4 & \underline{82.0} & \textbf{82.1} & 81.6 \\
  & Words & 324.2 & 326.4 & 323.7 & 312.7 & 348.5 & 315.3 & 313.8 & 323.0 & 311.5 & 311.7 \\
\midrule
\multirow{5}{*}{InternVL3~8B} & $C_S\downarrow$ & 30.4 & 32.4 & 30.8 & 27.2 & 30.2 & 33.6 & \textbf{25.8} & 30.4 & \underline{27.0} & 27.4 \\
  & $C_I\downarrow$ & 9.1 & 10.1 & 8.8 & 7.8 & 8.2 & 10.0 & 8.1 & 8.4 & \underline{7.7} & \textbf{7.4} \\
  & Recall$\uparrow$ & 70.4 & 70.3 & \underline{70.6} & 70.1 & 69.4 & 70.2 & 67.7 & \textbf{71.1} & 70.2 & 70.3 \\
  & F1$\uparrow$ & 79.3 & 78.9 & 79.6 & 79.6 & 79.0 & 78.9 & 78.0 & \textbf{80.1} & 79.7 & \underline{79.9} \\
  & Words & 104.9 & 104.4 & 103.8 & 103.7 & 111.4 & 105.7 & 100.6 & 105.2 & 106.9 & 114.8 \\
\midrule
\multirow{5}{*}{InstructBLIP~7B} & $C_S\downarrow$ & 46.4 & 52.0 & 63.0 & \textbf{39.0} & 73.6 & 67.8 & 49.2 & \underline{41.6} & 44.0 & 44.0 \\
  & $C_I\downarrow$ & 13.3 & 15.4 & 16.8 & 13.5 & 25.4 & 19.6 & 13.7 & \underline{12.5} & \textbf{11.3} & 12.7 \\
  & Recall$\uparrow$ & 71.2 & \underline{72.6} & 71.9 & 69.3 & \textbf{72.8} & 72.4 & 71.9 & 70.0 & 71.7 & 72.0 \\
  & F1$\uparrow$ & 78.2 & 78.1 & 77.1 & 77.0 & 73.7 & 76.2 & 78.4 & 77.8 & \textbf{79.3} & \underline{78.9} \\
  & Words & 91.9 & 88.6 & 92.6 & 78.4 & 114.3 & 86.3 & 93.1 & 98.0 & 99.6 & 99.5 \\
\bottomrule
\end{tabular}
\vspace{2pt}
\end{table*}

\begin{table*}[!t]
\centering
\caption{Fine-grained discriminative and open-ended generative results on AMBER~\cite{wang2023amber} across eight LVLMs (\%). \textbf{(a)} Attr$\uparrow$ (state, number, action) and Rel$\uparrow$ report binary F1 on fine-grained object attributes and inter-object relations. \textbf{(b)} Hal$\downarrow$ and Cog$\downarrow$ denote hallucinated response rate and human-cognition-aligned hallucination rate on 1,004 images. Bold and underline mark the best and second-best results per column within each block. $^{\ast}$: five-beam search.}
\label{tab:amber}
\small
\setlength{\tabcolsep}{3.116pt}
\begin{tabular}{l cc cc cc cc cc cc cc cc !{\vrule width 0.4pt} cc}
\toprule
\multirow{2}{*}[-0.25ex]{Method} & \multicolumn{2}{c}{LLaVA-1.5} & \multicolumn{2}{c}{LLaVA-NeXT} & \multicolumn{2}{c}{LLaVA-OV} & \multicolumn{2}{c}{Qwen2-VL} & \multicolumn{2}{c}{Qwen2.5-VL} & \multicolumn{2}{c}{Qwen3-VL} & \multicolumn{2}{c}{InternVL3} & \multicolumn{2}{c !{\vrule width 0.4pt}}{InstructBLIP} & \multicolumn{2}{c}{\textbf{Average}} \\
\cmidrule(lr){2-3}\cmidrule(lr){4-5}\cmidrule(lr){6-7}\cmidrule(lr){8-9}\cmidrule(lr){10-11}\cmidrule(lr){12-13}\cmidrule(lr){14-15}\cmidrule(lr){16-17}\cmidrule(lr){18-19}
 & \multicolumn{18}{c}{\textit{(a) Fine-Grained Discriminative Probing: Attribute F1 (Attr$\uparrow$) / Relation F1 (Rel$\uparrow$)}} \\
\cmidrule(lr){2-19}
 & Attr$\uparrow$ & Rel$\uparrow$ & Attr$\uparrow$ & Rel$\uparrow$ & Attr$\uparrow$ & Rel$\uparrow$ & Attr$\uparrow$ & Rel$\uparrow$ & Attr$\uparrow$ & Rel$\uparrow$ & Attr$\uparrow$ & Rel$\uparrow$ & Attr$\uparrow$ & Rel$\uparrow$ & Attr$\uparrow$ & Rel$\uparrow$ & Attr$\uparrow$ & Rel$\uparrow$ \\
\midrule
Greedy & 77.5 & 41.2 & 81.4 & 52.0 & \textbf{88.0} & 77.4 & 86.2 & 74.4 & 83.0 & 72.3 & 87.4 & 87.1 & 86.4 & 80.2 & \textbf{79.7} & \underline{76.6} & 83.7 & 70.1 \\
VCD & 76.7 & 49.7 & 80.3 & 59.6 & 87.6 & 78.1 & 85.1 & 72.3 & 83.1 & 72.0 & 86.8 & 84.9 & 85.8 & 81.6 & \underline{79.6} & 76.3 & 83.1 & 71.8 \\
M3ID & 75.8 & 47.3 & 79.8 & 63.7 & \underline{87.9} & 80.9 & 85.9 & \underline{76.5} & 83.5 & \textbf{79.9} & 87.1 & \textbf{90.3} & 86.4 & 82.6 & 78.3 & 59.4 & 83.1 & 72.6 \\
OPERA$^{\ast}$ & 80.0 & \underline{67.5} & 82.2 & 70.3 & \textbf{88.0} & 77.0 & \textbf{86.3} & \underline{76.5} & \underline{84.7} & 76.5 & \underline{87.6} & 85.9 & 86.4 & 81.3 & 76.8 & 68.7 & \underline{84.0} & 75.5 \\
ICD & 79.2 & 67.7 & 81.5 & 67.9 & \underline{87.9} & 75.8 & 85.6 & 75.5 & 84.3 & 74.2 & 87.1 & 83.5 & 85.9 & 79.6 & \textbf{79.7} & 69.3 & 83.9 & 74.2 \\
SID & 77.5 & 58.0 & 80.8 & 61.5 & 87.4 & 80.4 & 85.6 & 74.6 & 83.9 & \underline{79.2} & 87.2 & \underline{90.1} & 86.4 & \underline{84.9} & 78.0 & 35.4 & 83.4 & 70.5 \\
AGLA & 79.4 & 64.5 & 81.7 & 69.1 & 87.6 & 77.1 & 86.0 & \textbf{77.0} & \textbf{84.8} & 75.8 & \textbf{87.8} & 86.2 & 86.5 & 81.8 & 76.1 & 69.7 & 83.7 & 75.2 \\
ReVisiT & 78.0 & 51.2 & 81.5 & 56.6 & 87.5 & 73.7 & 85.7 & 71.3 & 82.5 & 70.3 & 87.3 & 87.2 & 86.4 & 80.4 & 73.5 & 74.7 & 82.8 & 70.7 \\
\textbf{IKOD} & \underline{80.4} & \textbf{71.2} & \underline{82.9} & \textbf{82.9} & 84.2 & \textbf{85.8} & 86.0 & 76.1 & 83.6 & 73.5 & 87.1 & 88.3 & \textbf{86.9} & 82.1 & 77.9 & \textbf{78.3} & 83.6 & \textbf{79.8} \\
\textbf{IKOD-A} & \textbf{80.5} & 56.0 & \textbf{83.2} & \underline{72.1} & 87.3 & \underline{85.0} & \textbf{86.3} & 75.6 & 83.7 & 73.1 & 87.3 & 88.1 & \underline{86.7} & \textbf{85.9} & 77.9 & \textbf{78.3} & \textbf{84.1} & \underline{76.8} \\
\midrule
 & \multicolumn{18}{c}{\textit{(b) Open-Ended Generative Captioning: Hallucinated Response Rate (Hal$\downarrow$) / Cognitive Bias Rate (Cog$\downarrow$)}} \\
\cmidrule(lr){2-19}
 & Hal$\downarrow$ & Cog$\downarrow$ & Hal$\downarrow$ & Cog$\downarrow$ & Hal$\downarrow$ & Cog$\downarrow$ & Hal$\downarrow$ & Cog$\downarrow$ & Hal$\downarrow$ & Cog$\downarrow$ & Hal$\downarrow$ & Cog$\downarrow$ & Hal$\downarrow$ & Cog$\downarrow$ & Hal$\downarrow$ & Cog$\downarrow$ & Hal$\downarrow$ & Cog$\downarrow$ \\
\midrule
Greedy & 32.4 & 3.7 & 51.7 & 4.6 & 69.1 & 8.0 & 41.0 & 3.8 & 20.5 & 0.9 & \underline{60.0} & 3.9 & 72.2 & 8.9 & 34.9 & 3.7 & 47.7 & 4.7 \\
VCD & 38.4 & 4.4 & 61.1 & 5.7 & 69.3 & 6.7 & 45.2 & 3.9 & 24.0 & 1.0 & 65.3 & 4.1 & 77.2 & 10.2 & 40.4 & 3.5 & 52.6 & 4.9 \\
M3ID & 32.5 & 3.2 & 50.6 & 4.4 & 68.7 & 7.0 & 44.6 & 4.0 & 21.7 & 1.1 & 61.3 & 3.9 & 73.7 & 8.3 & 36.6 & 3.8 & 48.7 & 4.5 \\
OPERA$^{\ast}$ & \textbf{24.4} & \textbf{2.6} & 50.7 & 5.0 & 67.9 & 9.3 & 40.7 & 4.2 & \textbf{13.7} & \textbf{0.5} & \textbf{55.4} & \textbf{3.0} & 72.2 & 10.3 & \textbf{27.7} & \textbf{2.5} & 44.1 & 4.7 \\
ICD & 33.4 & 4.0 & 52.8 & 4.7 & 67.3 & 7.9 & 61.8 & 7.1 & 25.2 & 1.3 & 66.7 & 4.3 & 78.0 & 9.9 & 44.0 & 5.0 & 53.6 & 5.5 \\
SID & 31.7 & 3.3 & 50.8 & 4.1 & 67.4 & 7.0 & 40.9 & \underline{3.5} & 24.8 & 0.9 & 63.2 & 3.8 & 74.1 & 8.4 & 42.0 & 4.5 & 49.4 & 4.4 \\
AGLA & 32.1 & 3.7 & 51.8 & 4.8 & 68.5 & 8.4 & \underline{40.4} & 3.8 & \underline{19.2} & 0.9 & \underline{60.0} & \underline{3.5} & 69.1 & 8.8 & 33.6 & 4.1 & 46.8 & 4.8 \\
ReVisiT & 38.5 & 3.8 & \underline{50.3} & 5.4 & 61.8 & 6.7 & 43.8 & 4.3 & 23.1 & 1.1 & 62.0 & \underline{3.5} & \underline{68.6} & \underline{8.2} & 34.0 & 3.3 & 47.8 & 4.5 \\
\textbf{IKOD} & \underline{26.1} & \underline{2.7} & \textbf{40.7} & \textbf{3.0} & \textbf{41.2} & \textbf{3.4} & \textbf{39.8} & \textbf{3.4} & 21.8 & 1.0 & 61.4 & 3.7 & 70.7 & 8.8 & 30.9 & \underline{3.2} & \textbf{41.6} & \textbf{3.6} \\
\textbf{IKOD-A} & 30.1 & 3.4 & 45.5 & \underline{3.5} & \underline{53.7} & \underline{5.9} & 41.0 & 3.9 & 20.9 & \underline{0.8} & 61.0 & 3.8 & \textbf{63.0} & \textbf{6.5} & \underline{30.0} & 3.3 & \underline{43.2} & \underline{3.9} \\
\bottomrule
\end{tabular}
\end{table*}

\subsection{Position Encoding and Computational Cost}
\label{sec:impl}
IKOD operates purely on the KV cache and logits without modifying model weights or attention kernels. Under 1D Rotary Position Embedding (RoPE) in LLaVA, InternVL3, and InstructBLIP, cached keys and values keep their original rotary phases while the query token on the merged branch uses the shortened cache length as its position index. For models with 3D Multimodal RoPE (M-RoPE) in the Qwen-VL family, the query token retains its original 3D temporal-spatial coordinates. With visual prefix length $U$ and $P = |\mathcal{P}_t|$ post-visual text tokens, the merged branch attends over $U + n_t + 1$ rather than $U + P + 1$ entries. Two lossless mechanisms minimize runtime overhead while preserving the exact output trajectory: (1) a \emph{plausibility-singleton short-circuit} skips KV merging and the auxiliary pass whenever $|\mathcal{V}_t \setminus \mathcal{E}| = 1$ (or $\alpha_t = 0$); and (2) a \emph{persistent visual-prefix buffer} caches $[1:U]$ once after prefill and updates only the $n_t$ centroids in-place across layers, as detailed in Section~\ref{sec:discussion} and the Supplementary Material.
\setcounter{topnumber}{4}
\setcounter{bottomnumber}{2}
\setcounter{totalnumber}{6}
\renewcommand{\topfraction}{0.9}
\renewcommand{\bottomfraction}{0.8}
\renewcommand{\textfraction}{0.08}
\renewcommand{\floatpagefraction}{0.85}
\renewcommand{\dbltopfraction}{0.9}
\renewcommand{\dblfloatpagefraction}{0.85}

\section{Experiments}
\label{sec:exp}

We evaluate IKOD and IKOD-A across probing, captioning, and general multimodal benchmarks to answer three questions: (1) Do IKOD and IKOD-A consistently curb hallucination across diverse LVLMs and decoding regimes? (2) Do they preserve general multimodal perception and reasoning? (3) How do anchor selection, mergeable region, bucket averaging, and collaborative fusion contribute to grounding and efficiency?

\subsection{Experimental Setup}
\label{sec:setup}
\noindent\textbf{Backbones.} 
We evaluate on eight open-source LVLMs across four model families: LLaVA-1.5~7B~\cite{LLaVA-1.5}, LLaVA-NeXT~7B~\cite{llavanext2024}, and LLaVA-OneVision~7B (LLaVA-OV)~\cite{llavaonevision2024} from the LLaVA family; Qwen2-VL~7B~\cite{qwen2vl2024}, Qwen2.5-VL~7B~\cite{qwen25vl2025}, and Qwen3-VL~8B~\cite{qwen3vl2025} from the Qwen-VL family; InternVL3~8B~\cite{internvl3_2025}; and InstructBLIP~7B~\cite{InstructBLIP}.

\noindent\textbf{Benchmarks.} 
We evaluate object and fine-grained hallucination on POPE~\cite{li2023evaluating}, CHAIR~\cite{rohrbach2018object}, and AMBER~\cite{wang2023amber}, alongside general multimodal understanding on MME perception~\cite{MME}, MMStar~\cite{mmstar2024}, MMBench-EN (dev)~\cite{mmbench2024}, and ScienceQA~\cite{scienceqa2022}. POPE probes object existence via 3,000 binary questions per split across the \textit{Random}, \textit{Popular}, and \textit{Adversarial} negative-sampling splits on 500 MSCOCO validation images~\cite{MSCOCO}. CHAIR evaluates 500 detailed MSCOCO captions on sentence-level ($C_S$) and instance-level ($C_I$) hallucination rates, object recall, object F1, and mean length (Words). AMBER evaluates attribute and relation F1 alongside generative hallucination (Hal) and cognitive bias (Cog) on 1,004 images; metric details are in the Supplementary Material.

\noindent\textbf{Baselines.} 
We compare IKOD and IKOD-A against Greedy decoding and seven representative training-free decoding methods: VCD~\cite{vcd}, M3ID~\cite{favero2024multi}, OPERA~\cite{huang2024opera}, ICD~\cite{icd2024}, SID~\cite{sid2025}, AGLA~\cite{agla}, and ReVisiT~\cite{revisit2025}. These baselines cover visual and instruction contrastive/ensemble decoding~\cite{vcd,favero2024multi,icd2024,sid2025,agla}, attention-pattern calibration~\cite{huang2024opera}, and internal visual-representation intervention~\cite{revisit2025}. Unless noted, all methods run under greedy decoding in a unified Hugging Face codebase (except OPERA, which requires five-beam search for rollback), with other sampling strategies evaluated in Section~\ref{sec:sampling}. For VCD, M3ID, OPERA, ICD, SID, and AGLA, we use default hyperparameters from the original papers on all backbones, and for ReVisiT its published exit-layer settings; full details are in the Supplementary Material.

\noindent\textbf{Implementation details.} 
We set the plausibility threshold to $\beta=0.1$ following prior work~\cite{vcd,icd2024}, and fix the retained anchor ratio to $\lambda=0.4$ across all architectures and benchmarks. IKOD sets only $\alpha$ by model family on POPE ($\alpha=2.0$ for LLaVA and Qwen-VL, $0.1$ for InternVL3, and $0.5$ for InstructBLIP) and transfers it unchanged to all evaluations, while IKOD-A fixes $(\alpha_{\min}, \alpha_{\max}, \kappa) = (0, 1, 10)$ across all models. Default maximum generation length is 12 tokens on POPE and 512 on CHAIR on A100 GPUs.

\subsection{Main Results}
\label{sec:main}

\noindent\textbf{Results on POPE.} 
Table~\ref{tab:pope} reports POPE-COCO F1 across all eight LVLMs. IKOD and IKOD-A lift eight-backbone mean F1 from 86.95\% under Greedy to 87.97\% and 87.62\%, topping all seven baselines and ranking first and second on \textit{Random}, \textit{Popular}, and \textit{Adversarial}. Even on single-token queries, folding low-$s_m$ function words in $\mathcal{P}_1$ curbs softmax competition against the image prefix, lifting F1 by 3.52 points on LLaVA-NeXT and 1.75 on LLaVA-OneVision and improving 19 of 24 split evaluations under IKOD-A. By contrast, contrastive methods (VCD, M3ID, ICD, and SID) ease ``no'' bias on LLaVA-NeXT and Qwen2.5-VL via logit subtraction but induce false positives on LLaVA-1.5 and drop InstructBLIP by 1.44--2.28 points, while OPERA and AGLA yield minor single-token gains (87.14\% and 87.09\%) and ReVisiT drops seven backbones.

\noindent\textbf{Results on CHAIR.} 
In Table~\ref{tab:chair}, IKOD and IKOD-A curb open-ended CHAIR hallucinations without shortening captions, generating 171.1 and 157.7 words versus 153.4 under Greedy while cutting eight-backbone mean $C_I$ from 10.10\% to 7.95\% and 9.24\% and lifting object F1 from 77.22\% to 77.31\% and 77.64\%. As $\mathcal{P}_t$ grows over hundreds of tokens, merging low-$s_m$ text entries bounds post-visual softmax competition under Theorem~\ref{thm:share} and cuts both $C_I$ and $C_S$ below Greedy on every backbone, lowering mean $C_S$ from 36.85\% to 32.35\%. Whereas static $\alpha=2.0$ over-constrains early mentions on LLaVA-OneVision and lowers recall from 65.7\% to 59.7\%, IKOD-A ramps $\alpha_t$ up only as $S_t$ decays, recovering recall to 64.6\% and raising mean recall to 68.31\%. By contrast, OPERA cuts $C_S$ to 31.20\% by truncating captions to 135.7 words and losing 3.0 recall points, logit subtraction in VCD, M3ID, ICD, and SID accumulates drift and raises $C_I$ to 10.94--11.91\%, and AGLA and ReVisiT drop object F1 below Greedy to 76.64\% and 76.92\%.

\noindent\textbf{Results on AMBER.} 
Table~\ref{tab:amber} further evaluates fine-grained attribute and inter-object relation probing~(a) alongside open-ended generative hallucination~(b) on AMBER~\cite{wang2023amber}. Whereas VCD, M3ID, SID, and ReVisiT lower mean attribute F1 from 83.7\% under Greedy to 82.8\%--83.4\%, folding low-$s_m$ prompt tokens refocuses attention onto fine-grained visual cues and lifts attribute F1 to 84.1\% under IKOD-A ahead of OPERA (84.0\%), ICD (83.9\%), and AGLA (83.7\%). On relation queries where both objects appear in the image, refocusing spatial attention overcomes co-occurrence bias on LLaVA-1.5 and LLaVA-NeXT (+30.0 and +30.9), lifting mean relation F1 from 70.1\% under Greedy and 70.5\%--75.5\% across all seven baselines to 79.8\% and 76.8\%. In generation~(b), while VCD, M3ID, ICD, and SID raise Hal to 48.7\%--53.6\%, AGLA and ReVisiT stay at 46.8\%--47.8\%, and OPERA leaves Cog at 4.7\%, IKOD and IKOD-A rank first and second in Hal (41.6\% and 43.2\% vs.\ 47.7\%) and Cog (3.6\% and 3.9\%), cutting Hal by 27.9, 11.0, and 9.2 points on LLaVA-OneVision, LLaVA-NeXT, and InternVL3.

\noindent\textbf{General multimodal understanding and reasoning.} 
Table~\ref{tab:noreg} evaluates general multimodal perception and reasoning across MME$^{P}$, MMStar, MMBench, and ScienceQA. With the fixed POPE weights, IKOD improves MME$^{P}$ on Qwen2.5-VL from 1,639 to 1,655 and on Qwen3-VL from 1,712 to 1,727, edges up LLaVA-1.5 on MMStar to 33.8\%, and keeps ScienceQA accuracy within 0.46 points of Greedy across the seven ViT-MLP backbones. Because short VQA and multiple-choice answers accumulate little decay, a large static weight can mildly perturb option logits, whereas IKOD-A applies a gentle gate under low decay: it lifts Qwen2.5-VL to 58.3\% on MMStar and LLaVA-NeXT to 66.8\% on MMBench, recovers LLaVA-OneVision from 1,354 to 1,506 on MME$^{P}$, and retains 99.0\% of Greedy across the seven ViT-MLP backbones.

\begin{table*}[!t]
\centering
\caption{General multimodal understanding and reasoning results across all eight LVLMs. MME$^{P}$ denotes the MME perception score; MMS, MMB, and SQA denote MMStar, MMBench-EN (dev), and ScienceQA accuracy (\%).}
\label{tab:noreg}
\small
\setlength{\tabcolsep}{1.135pt}
\setlength{\cmidrulekern}{1.135pt}
\begin{tabular}{l cc cc cc cc cc cc cc cc !{\vrule width 0.4pt} cc}
\toprule
\multirow{2}{*}[-0.25ex]{Method} & \multicolumn{2}{c}{LLaVA-1.5} & \multicolumn{2}{c}{LLaVA-NeXT} & \multicolumn{2}{c}{LLaVA-OV} & \multicolumn{2}{c}{Qwen2-VL} & \multicolumn{2}{c}{Qwen2.5-VL} & \multicolumn{2}{c}{Qwen3-VL} & \multicolumn{2}{c}{InternVL3} & \multicolumn{2}{c !{\vrule width 0.4pt}}{InstructBLIP} & \multicolumn{2}{c}{\textbf{Average}} \\
\cmidrule(lr){2-3}\cmidrule(lr){4-5}\cmidrule(lr){6-7}\cmidrule(lr){8-9}\cmidrule(lr){10-11}\cmidrule(lr){12-13}\cmidrule(lr){14-15}\cmidrule(lr){16-17}\cmidrule(lr){18-19}
 & \multicolumn{18}{c}{\textit{(a) Perception \& Vision-Indispensable Reasoning: MME Perception (MME$^{P}\uparrow$) / MMStar (MMS$\uparrow$)}} \\
\cmidrule(lr){2-19}
 & MME$^{P}$ & MMS & MME$^{P}$ & MMS & MME$^{P}$ & MMS & MME$^{P}$ & MMS & MME$^{P}$ & MMS & MME$^{P}$ & MMS & MME$^{P}$ & MMS & MME$^{P}$ & MMS & MME$^{P}$ & MMS \\
\midrule
Greedy & \textbf{1455} & 33.3 & \textbf{1525} & \textbf{37.5} & \textbf{1559} & \textbf{58.9} & \textbf{1681} & \textbf{57.5} & 1639 & \underline{58.2} & 1712 & \textbf{64.2} & \textbf{1701} & \textbf{66.2} & \textbf{1247} & \textbf{3.7} & \textbf{1565} & \textbf{47.4} \\
\textbf{IKOD} & 1364 & \textbf{33.8} & 1434 & \underline{36.8} & 1354 & 54.8 & \underline{1678} & 51.6 & \textbf{1655} & 57.5 & \textbf{1727} & 60.2 & \underline{1696} & \underline{65.8} & \underline{1193} & \underline{0.4} & 1513 & 45.1 \\
\textbf{IKOD-A} & \underline{1438} & \underline{33.6} & \underline{1486} & 36.6 & \underline{1506} & \underline{57.9} & 1672 & \underline{54.2} & \textbf{1655} & \textbf{58.3} & \underline{1722} & \underline{63.8} & 1688 & \underline{65.8} & \underline{1193} & \underline{0.4} & \underline{1545} & \underline{46.3} \\
\midrule
 & \multicolumn{18}{c}{\textit{(b) General Multimodal Comprehension \& Science Reasoning: MMBench-EN (MMB$\uparrow$) / ScienceQA-IMG (SQA$\uparrow$)}} \\
\cmidrule(lr){2-19}
 & MMB & SQA & MMB & SQA & MMB & SQA & MMB & SQA & MMB & SQA & MMB & SQA & MMB & SQA & MMB & SQA & MMB & SQA \\
\midrule
Greedy & \textbf{61.9} & \textbf{68.6} & \underline{66.7} & \textbf{73.3} & \textbf{81.1} & \textbf{89.6} & \textbf{78.7} & \textbf{85.3} & \textbf{80.8} & \textbf{81.0} & \textbf{84.6} & \textbf{92.9} & \textbf{84.3} & \textbf{96.4} & \textbf{13.4} & \textbf{38.6} & \textbf{68.9} & \textbf{78.2} \\
\textbf{IKOD} & 54.7 & 67.6 & 62.7 & 72.7 & 76.1 & \underline{89.5} & 73.9 & 83.9 & 77.5 & \underline{80.9} & 83.2 & \textbf{92.9} & \underline{84.1} & \textbf{96.4} & \underline{6.2} & \underline{27.6} & 64.8 & 76.4 \\
\textbf{IKOD-A} & \underline{60.0} & \underline{68.4} & \textbf{66.8} & \underline{72.9} & \underline{80.8} & \textbf{89.6} & \underline{77.3} & \underline{84.2} & \underline{79.6} & \textbf{81.0} & \underline{83.8} & \textbf{92.9} & 82.0 & \textbf{96.4} & \underline{6.2} & \underline{27.6} & \underline{67.1} & \underline{76.6} \\
\bottomrule
\end{tabular}
\end{table*}

\begin{table*}[!t]
\centering
\caption{Component ablations on attention-guided post-visual KV merging on CHAIR ($C_I{\downarrow}$ / F1${\uparrow}$, \%). Each variant in (a)--(c) alters a single design factor while holding all other settings fixed to full IKOD. $^{\dagger}$At $t{=}1$, $|\mathcal{P}_{\mathrm{gen}}|{=}0$ (degenerating to Greedy at 86.95\% POPE F1 vs.\ 87.97\% for IKOD).}
\label{tab:abl_kv}
\small
\setlength{\tabcolsep}{1.55pt}
\setlength{\cmidrulekern}{1.55pt}
\begin{tabular}{l cc cc cc cc cc cc cc cc !{\vrule width 0.4pt} cc}
\toprule
\multirow{2}{*}[-0.25ex]{Setting} & \multicolumn{2}{c}{LLaVA-1.5} & \multicolumn{2}{c}{LLaVA-NeXT} & \multicolumn{2}{c}{LLaVA-OV} & \multicolumn{2}{c}{Qwen2-VL} & \multicolumn{2}{c}{Qwen2.5-VL} & \multicolumn{2}{c}{Qwen3-VL} & \multicolumn{2}{c}{InternVL3} & \multicolumn{2}{c !{\vrule width 0.4pt}}{InstructBLIP} & \multicolumn{2}{c}{\textbf{Average}} \\
\cmidrule(lr){2-3}\cmidrule(lr){4-5}\cmidrule(lr){6-7}\cmidrule(lr){8-9}\cmidrule(lr){10-11}\cmidrule(lr){12-13}\cmidrule(lr){14-15}\cmidrule(lr){16-17}\cmidrule(lr){18-19}
 & $C_I{\downarrow}$ & F1${\uparrow}$ & $C_I{\downarrow}$ & F1${\uparrow}$ & $C_I{\downarrow}$ & F1${\uparrow}$ & $C_I{\downarrow}$ & F1${\uparrow}$ & $C_I{\downarrow}$ & F1${\uparrow}$ & $C_I{\downarrow}$ & F1${\uparrow}$ & $C_I{\downarrow}$ & F1${\uparrow}$ & $C_I{\downarrow}$ & F1${\uparrow}$ & $C_I{\downarrow}$ & F1${\uparrow}$ \\
\midrule
Greedy (base) & 15.9 & 80.8 & 9.2 & 73.0 & 8.2 & 76.6 & 7.4 & 77.7 & 6.8 & 71.1 & 10.9 & 81.1 & 9.1 & 79.3 & 13.3 & 78.2 & 10.1 & 77.2 \\
\midrule
\multicolumn{19}{l}{\textit{(a) Anchor selection criterion}} \\
\hspace{0.4em}High-attention & \textbf{10.4} & 81.0 & 7.5 & 72.0 & \textbf{5.4} & 73.7 & 8.2 & 75.7 & 7.8 & 71.5 & 9.4 & 81.7 & 8.0 & \textbf{79.8} & \textbf{10.3} & \textbf{79.8} & 8.4 & 76.9 \\
\hspace{0.4em}Uniform stride & 12.9 & \textbf{81.4} & 6.6 & \textbf{74.5} & \textbf{5.4} & \textbf{75.2} & 5.5 & 77.2 & 7.3 & \textbf{72.4} & \textbf{9.2} & \textbf{82.2} & 8.8 & 79.3 & 11.4 & 79.0 & 8.4 & \textbf{77.7} \\
\hspace{0.4em}\textbf{Low-attention (IKOD)} & 12.5 & \textbf{81.4} & \textbf{6.1} & 73.9 & 6.4 & 72.9 & \textbf{4.0} & \textbf{78.4} & \textbf{6.4} & 70.8 & \textbf{9.2} & 82.1 & \textbf{7.7} & 79.7 & 11.3 & 79.3 & \textbf{8.0} & 77.3 \\
\midrule
\multicolumn{19}{l}{\textit{(b) Mergeable post-visual text region $\mathcal{P}_t$}} \\
\hspace{0.4em}Prompt suffix & 15.1 & 80.7 & 6.6 & \textbf{74.3} & 8.2 & \textbf{77.6} & 7.4 & \textbf{78.5} & 12.4 & \textbf{71.2} & 10.3 & 81.4 & 8.7 & 79.6 & 13.4 & 78.1 & 10.3 & \textbf{77.7} \\
\hspace{0.4em}Generated history$^{\dagger}$ & \textbf{11.2} & 80.8 & 6.2 & 73.0 & \textbf{6.4} & 74.0 & 5.2 & 78.0 & \textbf{6.2} & 70.9 & 9.3 & 81.5 & 8.0 & 79.2 & \textbf{10.3} & \textbf{79.5} & \textbf{7.9} & 77.1 \\
\hspace{0.4em}\textbf{All post-visual (IKOD)} & 12.5 & \textbf{81.4} & \textbf{6.1} & 73.9 & \textbf{6.4} & 72.9 & \textbf{4.0} & 78.4 & 6.4 & 70.8 & \textbf{9.2} & \textbf{82.1} & \textbf{7.7} & \textbf{79.7} & 11.3 & 79.3 & 8.0 & 77.3 \\
\midrule
\multicolumn{19}{l}{\textit{(c) Token aggregation operator}} \\
\hspace{0.4em}Prune (high-attn) & \textbf{10.9} & 80.8 & 7.4 & 72.2 & 6.8 & 73.2 & 4.6 & 78.0 & \textbf{6.2} & \textbf{71.6} & \textbf{9.1} & 81.8 & \textbf{7.6} & 79.6 & 15.3 & 78.5 & 8.5 & 77.0 \\
\hspace{0.4em}Prune (low-attn) & 15.4 & 80.7 & 7.6 & 72.9 & \textbf{5.6} & \textbf{75.2} & 7.9 & 77.7 & 8.6 & 70.6 & 9.2 & 81.3 & 7.9 & \textbf{79.7} & 11.6 & 78.6 & 9.2 & 77.1 \\
\hspace{0.4em}\textbf{Bucket centroid (IKOD)} & 12.5 & \textbf{81.4} & \textbf{6.1} & \textbf{73.9} & 6.4 & 72.9 & \textbf{4.0} & \textbf{78.4} & 6.4 & 70.8 & 9.2 & \textbf{82.1} & 7.7 & \textbf{79.7} & \textbf{11.3} & \textbf{79.3} & \textbf{8.0} & \textbf{77.3} \\
\bottomrule
\end{tabular}
\end{table*}

\subsection{Ablation Studies}
\label{sec:aba}
We conduct ablations across all eight LVLMs to evaluate the core design choices and hyperparameter sensitivity of IKOD in Section~\ref{sec:method}.

\noindent\textbf{KV merging design.} 
Table~\ref{tab:abl_kv}(a)--(c) and Figure~\ref{fig:mech8b}(c) dissect the three structural choices of merged-cache construction across all eight backbones. In group~(a), lowest-attention anchors target uninformative function words and punctuation, comprising 55.6\% of anchors versus 4.2\% grounded nouns, and partition late decayed spans into fine 3.7-token buckets, lifting visual attention at hallucinated steps by 6.2 percentage points to achieve 8.0\% $C_I$ and 77.3\% F1. By contrast, highest-attention anchors cluster in early grounded tokens and leave late spans in coarse 10.2-token buckets with only a 2.4-point attention gain at 8.4\% $C_I$ and 76.9\% F1, while uniform stride ignores local decay at 8.4\% $C_I$. In group~(b), merging only the prompt suffix leaves $C_I$ at 10.3\%, whereas merging only generated history reaches 7.9\% $C_I$ on long captions but slips to 77.1\% F1 and is inactive on single-token VQA where $|\mathcal{P}_{\mathrm{gen}}|=0$; merging the full post-visual region $\mathcal{P}_t$ unifies both regimes. In group~(c), hard pruning degrades grounding: keeping only low-attention anchors discards intra-bucket object nouns, dropping grounded value cosine similarity from 76.7\% to 68.1\% and raising $C_I$ to 9.2\%, while keeping only high-attention tokens disrupts phrase continuity at 8.5\% $C_I$ and 77.0\% F1, confirming that bucket centroids compress context while preserving 91.2\% output similarity.

\begin{table}[!htb]
\centering
\caption{Ablation on dual-branch collaborative decoding and fusion sign. Eight-LVLM mean POPE F1 and CHAIR metrics (\%) at $\lambda{=}0.4$. Bold and underline mark the best and second-best results.}
\label{tab:abl_collab}
\small
\setlength{\tabcolsep}{5.038pt}
\begin{tabular}{ccc ccccc}
\toprule
$\ell_t^{\mathrm{base}}$ & $\ell_t^{\mathrm{merged}}$ & Sign & POPE$\uparrow$ & $C_S\downarrow$ & $C_I\downarrow$ & Recall$\uparrow$ & F1$\uparrow$ \\
\midrule
\checkmark & $\times$ & -- & \underline{86.95} & 36.9 & 10.1 & \textbf{68.1} & \underline{77.2} \\
$\times$ & \checkmark & -- & 86.46 & \textbf{31.0} & \underline{8.5} & 66.1 & 76.4 \\
\checkmark & \checkmark & $-\alpha$ & 79.30 & 41.8 & 15.7 & 60.5 & 70.0 \\
\checkmark & \checkmark & $+\alpha$ & \textbf{87.97} & \underline{32.4} & \textbf{8.0} & \underline{67.1} & \textbf{77.3} \\
\bottomrule
\end{tabular}
\end{table}

\noindent\textbf{Collaborative decoding.} 
Table~\ref{tab:abl_collab} isolates the dual-branch fusion design in Section~\ref{sec:co} across all eight backbones. Decoding solely from the merged branch ($\ell_t^{\mathrm{merged}}$ only) lowers mean CHAIR $C_I$ from 10.1\% to 8.5\% and $C_S$ to 31.0\%, yet coarse-bucket aggregation sacrifices lexical precision, dropping POPE F1 from 86.95\% to 86.46\% and CHAIR object F1 from 77.2\% to 76.4\%. Reversing the fusion sign to contrastive subtraction ($\ell_t^{\mathrm{base}} - \alpha\ell_t^{\mathrm{merged}}$) severely penalizes visually refocused predictions, dropping POPE F1 to 79.30\% and elevating CHAIR $C_I$ to 15.7\%. By contrast, additive collaboration ($\ell_t^{\mathrm{base}} + \alpha\ell_t^{\mathrm{merged}}$) achieves the best overall results at 87.97\% POPE F1, 8.0\% CHAIR $C_I$, and 77.3\% object F1, confirming that $\ell_t^{\mathrm{merged}}$ acts strictly as a positive visual collaborator.

\begin{figure*}[t]
\centering
\includegraphics[width=0.98\textwidth]{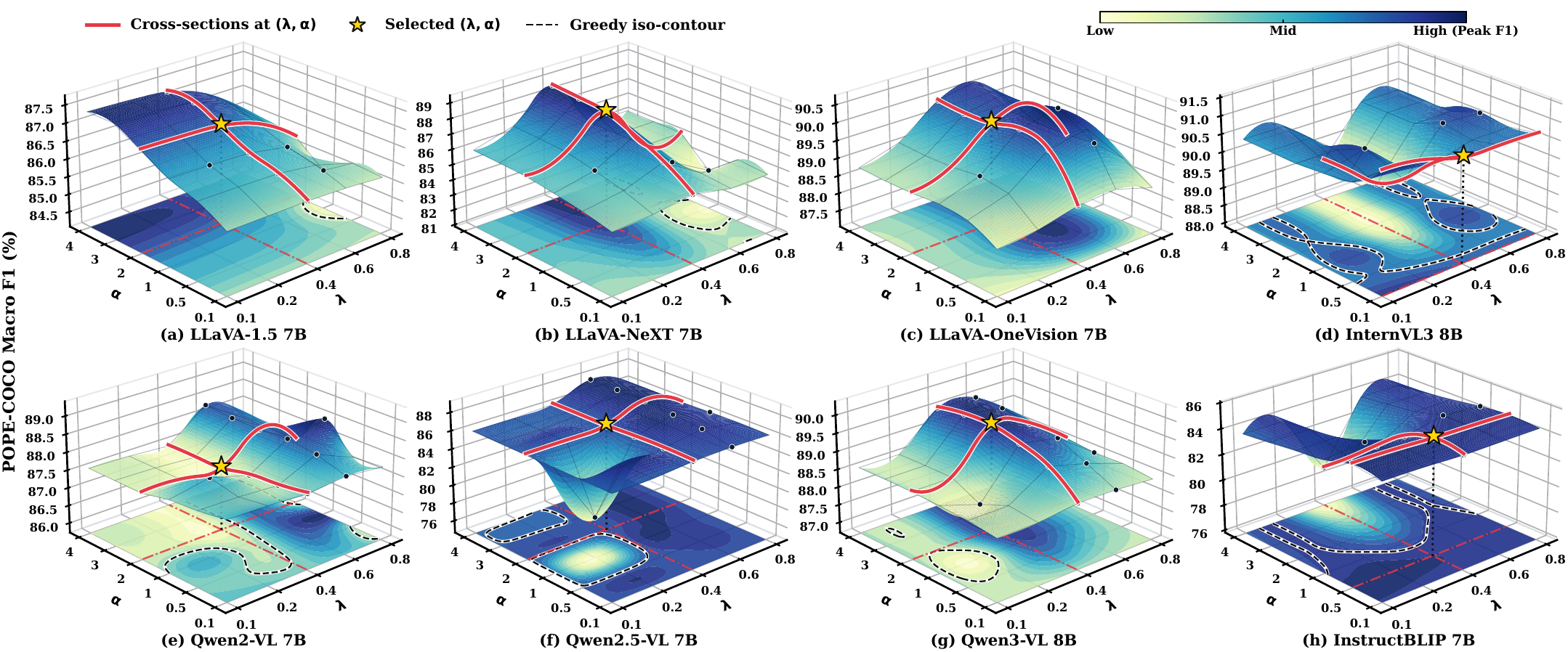}
\caption{Joint sensitivity to the retained anchor ratio $\lambda$ and collaboration weight $\alpha$ on POPE. Each 3D surface and floor projection maps three-split mean F1 (\%) over $\lambda\in\{0.1,0.2,0.4,0.6,0.8\}$ and $\alpha\in\{0.1,0.5,1,2,3,4\}$ across the LLaVA family~(a--c), InternVL3~(d), the Qwen-VL family~(e--g), and InstructBLIP~(h). Dashed floor contours mark Greedy, crimson curves trace orthogonal cross-sections along $\lambda$ and $\alpha$, and gold stars mark the selected IKOD settings $(\lambda{=}0.4,\alpha)$.}
\label{fig:hyperparam}
\end{figure*}

\noindent\textbf{Sensitivity to $\lambda$ and $\alpha$.} 
Figure~\ref{fig:hyperparam} maps the joint POPE-COCO F1 terrain over $(\lambda,\alpha)$ across all eight backbones, with corresponding CHAIR trajectories in the Supplementary Material. Six of the eight models peak at a balanced ratio near $\lambda=0.4$, bounded between coarse-bucket collapse at $\lambda\le 0.2$ and unmitigated dilution near the Greedy contour at $\lambda=0.8$. While Qwen2-VL and Qwen2.5-VL slightly favor $\lambda=0.6$ on single-token POPE where $\mathcal{P}_1$ holds only the short question prompt under 3D M-RoPE, multi-sentence CHAIR generation requires $\lambda=0.4$ to compress accumulated history and halve Qwen2-VL $C_I$, supporting a universal $\lambda=0.4$. Along $\lambda=0.4$, optimal $\alpha$ tracks connector capacity: ViT-MLP and M-RoPE decoders in LLaVA and Qwen-VL plateau near $\alpha=2.0$, InternVL3 peaks at $\alpha=0.1$, and InstructBLIP at $\alpha=0.5$ before over-fusion at $\alpha\ge 3.0$, motivating IKOD-A.

\subsection{Further Analysis}
\label{sec:discussion}
\label{sec:sampling}

\begin{table}[!htb]
\centering
\caption{Zero-shot cross-scale transfer on CHAIR across four 13B--32B backbones. IKOD and IKOD-A directly apply the 7B/8B settings on 500 MSCOCO validation images. Bold and underline mark the best and second-best results per block.}
\label{tab:scale13b}
\small
\setlength{\tabcolsep}{5.876pt}
\begin{tabular}{ll cccc}
\toprule
Model & Method & $C_S\downarrow$ & $C_I\downarrow$ & Recall$\uparrow$ & F1$\uparrow$ \\
\midrule
\multirow{3}{*}{LLaVA-1.5-13B} & Greedy & 50.4 & 15.3 & \underline{78.4} & 81.4 \\
 & \textbf{IKOD} & \textbf{38.8} & \textbf{10.0} & 77.0 & \textbf{83.0} \\
 & \textbf{IKOD-A} & \underline{44.4} & \underline{12.9} & \textbf{78.8} & \underline{82.7} \\
\midrule
\multirow{3}{*}{InstructBLIP-13B} & Greedy & \underline{48.4} & 14.4 & \textbf{68.5} & 76.1 \\
 & \textbf{IKOD} & \textbf{42.2} & \underline{11.9} & 67.7 & \underline{76.6} \\
 & \textbf{IKOD-A} & \textbf{42.2} & \textbf{11.8} & \underline{68.4} & \textbf{77.0} \\
\midrule
\multirow{3}{*}{InternVL3-14B} & Greedy & \textbf{32.4} & 9.7 & 70.8 & 79.4 \\
 & \textbf{IKOD} & \textbf{32.4} & \underline{9.3} & \underline{71.1} & \underline{79.7} \\
 & \textbf{IKOD-A} & \underline{33.4} & \textbf{9.0} & \textbf{71.5} & \textbf{80.1} \\
\midrule
\multirow{3}{*}{Qwen2.5-VL-32B} & Greedy & 54.0 & 9.6 & \textbf{74.0} & 81.4 \\
 & \textbf{IKOD} & \underline{50.2} & \textbf{8.3} & 73.3 & \underline{81.5} \\
 & \textbf{IKOD-A} & \textbf{48.2} & \underline{8.7} & \underline{73.9} & \textbf{81.7} \\
\midrule
\multirow{3}{*}{\textbf{Average}} & Greedy & 46.3 & 12.3 & \underline{72.9} & 79.6 \\
 & \textbf{IKOD} & \textbf{40.9} & \textbf{9.9} & 72.3 & \underline{80.2} \\
 & \textbf{IKOD-A} & \underline{42.1} & \underline{10.6} & \textbf{73.2} & \textbf{80.4} \\
\bottomrule
\end{tabular}
\vspace{2pt}
\end{table}

\begin{table}[!htb]
\centering
\caption{Robustness across stochastic sampling strategies and temperatures on POPE. Eight-LVLM three-split mean metrics (\%). Bold and underline mark the best and second-best results per block.}
\label{tab:chairsamp}
\small
\setlength{\tabcolsep}{3.078pt}
\begin{tabular}{llcccc}
\toprule
Sampling Strategy & Method & Acc.$\uparrow$ & Prec.$\uparrow$ & Recall$\uparrow$ & F1$\uparrow$ \\
\midrule
\multirow{3}{*}{Top-$P$ ($p{=}0.9, T{=}1.0$)} & Base & 86.46 & \textbf{91.90} & 80.43 & 85.59 \\
 & \textbf{IKOD} & \textbf{87.40} & 89.97 & \textbf{84.90} & \textbf{87.15} \\
 & \textbf{IKOD-A} & \underline{86.77} & \underline{90.53} & \underline{82.86} & \underline{86.28} \\
\midrule
\multirow{3}{*}{Top-$K$ ($k{=}50, T{=}1.0$)} & Base & 85.48 & \textbf{90.61} & 79.62 & 84.56 \\
 & \textbf{IKOD} & \textbf{87.21} & 89.85 & \textbf{84.59} & \textbf{86.94} \\
 & \textbf{IKOD-A} & \underline{86.54} & \underline{90.40} & \underline{82.51} & \underline{86.03} \\
\midrule
\multirow{3}{*}{\begin{tabular}[c]{@{}l@{}}Multinomial\\($p{=}1.0, T{=}1.0$)\end{tabular}} & Base & 85.23 & \textbf{90.29} & 79.51 & 84.35 \\
 & \textbf{IKOD} & \textbf{87.20} & 89.83 & \textbf{84.58} & \textbf{86.92} \\
 & \textbf{IKOD-A} & \underline{86.51} & \underline{90.27} & \underline{82.66} & \underline{86.02} \\
\midrule
\multirow{3}{*}{Top-$K$ ($k{=}50, T{=}0.7$)} & Base & 86.60 & \textbf{92.09} & 80.49 & 85.72 \\
 & \textbf{IKOD} & \textbf{87.63} & 90.28 & \textbf{84.96} & \textbf{87.35} \\
 & \textbf{IKOD-A} & \underline{87.17} & \underline{90.96} & \underline{83.21} & \underline{86.68} \\
\midrule
\multirow{3}{*}{Top-$K$ ($k{=}50, T{=}1.5$)} & Base & 82.62 & 86.91 & 77.36 & 81.65 \\
 & \textbf{IKOD} & \textbf{86.60} & \underline{89.34} & \textbf{83.89} & \textbf{86.31} \\
 & \textbf{IKOD-A} & \underline{85.79} & \textbf{89.60} & \underline{81.80} & \underline{85.27} \\
\bottomrule
\end{tabular}
\vspace{2pt}
\end{table}

\noindent\textbf{Transfer to larger backbones.} 
Table~\ref{tab:scale13b} evaluates zero-shot transfer of the 7B/8B settings to 13B--32B backbones on CHAIR across all four families, with corresponding 13B--32B POPE results reported in the Supplementary Material. Without hyperparameter retuning, IKOD and IKOD-A achieve the largest reductions on LLaVA-1.5-13B and InstructBLIP-13B, where Greedy $C_I$ remains above 13\%. Specifically, IKOD cuts $C_S$ from 47.0\% to 37.8\% and $C_I$ from 13.7\% to 10.3\% on LLaVA-1.5-13B and lowers $C_I$ from 16.4\% to 12.4\% on InstructBLIP-13B. Even on InternVL3-14B and Qwen2.5-VL-32B where Greedy $C_I$ is already below 10\%, both variants further reduce hallucination and improve grounding, lowering four-family mean $C_I$ from 12.3\% to 9.9\% and 10.6\% while lifting object F1 from 79.6\% to 80.2\% and 80.4\%.

\noindent\textbf{Sampling-strategy and random-seed robustness.} 
To verify that our gains extend beyond greedy decoding, Table~\ref{tab:chairsamp} evaluates IKOD and IKOD-A across five stochastic sampling strategies and temperatures over all eight backbones, with per-backbone breakdowns deferred to the Supplementary Material. IKOD and IKOD-A rank first and second in average POPE accuracy, recall, and F1 under every sampler, with the largest gains of 4.66 and 3.62 F1 points emerging under high-entropy $T{=}1.5$ decoding where language-prior drift is most severe. Across three independent Top-$P$ random seeds on CHAIR in Table~\ref{tab:seed_rob}, IKOD and IKOD-A likewise reduce eight-backbone mean $C_I$ from 12.1\% to 9.0\% and 9.7\% and cut $C_S$ from 41.7\% to 34.1\% and 38.1\%, while raising object F1 from 75.8\% to 77.2\% and 77.7\%.

\begin{table}[!htb]
\centering
\caption{Protocol robustness on CHAIR across eight LVLMs under alternative prompt phrasings (512-token cap), a strictly length-matched 128-token budget, and Top-$P$ ($p{=}0.9, T{=}1.0$) sampling across three random seeds (42, 1234, 2026). Bold and underline mark the best and second-best results per block.}
\label{tab:prompt_len}
\label{tab:short_budget}
\label{tab:seed_rob}
\small
\setlength{\tabcolsep}{2.2pt}
\resizebox{\linewidth}{!}{\begin{tabular}{llccccc}
\toprule
Prompt / Setting & Method & $C_S\downarrow$ & $C_I\downarrow$ & Recall$\uparrow$ & F1$\uparrow$ & Words \\
\midrule
\multirow{3}{*}{\textit{Describe this image.}} & Greedy & 38.3 & 9.7 & \textbf{69.6} & \textbf{78.4} & 191.3 \\
 & \textbf{IKOD} & \textbf{32.5} & \textbf{8.0} & 67.2 & \underline{77.4} & 204.9 \\
 & \textbf{IKOD-A} & \underline{34.7} & \underline{8.3} & \underline{68.8} & \textbf{78.4} & 197.0 \\
\midrule
\multirow{3}{*}{\begin{tabular}[c]{@{}l@{}}\textit{Please provide a}\\\textit{detailed description...}\end{tabular}} & Greedy & 37.0 & 10.2 & \textbf{68.2} & \textbf{77.3} & 154.0 \\
 & \textbf{IKOD} & \textbf{31.1} & \textbf{7.8} & 66.2 & 76.7 & 176.1 \\
 & \textbf{IKOD-A} & \underline{34.9} & \underline{9.2} & \underline{67.5} & \underline{77.2} & 162.5 \\
\midrule
\multirow{3}{*}{\begin{tabular}[c]{@{}l@{}}\textit{What is showing}\\\textit{in this image?}\end{tabular}} & Greedy & 17.8 & 6.1 & \underline{56.2} & 68.9 & 64.3 \\
 & \textbf{IKOD} & \underline{16.2} & \underline{5.3} & 56.0 & \underline{69.2} & 67.3 \\
 & \textbf{IKOD-A} & \textbf{15.7} & \textbf{5.2} & \textbf{56.4} & \textbf{69.5} & 66.1 \\
\midrule
\multirow{3}{*}{128-token budget} & Greedy & 30.4 & 9.8 & \underline{61.8} & 72.8 & 91.7 \\
 & \textbf{IKOD} & \textbf{26.6} & \textbf{7.7} & 61.4 & \underline{73.2} & 93.1 \\
 & \textbf{IKOD-A} & \underline{28.5} & \underline{8.5} & \textbf{62.2} & \textbf{73.5} & 92.8 \\
\midrule
\multirow{3}{*}{\begin{tabular}[c]{@{}l@{}}Top-$P$ ($p{=}0.9, T{=}1.0$)\\(3 Seeds, Mean$\pm$Std)\end{tabular}} & Base & 41.7$\pm$0.8 & 12.1$\pm$0.1 & 66.9$\pm$0.6 & 75.8$\pm$0.4 & 154.0 \\
 & \textbf{IKOD} & \textbf{34.1}$\pm$0.3 & \textbf{9.0}$\pm$0.5 & \underline{67.4}$\pm$0.1 & \underline{77.2}$\pm$0.2 & 165.4 \\
 & \textbf{IKOD-A} & \underline{38.1}$\pm$1.1 & \underline{9.7}$\pm$0.2 & \textbf{68.6}$\pm$0.1 & \textbf{77.7}$\pm$0.1 & 154.5 \\
\bottomrule
\end{tabular}}
\end{table}

\noindent\textbf{Prompt-phrasing and generation-budget robustness.} 
To verify that IKOD generalizes across prompt formulations and shorter generation budgets, Table~\ref{tab:prompt_len} evaluates three alternative CHAIR instructions alongside a length-matched 128-token budget across all eight LVLMs. Across concise imperatives, polite detailed requests, and brief queries averaging 191.3, 154.0, and 64.3 words under Greedy, both IKOD and IKOD-A consistently cut $C_S$ by up to 5.9 points and $C_I$ by up to 2.4 points without shortening responses, while IKOD-A preserves or improves mean object F1 across all three length regimes, reaching 69.5\% versus 68.9\% on brief queries. Under the 128-token budget at 91.7--93.1 words, within 1.4 words of Greedy, IKOD and IKOD-A lower $C_I$ on all eight backbones, cutting mean $C_I$ from 9.8\% to 7.7\% and 8.5\% and lifting F1 to 73.2\% and 73.5\%.

\begin{figure*}[!t]
 \centering
 \includegraphics[width=\textwidth]{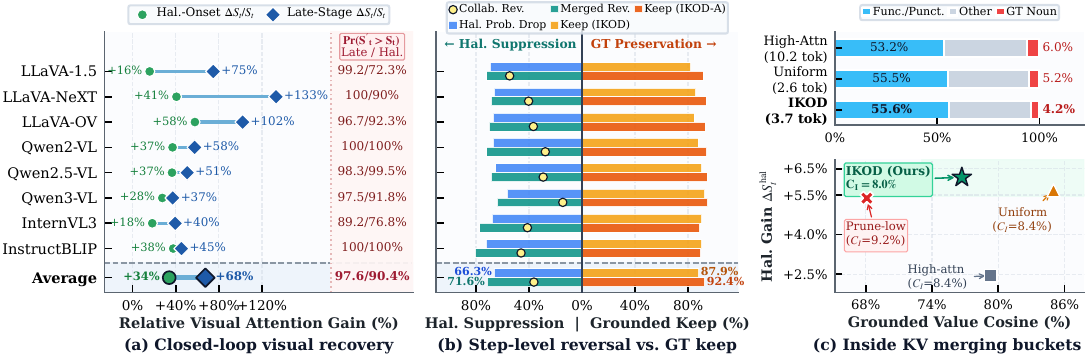}
 \caption{Dataset-wide closed-loop mechanistic evaluation across all eight LVLMs on CHAIR ($\lambda{=}0.4$). (a) Horizontal dumbbell trajectories of relative visual-attention gain from hallucination-onset steps to late generation ($t \ge T/2$), paired with the empirical satisfaction rate $\Pr(S'_t > S_t)$ of Theorem~\ref{thm:share} on the right. (b) Diverging step-level comparison between hallucination suppression on the left (hallucinated-noun probability drop, merged-branch top-1 reversal, and single-step collaborative reversal) and grounded-object preservation on the right (IKOD and IKOD-A). (c) Internal bucket anatomy showing anchor lexical composition and late-span bucket size (top) alongside the trade-off between grounded value cosine fidelity, hallucination-onset visual gain $\Delta S_t^{\mathrm{hal}}$, and CHAIR $C_I$ across merging and pruning rules (bottom).}
 \label{fig:mech8b}
\end{figure*}

\begin{figure*}[!t]
 \centering
 \includegraphics[width=\textwidth]{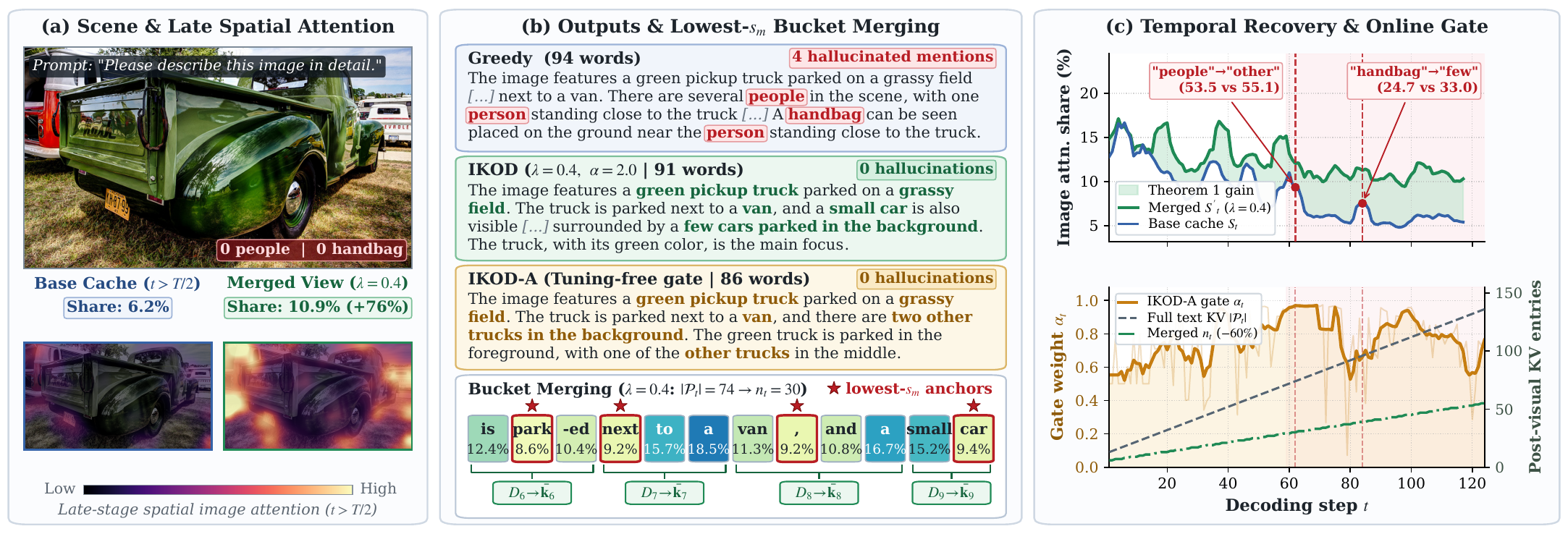}
 \caption{Mechanistic case study on LLaVA-1.5 (COCO val2014). (a) Input outdoor vehicle scene containing zero people or handbags, paired with late-stage ($t > T/2$) spatial image-attention overlays comparing the full base cache against the merged view ($\lambda{=}0.4$). (b) Generated descriptions from Greedy, IKOD, and tuning-free IKOD-A, alongside live post-visual token merging where contiguous intervals centered at the lowest-$s_m$ anchors collapse into bucket centroids $\tilde{\boldsymbol{k}}_b$. (c) Per-step visual attention share verifying Theorem~\ref{thm:share} ($S'_t > S_t$) with exact logit-reversal points, stacked above IKOD-A's online decay gate $\alpha_t$ and compressed post-visual KV length.}
 \label{fig:attnvizmain}
\end{figure*}

\noindent\textbf{Inference efficiency and effectiveness trade-off.} 
Table~\ref{tab:efficiency} jointly profiles 512-token decoding cost and eight-backbone grounding performance. Because the merged branch shares a single vision-encoder and prompt prefill, reuses the cached visual prefix $\mathcal{I}$ across layers, and queries only the current token against $(\tilde{K}_t,\tilde{V}_t)$ as detailed in Section~\ref{sec:impl}, IKOD and IKOD-A require only 27.6 and 27.8\,ms/token, corresponding to 1.14$\times$ and 1.15$\times$ Greedy, 40--44\% faster than two-pass contrastive baselines, and 9.2$\times$ faster than OPERA at 1.12$\times$ peak memory versus 1.74$\times$ for ReVisiT. Crucially, whereas every baseline either lowers object F1 below Greedy or elevates $C_I$, IKOD and IKOD-A are the only methods that simultaneously improve POPE F1, CHAIR $C_I$, and object F1.

\begin{table}[!t]
\centering
\caption{Inference efficiency and effectiveness trade-off across eight LVLMs. Eight-LVLM mean 512-token captioning cost alongside mean POPE F1, CHAIR $C_I$, and object F1 (\%). $^{\ast}$: five-beam search.}
\label{tab:efficiency}
\small
\setlength{\tabcolsep}{4.255pt}
\begin{tabular}{l cccccc}
\toprule
Method & ms/tok$\downarrow$ & Time$\times\downarrow$ & Mem$\times\downarrow$ & POPE$\uparrow$ & $C_I\downarrow$ & F1$\uparrow$ \\
\midrule
Greedy & 24.3 & 1.00 & 1.00 & 86.95 & 10.1 & 77.2 \\
VCD & 47.7 & 1.97 & 1.03 & 87.60 & 11.9 & 75.9 \\
M3ID & 46.8 & 1.92 & 1.00 & 86.96 & 10.9 & \underline{77.5} \\
OPERA$^{\ast}$ & 253.6 & 10.67 & 1.17 & 87.14 & 9.6 & 75.3 \\
ICD & 48.0 & 1.98 & 1.04 & 86.03 & 11.9 & 75.5 \\
SID & 49.6 & 2.04 & 1.03 & 87.03 & 11.5 & 76.7 \\
AGLA & 49.7 & 2.05 & 1.18 & 87.08 & 9.9 & 76.6 \\
ReVisiT & 25.5 & 1.05 & 1.74 & 85.48 & 10.2 & 76.9 \\
\midrule
\textbf{IKOD} & 27.6 & 1.14 & 1.12 & \textbf{87.97} & \textbf{8.0} & 77.3 \\
\textbf{IKOD-A} & 27.8 & 1.15 & 1.12 & \underline{87.62} & \underline{9.2} & \textbf{77.6} \\
\bottomrule
\end{tabular}
\end{table}

\noindent\textbf{Mechanistic verification and case study.} 
Because Theorem~\ref{thm:share} is a step-wise property conditioned on a shared prefix $y_{<t}$, Figure~\ref{fig:mech8b} evaluates $\mathcal{C}_t$ and $\tilde{\mathcal{C}}_t$ at $\lambda{=}0.4$ on identical prefixes across 3,419 grounded-object steps, 1,694 hallucination-onset steps, and 801,118 total CHAIR decoding steps over all eight LVLMs. Across every backbone, base visual attention is lower on hallucinated nouns than on grounded nouns, dropping from an eight-model mean of 20.0\% to 17.8\%, while 58.0\% of hallucinations occur in the second half of generation ($t \ge T/2$) as base attention decays from 17.9\% early to 13.8\% late. In Figure~\ref{fig:mech8b}(a), Theorem~\ref{thm:share} ($S'_t > S_t$) holds on 97.6\% of late decoding steps and 90.4\% of hallucination-onset steps, lifting mean late visual attention by 67.5\% relative from 13.8\% to 21.5\% and restoring attention at hallucination-onset steps by 34.0\% relative from 17.8\% to 24.0\%, which drives IKOD-A's online gate $\alpha_t$ from 0.60 early to 0.78 late and 0.94 at the 90th percentile. At hallucination-onset steps in Figure~\ref{fig:mech8b}(b), the merged branch cuts hallucinated-noun probability by 66.3\% relative and reverses 71.6\% of hallucinated top-1 tokens on the merged branch and 36.3\% on the single step alone under collaborative fusion, while preserving 87.9\% and 92.4\% of grounded predictions under IKOD and IKOD-A. Figure~\ref{fig:attnvizmain} illustrates this on LLaVA-1.5: at $\sim$90 words, Greedy's late-stage ($t > T/2$) image attention drops to 6.2\% and induces four co-occurrence hallucinations (\textit{people}, \textit{person}, and \textit{handbag}), whereas folding post-visual intervals around low-$s_m$ function-word anchors at $\lambda{=}0.4$ cuts post-visual KV length by 60\%, restores late spatial attention to 10.9\%, and reverses hallucinatory logits (\texttt{"people"}$\to$\texttt{"other"} and \texttt{"handbag"}$\to$\texttt{"few"}) as $\alpha_t$ ramps above $0.9$ past $t{=}60$.

\section{Conclusion}
\label{sec:conclusion}
In this paper, we showed across eight LVLMs that visual attention decays steadily as generated text accumulates, closely tracking object hallucination. To counter this dilution, we proposed IKOD, a training-free collaborative decoder that restores visual focus via image-attention-guided key-value merging. By selecting low-image-attention text positions as anchors and averaging surrounding post-visual text KV entries within contiguous buckets, IKOD forms a compact merged branch that provably increases visual attention share and fuses its logits with the base branch under a plausibility constraint. Our adaptive variant, IKOD-A, further schedules the collaboration weight online from the measured attention decay under a single universal configuration. Extensive experiments on hallucination and general multimodal benchmarks across eight LVLMs demonstrate that IKOD and IKOD-A consistently mitigate object hallucination while preserving object recall and general multimodal capabilities at minimal inference cost.

\textbf{Limitations and Future Work.}
First, IKOD uses a uniform anchor ratio $\lambda$ and layer-averaged decay to remain tuning-free; adapting $\lambda$ and $\alpha_t$ to layer- or head-specific visual sensitivity could offer finer control across connectors. Second, while reusing the visual prefix and querying only the compressed post-visual cache limit 512-token overhead to 1.14--1.15$\times$ Greedy, multi-step centroid reuse could push latency even closer to 1.0$\times$. Finally, extending attention-guided bucket merging to interleaved multi-image dialogues and streaming video LVLMs is a promising direction.

\bibliographystyle{IEEEtran}
\bibliography{refs}

\begin{thebibliography}{10}
\providecommand{\url}[1]{#1}
\csname url@samestyle\endcsname
\providecommand{\newblock}{\relax}
\providecommand{\bibinfo}[2]{#2}
\providecommand{\BIBentrySTDinterwordspacing}{\spaceskip=0pt\relax}
\providecommand{\BIBentryALTinterwordstretchfactor}{4}
\providecommand{\BIBentryALTinterwordspacing}{\spaceskip=\fontdimen2\font plus
\BIBentryALTinterwordstretchfactor\fontdimen3\font minus \fontdimen4\font\relax}
\providecommand{\BIBforeignlanguage}[2]{{%
\expandafter\ifx\csname l@#1\endcsname\relax
\typeout{** WARNING: IEEEtran.bst: No hyphenation pattern has been}%
\typeout{** loaded for the language `#1'. Using the pattern for}%
\typeout{** the default language instead.}%
\else
\language=\csname l@#1\endcsname
\fi
#2}}
\providecommand{\BIBdecl}{\relax}
\BIBdecl

\bibitem{li2023evaluating}
Y.~Li, Y.~Du, K.~Zhou, J.~Wang, W.~X. Zhao, and J.-R. Wen, ``Evaluating object hallucination in large vision-language models,'' in \emph{Proceedings of the 2023 Conference on Empirical Methods in Natural Language Processing (EMNLP)}, 2023, pp. 292--305.

\bibitem{rohrbach2018object}
A.~Rohrbach, L.~A. Hendricks, K.~Burns, T.~Darrell, and K.~Saenko, ``Object hallucination in image captioning,'' in \emph{Proceedings of the 2018 Conference on Empirical Methods in Natural Language Processing (EMNLP)}, 2018, pp. 4035--4045.

\bibitem{chen2024detecting}
J.~Chen \emph{et~al.}, ``Detecting and evaluating medical hallucinations in large vision language models,'' \emph{arXiv preprint arXiv:2406.10185}, 2024.

\bibitem{kang2023deficiency}
H.~Kang and X.-Y. Liu, ``Deficiency of large language models in finance: An empirical examination of hallucination,'' in \emph{I Can't Believe It's Not Better Workshop: Failure Modes in the Age of Foundation Models}, 2023.

\bibitem{ding2024open6dor}
Y.~Ding \emph{et~al.}, ``{Open6DOR}: Benchmarking open-instruction {6-DoF} object rearrangement and a {VLM}-based approach,'' in \emph{2024 IEEE/RSJ International Conference on Intelligent Robots and Systems (IROS)}.\hskip 1em plus 0.5em minus 0.4em\relax IEEE, 2024, pp. 7359--7366.

\bibitem{vcd}
S.~Leng \emph{et~al.}, ``Mitigating object hallucinations in large vision-language models through visual contrastive decoding,'' in \emph{Proceedings of the IEEE/CVF Conference on Computer Vision and Pattern Recognition}, 2024, pp. 13\,872--13\,882.

\bibitem{huang2024opera}
Q.~Huang \emph{et~al.}, ``{OPERA}: Alleviating hallucination in multi-modal large language models via over-trust penalty and retrospection-allocation,'' in \emph{Proceedings of the IEEE/CVF Conference on Computer Vision and Pattern Recognition}, 2024, pp. 13\,418--13\,427.

\bibitem{agla}
W.~An \emph{et~al.}, ``Mitigating object hallucinations in large vision-language models with assembly of global and local attention,'' in \emph{Proceedings of the IEEE/CVF Conference on Computer Vision and Pattern Recognition}, 2025, pp. 29\,915--29\,926.

\bibitem{liu2023aligning}
F.~Liu, K.~Lin, L.~Li, J.~Wang, Y.~Yacoob, and L.~Wang, ``Mitigating hallucination in large multi-modal models via robust instruction tuning,'' in \emph{International Conference on Learning Representations}, vol. 2024, 2024, pp. 57\,689--57\,733.

\bibitem{zhao2023svit}
B.~Zhao, B.~Wu, M.~He, and T.~Huang, ``{SVIT}: Scaling up visual instruction tuning,'' \emph{arXiv preprint arXiv:2307.04087}, 2023.

\bibitem{rlhfv}
T.~Yu \emph{et~al.}, ``{RLHF-V}: Towards trustworthy {MLLMs} via behavior alignment from fine-grained correctional human feedback,'' in \emph{Proceedings of the IEEE/CVF Conference on Computer Vision and Pattern Recognition}, 2024, pp. 13\,807--13\,816.

\bibitem{zhou2023analyzing}
Y.~Zhou \emph{et~al.}, ``Analyzing and mitigating object hallucination in large vision-language models,'' in \emph{International Conference on Learning Representations}, vol. 2024, 2024, pp. 56\,969--56\,998.

\bibitem{yin2024woodpecker}
S.~Yin \emph{et~al.}, ``Woodpecker: Hallucination correction for multimodal large language models,'' \emph{Science China Information Sciences}, vol.~67, no.~12, p. 220105, 2024.

\bibitem{zhang2024debiasing}
Y.-F. Zhang \emph{et~al.}, ``Debiasing multimodal large language models via penalization of language priors,'' in \emph{Proceedings of the 33rd ACM International Conference on Multimedia}, 2025, pp. 4232--4241.

\bibitem{revisit2025}
B.~Cho and J.~Kim, ``Revisit what you see: Revealing visual semantics in vision tokens to guide {LVLM} decoding,'' in \emph{Proceedings of the 64th Annual Meeting of the Association for Computational Linguistics (Volume 1: Long Papers)}, 2026, pp. 5794--5824.

\bibitem{favero2024multi}
A.~Favero \emph{et~al.}, ``Multi-modal hallucination control by visual information grounding,'' in \emph{Proceedings of the IEEE/CVF Conference on Computer Vision and Pattern Recognition}, 2024, pp. 14\,303--14\,312.

\bibitem{vista2025}
Z.~Li \emph{et~al.}, ``The hidden life of tokens: Reducing hallucination of large vision-language models via visual information steering,'' in \emph{International Conference on Machine Learning}.\hskip 1em plus 0.5em minus 0.4em\relax PMLR, 2025, pp. 35\,799--35\,819.

\bibitem{tarac2025}
L.~Jiang \emph{et~al.}, ``{TARAC}: Mitigating hallucination in {LVLMs} via temporal attention real-time accumulative connection,'' \emph{arXiv preprint arXiv:2504.04099}, 2025.

\bibitem{rudder2026}
Z.~Zou, Y.~Gao, J.~Guan, B.~Li, and P.~Marttinen, ``Adaptive residual-update steering for low-overhead hallucination mitigation in large vision-language models,'' in \emph{International Conference on Machine Learning}.\hskip 1em plus 0.5em minus 0.4em\relax PMLR, 2026, pp. 168\,675--168\,694.

\bibitem{afip2026}
Q.~Li, Z.~Liu, W.~Luo, T.~Luo, and C.~Hou, ``Correcting visual blur induced by attention distraction to reduce hallucinations: Algorithm and theory,'' in \emph{International Conference on Machine Learning}.\hskip 1em plus 0.5em minus 0.4em\relax PMLR, 2026, pp. 68\,534--68\,569.

\bibitem{sumgd2025}
K.~Min, M.~Kim, K.-i. Lee, D.~Lee, and K.~Jung, ``Mitigating hallucinations in large vision-language models via summary-guided decoding,'' in \emph{Findings of the Association for Computational Linguistics: NAACL 2025}, 2025, pp. 4183--4198.

\bibitem{halback2026}
B.~Li, H.~Yu, Y.~Han, J.~Ren, M.~Dai, and S.~Bao, ``Mitigating multimodal hallucinations through visual attention tracing and origin-point regeneration,'' \emph{Scientific Reports}, vol.~16, no.~1, p. 25596, 2026.

\bibitem{vgd2026}
L.~Yang \emph{et~al.}, ``Hallucinations leave a grounding signature: Verifier-guided decoding for selective object correction,'' \emph{arXiv preprint arXiv:2607.27823}, 2026.

\bibitem{saliencydrops2026}
X.~Zhang \emph{et~al.}, ``Hallucination begins where saliency drops,'' in \emph{International Conference on Learning Representations}, 2026.

\bibitem{icd2024}
X.~Wang, J.~Pan, L.~Ding, and C.~Biemann, ``Mitigating hallucinations in large vision-language models with instruction contrastive decoding,'' in \emph{Findings of the Association for Computational Linguistics: ACL 2024}, 2024, pp. 15\,840--15\,853.

\bibitem{sid2025}
F.~Huo, W.~Xu, Z.~Zhang, H.~Wang, Z.~Chen, and P.~Zhao, ``Self-introspective decoding: Alleviating hallucinations for large vision-language models,'' in \emph{International Conference on Learning Representations}, vol. 2025, 2025, pp. 24\,272--24\,295.

\bibitem{code2024}
J.~Kim, H.~Kim, Y.~Kim, and Y.~M. Ro, ``{CODE}: Contrasting self-generated description to combat hallucination in large multi-modal models,'' \emph{Advances in Neural Information Processing Systems}, vol.~37, pp. 133\,571--133\,599, 2024.

\bibitem{chuang2023dola}
Y.-S. Chuang, Y.~Xie, H.~Luo, Y.~Kim, J.~R. Glass, and P.~He, ``{DoLa}: Decoding by contrasting layers improves factuality in large language models,'' in \emph{International Conference on Learning Representations}, vol. 2024, 2024, pp. 54\,158--54\,183.

\bibitem{pai2024}
S.~Liu, K.~Zheng, and W.~Chen, ``Paying more attention to image: A training-free method for alleviating hallucination in {LVLMs},'' in \emph{Proceedings of the European Conference on Computer Vision (ECCV)}, 2024, pp. 125--140.

\bibitem{cam2024}
Y.~Zhang \emph{et~al.}, ``{CaM}: Cache merging for memory-efficient {LLM}s inference,'' in \emph{International Conference on Machine Learning}.\hskip 1em plus 0.5em minus 0.4em\relax PMLR, 2024, pp. 58\,840--58\,850.

\bibitem{meda2025}
Z.~Wan, H.~Shen, X.~Wang, C.~Liu, Z.~Mai, and M.~Zhang, ``{MEDA}: Dynamic {KV} cache allocation for efficient multimodal long-context inference,'' in \emph{Proceedings of the 2025 Conference of the Nations of the Americas Chapter of the Association for Computational Linguistics: Human Language Technologies (Volume 1: Long Papers)}, 2025, pp. 2485--2497.

\bibitem{liu2024efficient}
Z.~Liu \emph{et~al.}, ``Efficient inference of vision instruction-following models with elastic cache,'' in \emph{Proceedings of the European Conference on Computer Vision (ECCV)}, 2024, pp. 54--69.

\bibitem{vasparse2025}
X.~Zhuang, Z.~Zhu, Y.~Xie, L.~Liang, and Y.~Zou, ``{VASparse}: Towards efficient visual hallucination mitigation via visual-aware token sparsification,'' in \emph{Proceedings of the IEEE/CVF Conference on Computer Vision and Pattern Recognition}, 2025, pp. 4189--4199.

\bibitem{kvsmooth2026}
S.~Jiang, F.~Chen, X.~Zhang, and K.~He, ``{KVSmooth}: Mitigating hallucinations in multimodal large language models via key-value smoothing,'' in \emph{Proceedings of the IEEE/CVF Conference on Computer Vision and Pattern Recognition}, 2026, pp. 32\,561--32\,571.

\bibitem{lime2026}
I.~Allouche and J.~Keshet, ``Mitigating multimodal {LLM}s hallucinations via relevance propagation at inference time,'' \emph{arXiv preprint arXiv:2605.01766}, 2026.

\bibitem{itad2025}
X.~Xu \emph{et~al.}, ``Mitigating hallucinations in multi-modal large language models via image token attention-guided decoding,'' in \emph{Proceedings of the 2025 Conference of the Nations of the Americas Chapter of the Association for Computational Linguistics: Human Language Technologies (Volume 1: Long Papers)}, 2025, pp. 1571--1590.

\bibitem{mcothall2026}
J.~Ma, W.~Suo, P.~Wang, and Y.~Zhang, ``Evaluating and mitigating multimodal chain-of-thought hallucination,'' in \emph{Proceedings of the IEEE/CVF Conference on Computer Vision and Pattern Recognition}, 2026, pp. 40\,224--40\,234.

\bibitem{MSCOCO}
T.-Y. Lin \emph{et~al.}, ``Microsoft {COCO}: Common objects in context,'' in \emph{Proceedings of the European Conference on Computer Vision (ECCV)}, 2014, pp. 740--755.

\bibitem{pas2026}
N.~Hoang-Xuan, M.~Vu, M.~T. Thai, and M.~Bhattarai, ``Pay attention to the stall: Scaling {LVLM} guided decoding by signal noise optimization,'' in \emph{Proceedings of the IEEE/CVF Conference on Computer Vision and Pattern Recognition}, 2026, pp. 18\,273--18\,283.

\bibitem{cd}
X.~L. Li \emph{et~al.}, ``Contrastive decoding: Open-ended text generation as optimization,'' in \emph{Proceedings of the 61st Annual Meeting of the Association for Computational Linguistics (Volume 1: Long Papers)}, 2023, pp. 12\,286--12\,312.

\bibitem{wang2023amber}
J.~Wang \emph{et~al.}, ``{AMBER}: An llm-free multi-dimensional benchmark for mllms hallucination evaluation,'' \emph{arXiv preprint arXiv:2311.07397}, 2023.

\bibitem{LLaVA-1.5}
H.~Liu, C.~Li, Y.~Li, and Y.~J. Lee, ``Improved baselines with visual instruction tuning,'' in \emph{Proceedings of the IEEE/CVF Conference on Computer Vision and Pattern Recognition}, 2024, pp. 26\,286--26\,296.

\bibitem{llavanext2024}
H.~Liu \emph{et~al.}, ``{LLaVA-NeXT}: Improved reasoning, {OCR}, and world knowledge,'' LLaVA Project Blog, 2024.

\bibitem{llavaonevision2024}
B.~Li \emph{et~al.}, ``{LLaVA-OneVision}: Easy visual task transfer,'' \emph{Transactions on Machine Learning Research}, 2025.

\bibitem{qwen2vl2024}
P.~Wang \emph{et~al.}, ``{Qwen2-VL}: Enhancing vision-language model's perception of the world at any resolution,'' \emph{arXiv preprint arXiv:2409.12191}, 2024.

\bibitem{qwen25vl2025}
S.~Bai \emph{et~al.}, ``{Qwen2.5-VL} technical report,'' \emph{arXiv preprint arXiv:2502.13923}, 2025.

\bibitem{qwen3vl2025}
S.~Bai \emph{et~al.}, ``{Qwen3-VL} technical report,'' \emph{arXiv preprint arXiv:2511.21631}, 2025.

\bibitem{internvl3_2025}
J.~Zhu \emph{et~al.}, ``{InternVL3}: Exploring advanced training and test-time recipes for open-source multimodal models,'' \emph{arXiv preprint arXiv:2504.10479}, 2025.

\bibitem{InstructBLIP}
W.~Dai \emph{et~al.}, ``{InstructBLIP}: Towards general-purpose vision-language models with instruction tuning,'' \emph{Advances in Neural Information Processing Systems}, vol.~36, pp. 49\,250--49\,267, 2023.

\bibitem{MME}
C.~Fu \emph{et~al.}, ``{MME}: A comprehensive evaluation benchmark for multimodal large language models,'' \emph{Advances in Neural Information Processing Systems}, vol.~38, pp. 162\,549--162\,567, 2025.

\bibitem{mmstar2024}
L.~Chen \emph{et~al.}, ``Are we on the right way for evaluating large vision-language models?'' \emph{Advances in Neural Information Processing Systems}, vol.~37, pp. 27\,056--27\,087, 2024.

\bibitem{mmbench2024}
Y.~Liu \emph{et~al.}, ``{MMBench}: Is your multi-modal model an all-around player?'' in \emph{Proceedings of the European Conference on Computer Vision (ECCV)}, 2024, pp. 216--233.

\bibitem{scienceqa2022}
P.~Lu \emph{et~al.}, ``Learn to explain: Multimodal reasoning via thought chains for science question answering,'' \emph{Advances in Neural Information Processing Systems}, vol.~35, pp. 2507--2521, 2022.

\bibitem{Xiong_2026_CVPR}
M.~Xiong \emph{et~al.}, ``{Prune2Drive}: A plug-and-play framework for accelerating vision-language models in autonomous driving,'' in \emph{Proceedings of the IEEE/CVF Conference on Computer Vision and Pattern Recognition}, 2026, pp. 25\,215--25\,224.

\end{thebibliography}

\clearpage
\raggedbottom
\makeatletter
\setlength{\@dblfptop}{0pt}
\setlength{\@dblfpsep}{10pt}
\setlength{\@dblfpbot}{0pt plus 1fil}
\makeatother
\setlength{\abovedisplayskip}{5.5pt plus 1.5pt minus 1.5pt}
\setlength{\belowdisplayskip}{5.5pt plus 1.5pt minus 1.5pt}
\setlength{\abovedisplayshortskip}{3pt plus 1pt minus 1pt}
\setlength{\belowdisplayshortskip}{4pt plus 1pt minus 1pt}
\twocolumn[{%
\begin{center}
{\Huge Supplementary Material for ``IKOD: Image-Attention-Guided Key-Value Merging against Object Hallucination in the Decoding of Large Vision-Language Models''\par}
\vspace{1.2em}
{\large Jiabing~Yang$^{\ast}$, Chenhang~Cui$^{\ast}$, Yixiang~Chen, Tao~Yu, Yan~Huang$^{\dagger}$, and~Liang~Wang,~\IEEEmembership{Fellow,~IEEE}\par}
\vspace{1.6em}
\end{center}
}]
This supplementary material provides benchmark and baseline details, eight-backbone visual-attention analyses, the full proof of Theorem~1, hyperparameter trajectories, and per-backbone experimental breakdowns supporting the main paper. Sections, equations, figures, and tables are numbered sequentially following the main paper.

\section{Benchmarks, Baselines, and Implementation Details}

\subsection{Evaluation Metrics and Benchmarks}
\label{sec:supp_benchmarks}
\noindent\textbf{POPE.} The Polling-based Object Probing Evaluation (POPE)~\cite{li2023evaluating} formulates object hallucination evaluation as a binary polling task (``\textit{Is there a \{object\} in the image?}'') across MSCOCO~\cite{MSCOCO}, A-OKVQA, and GQA. Each dataset selects 500 images and queries six balanced Yes/No questions (three present, three absent objects; 3,000 questions per split) under three negative-sampling strategies: \textit{Random} (uniformly sampled absent objects), \textit{Popular} (most frequent absent categories in the dataset), and \textit{Adversarial} (absent objects frequently co-occurring with present objects). Following established practice~\cite{li2023evaluating,huang2024opera,revisit2025}, we evaluate on the MSCOCO validation benchmark (POPE-COCO) and report per-split and three-split average F1.

\noindent\textbf{CHAIR.} Caption Hallucination Assessment with Image Relevance (CHAIR)~\cite{rohrbach2018object} evaluates open-ended captioning on MSCOCO~\cite{MSCOCO} by matching generated nouns (via synonym lemmatization over the 80 MSCOCO categories) against ground-truth annotations and reference captions along two dimensions, instance-level ($\text{CHAIR}_I$, denoted $C_I$) and sentence-level ($\text{CHAIR}_S$, denoted $C_S$):
\begin{gather*}
\text{CHAIR}_I = \frac{|\{\text{hallucinated objects}\}|}{|\{\text{all mentioned objects}\}|}, \\
\text{CHAIR}_S = \frac{|\{\text{captions with hallucinated objects}\}|}{|\{\text{all captions}\}|}.
\end{gather*}
Following established LVLM protocols~\cite{huang2024opera,sid2025,revisit2025}, we evaluate on 500 randomly sampled MSCOCO 2014 validation images (512-token cap) and report $C_S$, $C_I$, object recall, object F1, and mean caption length (Words).

\noindent\textbf{AMBER.} The LLM-free Multi-dimensional Benchmark (AMBER)~\cite{wang2023amber} evaluates both discriminative and generative hallucination across 1,004 images annotated with exhaustive object existence, attributes, and relations. On the discriminative split (14,216 binary queries), we report fine-grained attribute F1 ($\text{Attr}\uparrow$, covering state, number, and action) and inter-object relation F1 ($\text{Rel}\uparrow$). On the generative split (1,004 open-ended descriptions under a 512-token cap), we report the hallucinated response rate ($\text{Hal}\downarrow$, percentage of captions with $\text{CHAIR}>0$) and the human-cognition-aligned hallucination rate ($\text{Cog}\downarrow$).

\noindent\textbf{MME.} The Multimodal Large Language Model Evaluation (MME)~\cite{MME} benchmark comprises 14 subtasks (10 perception and 4 cognition), pairing each image with two manually designed Yes/No instructions scored by question-level and image-level accuracy ($\text{Accuracy} + \text{Accuracy}+$, 200 points per subtask). We evaluate on all 10 perception subtasks (2,374 questions over 1,187 images, maximum score 2,000), covering coarse-grained recognition (\textit{existence}, \textit{count}, \textit{position}, \textit{color}), fine-grained recognition (\textit{poster}, \textit{celebrity}, \textit{scene}, \textit{landmark}, \textit{artwork}), and \textit{OCR}.

\noindent\textbf{MMStar.} MMStar~\cite{mmstar2024} comprises 1,500 vision-indispensable multiple-choice samples curated to eliminate unimodal knowledge shortcuts and data leakage, balanced across six core capabilities (\textit{coarse perception}, \textit{fine-grained perception}, \textit{instance reasoning}, \textit{logical reasoning}, \textit{science \& technology}, and \textit{mathematics}, 250 samples each) and 18 detailed axes.

\noindent\textbf{MMBench.} MMBench~\cite{mmbench2024} evaluates 20 fine-grained leaf multimodal abilities under perception and reasoning via multiple-choice questions. We evaluate on the English development split (MMBench-EN dev, 4,329 questions).

\noindent\textbf{ScienceQA.} ScienceQA (SQA)~\cite{scienceqa2022} is a multimodal multiple-choice science QA benchmark covering natural, social, and language sciences across 26 topics and 127 categories. We evaluate on the full test split (4,241 questions).

\setlength{\IEEEilabelindent}{0pt}
\setlength{\IEEEelabelindent}{0pt}
\subsection{Overview of Backbones and Baselines}
\label{sec:supp_baselines}
\noindent\textbf{Backbone Models.} We evaluate eight open-source LVLMs spanning four model families:
\begin{itemize}
    \item \textbf{LLaVA family:} LLaVA-1.5~7B~\cite{LLaVA-1.5} connects a CLIP-ViT-L/14-336px encoder (576 visual tokens per image) to Vicuna-v1.5-7B via a two-layer MLP projector; LLaVA-NeXT~7B~\cite{llavanext2024} extends it with AnyRes dynamic high-resolution grid patching (up to four $336\times 336$ sub-crops plus a global thumbnail, up to 2,880 visual tokens) paired with Vicuna-v1.6-7B; and LLaVA-OneVision~7B~\cite{llavaonevision2024} pairs a SigLIP-SO400M/14-384px encoder (729 tokens per crop) with Qwen2-7B via a two-layer MLP under Higher-AnyRes token packing and bilinear spatial pooling.
    \item \textbf{Qwen-VL family:} Qwen2-VL~7B~\cite{qwen2vl2024} couples a 675M-parameter 2D-RoPE ViT supporting Naive Dynamic Resolution ($2\times 2$ adjacent patch MLP merger) with Qwen2-7B under 3D Multimodal Rotary Position Embedding (M-RoPE, decomposing position IDs into temporal, height, and width components); Qwen2.5-VL~7B~\cite{qwen25vl2025} introduces a windowed-attention ViT (four full-attention layers at $\{7,15,23,31\}$) with a $2\times 2$ spatial-grouping two-layer MLP merger and Qwen2.5-7B under absolute-time M-RoPE; and Qwen3-VL~8B~\cite{qwen3vl2025} builds on Qwen3-8B with Interleaved M-RoPE across frequency bands and DeepStack multi-level ViT-to-LLM residual fusion.
    \item \textbf{InternVL3:} InternVL3~8B~\cite{internvl3_2025} is trained under Native Multimodal Pre-Training, coupling InternViT-300M-448px-V2.5 with a Qwen2.5-7B language backbone (and Qwen2.5-14B in InternVL3-14B) via a two-layer MLP with $4\times$ pixel unshuffle (compressing each $448\times 448$ tile from 1,024 patches to 256 visual tokens) and Variable Visual Position Encoding (V2PE).
    \item \textbf{InstructBLIP:} InstructBLIP~7B~\cite{InstructBLIP} connects a frozen EVA-CLIP ViT-g/14 vision encoder to a frozen Vicuna-7B language model through an instruction-aware Query Transformer (Q-Former) that takes both the task instruction and 32 learnable query tokens to extract 32 task-relevant visual tokens via cross-attention.
\end{itemize}

\noindent\textbf{Decoding Baselines.} We compare against seven representative training-free decoding methods:
\begin{itemize}
    \item \textbf{VCD}~\cite{vcd}: Visual Contrastive Decoding (VCD) runs an auxiliary forward pass on a visual input corrupted by Gaussian diffusion noise and contrasts the original and distorted output distributions within an adaptive plausibility constraint to suppress unimodal language priors and statistical co-occurrence bias.
    \item \textbf{M3ID}~\cite{favero2024multi}: Multi-Modal Mutual Information Decoding (M3ID) contrasts the image-conditioned output distribution against an image-omitted text-only pass with a time-increasing weight along generation whenever prediction confidence is low, counteracting the progressive forgetting of visual context.
    \item \textbf{OPERA}~\cite{huang2024opera}: Over-trust Penalty and Retrospection-Allocation (OPERA) operates over five-beam search, penalizing candidate logits based on column-wise local self-attention aggregation patterns to mitigate over-trust in summary tokens and rolling decoding back to reallocate candidates when attention-column overlap becomes excessive.
    \item \textbf{ICD}~\cite{icd2024}: Instruction Contrastive Decoding (ICD) prepends a negative role disturbance prefix (e.g., ``\textit{You are a confused object detector.}'') in an auxiliary pass to amplify multimodal alignment uncertainty, and subtracts the disturbed distribution from the base distribution under an adaptive plausibility constraint.
    \item \textbf{SID}~\cite{sid2025}: Self-Introspective Decoding (SID) prunes high-attention visual tokens after an early decoder layer in its auxiliary pass---retaining only the least-attended visual tokens---to induce text-informed vision-and-text association hallucinations, which are then subtracted via contrastive decoding.
    \item \textbf{AGLA}~\cite{agla}: Assembly of Global and Local Attention (AGLA) uses an external BLIP-ITM model to compute image--prompt Grad-CAM relevance maps and mask out prompt-irrelevant pixel patches, additively ensembling logits from the original global image and the local task-focused view.
    \item \textbf{ReVisiT}~\cite{revisit2025}: ReVisiT projects cached vision-token hidden states from candidate decoder layers through the language modeling head during prefill, selects at each step the visual token and layer whose distribution minimizes Jensen-Shannon divergence to the confidence-constrained output distribution, and combines their log-probabilities without a second decoder pass.
\end{itemize}

\subsection{Experimental and Implementation Settings}
\label{sec:supp_impl}
All experiments are conducted using PyTorch and Hugging Face Transformers on NVIDIA A100 (80GB) GPUs. Unless otherwise specified (see the prompt and length analyses in Section~V-D of the main paper and Section~\ref{sec:supp_prompt_len}), POPE uses the prompt \texttt{"Is there a \{object\} in the image? Please answer this question with one word."} with a maximum generation length of 12 tokens, CHAIR uses \texttt{"Please describe this image in detail."} with a maximum generation length of 512 tokens, and AMBER uses its official discriminative and 512-token generative prompts; the four general multimodal benchmarks (MME, MMStar, MMBench-EN, and ScienceQA) follow the official \texttt{lmms-eval} evaluation protocol and default task prompts.

\noindent\textbf{Hyperparameter Configuration.}
Table~\ref{tab:protocols} summarizes the hyperparameter settings of all compared baselines alongside IKOD and IKOD-A. For VCD~\cite{vcd}, M3ID~\cite{favero2024multi}, OPERA~\cite{huang2024opera}, ICD~\cite{icd2024}, SID~\cite{sid2025}, and AGLA~\cite{agla}, we directly apply the fixed hyperparameters recommended in their original papers and official repositories across all eight backbones; for ReVisiT~\cite{revisit2025}, we follow its published per-family exit-layer and relative-top configuration (\textit{last} on LLaVA and InternVL3; \textit{all} on Qwen-VL and InstructBLIP). Both IKOD and IKOD-A fix the plausibility threshold $\beta=0.1$ and the universal retained-anchor ratio $\lambda=0.4$ across all architectures and benchmarks: IKOD sets only the collaboration weight $\alpha$ by model family on POPE and transfers it unchanged to CHAIR, AMBER, and the 13B--32B backbones, whereas IKOD-A fixes $(\alpha_{\min}, \alpha_{\max}, \kappa) = (0, 1, 10)$ under a single universal configuration across all models.

\begin{table}[!t]
\centering
\caption{Hyperparameter settings of all baselines and our methods. $^{\ast}$: five-beam search.}
\label{tab:protocols}
\small
\setlength{\tabcolsep}{3.5pt}
\resizebox{\linewidth}{!}{\begin{tabular}{l >{\raggedright\arraybackslash}p{188pt}}
\toprule
Method & Hyperparameter Settings \\
\midrule
VCD~\mbox{\cite{vcd}} & $\alpha{=}1.0$, $\beta{=}0.1$; diffusion steps $T_{\mathrm{noise}}{=}999$ (POPE), $500$ (CHAIR) \\
M3ID~\mbox{\cite{favero2024multi}} & Decay $\lambda{=}0.02$, confidence gate $0.3$, $\beta{=}0$ (POPE) / $0.1$ (CHAIR) \\
OPERA$^{\ast}$~\mbox{\cite{huang2024opera}} & $N_{\mathrm{beam}}{=}5$, scale $\sigma{=}50$, threshold $r{=}15$, $N_{\mathrm{can}}{=}5$, penalty $\alpha{=}1.0$ \\
ICD~\mbox{\cite{icd2024}} & $\alpha{=}1.0$, $\beta{=}0.1$; negative role disturbance prefix \\
SID~\mbox{\cite{sid2025}} & $\alpha{=}0.5$, $\beta{=}0.1$, pruning layer $i{=}2$, visual token retention $10\%$ \\
AGLA~\mbox{\cite{agla}} & $\alpha{=}2.0$, $\beta{=}0.5$; BLIP-ITM Grad-CAM block $6$ \\
ReVisiT~\mbox{\cite{revisit2025}} & Exit \textit{last} (LLaVA, InternVL3) / \textit{all} (Qwen-VL, InstructBLIP); relative-top $\tau{=}10^{-5}$ ($10^{-6}$ on InternVL3 CHAIR) \\
\midrule
\textbf{IKOD} & $\beta{=}0.1$, universal $\lambda{=}0.4$; family $\alpha{=}2.0$ (LLaVA, Qwen-VL), $0.1$ (InternVL3), $0.5$ (InstructBLIP) \\
\textbf{IKOD-A} & $\beta{=}0.1$, universal $\lambda{=}0.4$, $(\alpha_{\min},\alpha_{\max},\kappa){=}(0,1,10)$ \\
\bottomrule
\end{tabular}}
\end{table}

\begin{figure}[!t]
    \centering
    \includegraphics[width=\linewidth]{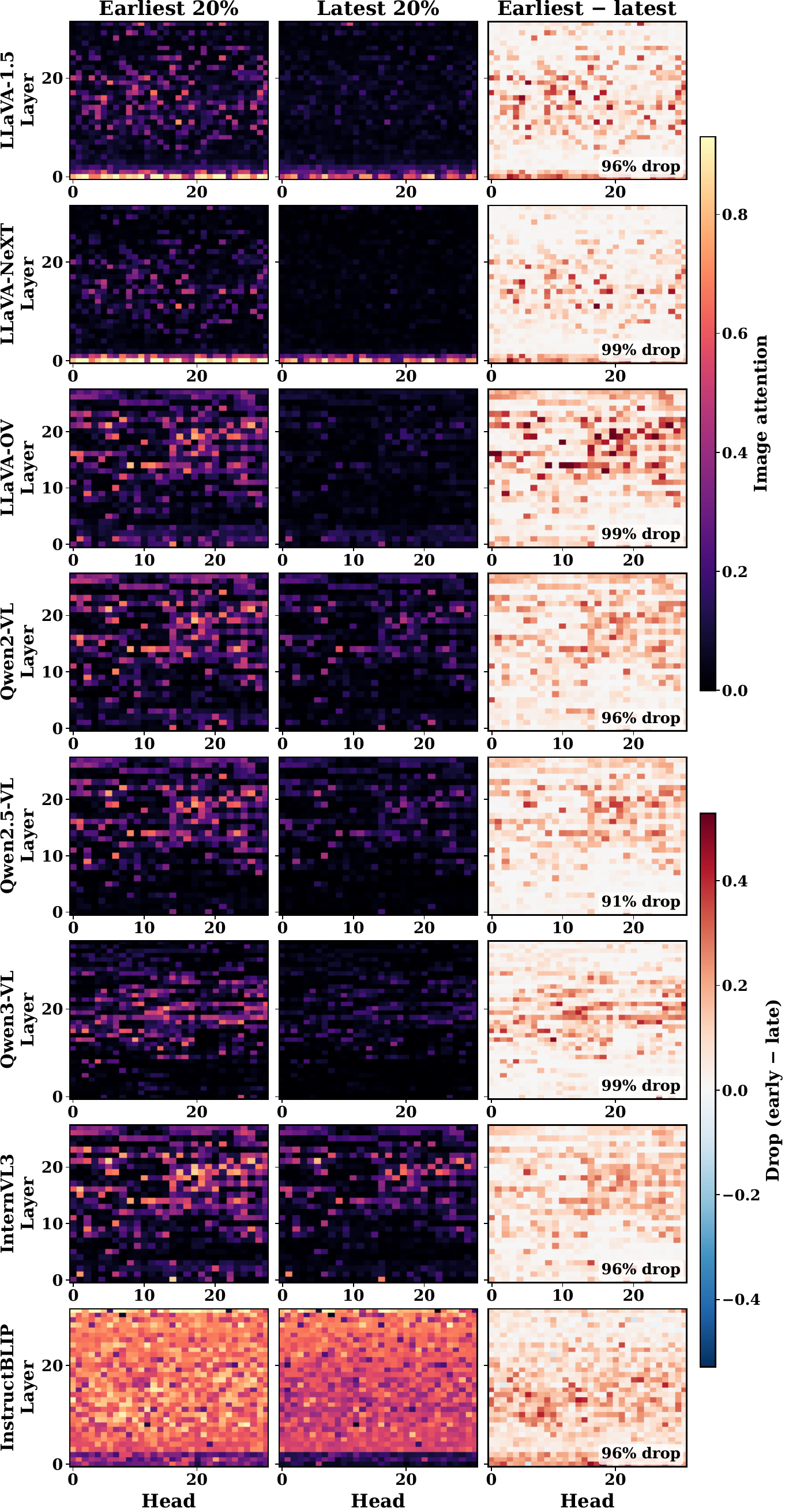}
    \caption{Per-layer and per-head image attention for the earliest 20\% (left) and latest 20\% (middle) of generated tokens across all eight backbones, alongside their late-minus-early difference (right; red indicates attention decay).}
    \label{fig:slayerhead}
\end{figure}

\section{Visual Attention Degradation Across All Eight Backbones}
\begin{figure*}[!t]
    \centering
    \includegraphics[width=\textwidth]{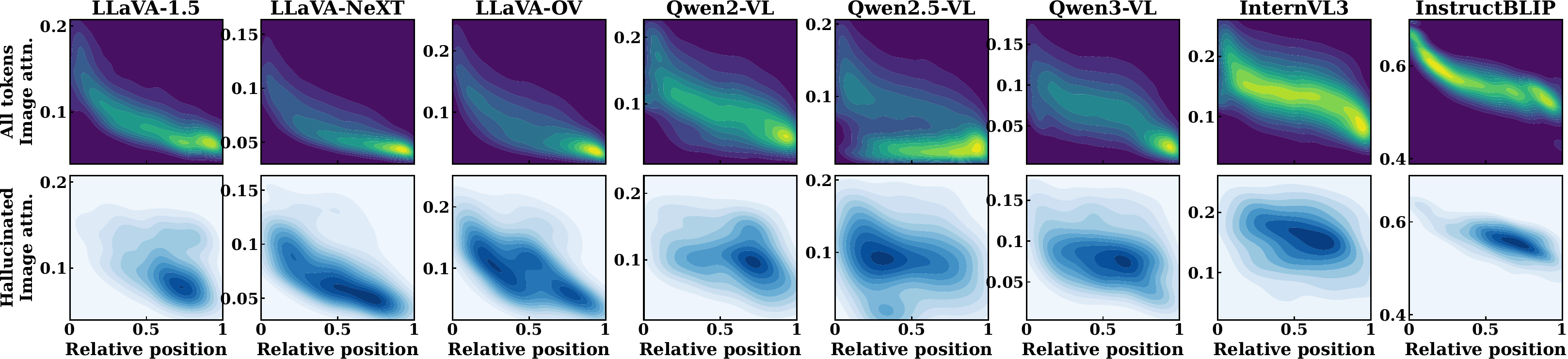}
    \caption{Joint position--attention density of generated and hallucinated tokens across all eight backbones. Kernel density of all generated tokens (top row) and hallucinated object mentions (bottom row) over relative token position and visual attention share on 500 MSCOCO captions.}
    \label{fig:sdensall}
\end{figure*}

While Figs.~2 and~3 in Section~III of the main paper illustrate three representative models, Figs.~\ref{fig:slayerhead} and~\ref{fig:sdensall} present the complete layer-head attention maps and joint position--attention densities across all eight LVLMs. Figure~\ref{fig:slayerhead} plots the per-layer and per-head image attention for the earliest 20\% and latest 20\% of generated tokens alongside their difference: across all eight models, between 91.1\% and 99.5\% of layer--head pairs allocate less attention to the image late in generation than early, with the strongest attenuation occurring across the middle-to-deep layers. Figure~\ref{fig:sdensall} shows the corresponding joint density of relative token position and image attention for all generated tokens (top row) and CHAIR-hallucinated object mentions (bottom row) across all eight backbones, confirming that hallucinated mentions consistently shift toward middle-to-late positions---occurring $4.8$ to $21.8$ percentage points later than grounded mentions within the same caption---and cluster within the low-image-attention band ($0.1\%$ to $9.7\%$ lower average visual attention than grounded tokens) across all four architectural families (LLaVA, Qwen-VL, InternVL, and InstructBLIP).

\section{Full Proof of Theorem~1}
This section provides the complete proof of Theorem~1 in Section~IV-A of the main paper. Following the main text, consider a single attention head at decoding step $t$ with query $\boldsymbol{q}\in\mathbb{R}^{d_h}$, cached keys $\{\boldsymbol{k}_m\}$, and scaled dot-product scores $z_m=\boldsymbol{q}^\top\boldsymbol{k}_m/\sqrt{d_h}$. In the full KV cache, let $A_{\mathcal{I}}=\sum_{j\in\mathcal{I}}e^{z_j}>0$, $B\ge 0$, and $A_{\mathcal{P}}=\sum_{m\in\mathcal{P}_t}e^{z_m}>0$ denote the exponential score sums in the softmax denominator over the visual prefix tokens $\mathcal{I}$, the unmerged text tokens (system-prefix tokens and the current query token $t$), and the mergeable post-visual text tokens $\mathcal{P}_t$, respectively, so that the pre-merge visual attention share is $S=A_{\mathcal{I}}/(A_{\mathcal{I}}+B+A_{\mathcal{P}})$.

\begin{theoremS1}[Visual attention recovery under bucket merging]
Fix any query $\boldsymbol{q}$ and visual prefix keys $\{\boldsymbol{k}_j\}_{j \in \mathcal{I}}$. If $\mathcal{P}_t$ is partitioned into nonempty buckets $\{D_b\}_{b=1}^{n_t}$ and each bucket is replaced by its centroid key $\bar{\boldsymbol{k}}_b = \frac{1}{|D_b|}\sum_{m \in D_b}\boldsymbol{k}_m$, the resulting visual attention share $S'$ satisfies $S' \ge S$, with strict inequality $S' > S$ whenever at least one bucket has $|D_b| \ge 2$. Moreover, if $\min_b |D_b| \ge b_0$, then $S' \ge A_{\mathcal{I}} / (A_{\mathcal{I}} + B + A_{\mathcal{P}}/b_0)$.
\end{theoremS1}
\begin{proof}
For each bucket $D_b$ ($b=1,\dots,n_t$) with cardinality $|D_b|\ge 1$, replacing the keys $\{\boldsymbol{k}_m\}_{m\in D_b}$ by their arithmetic mean $\bar{\boldsymbol{k}}_b=\frac{1}{|D_b|}\sum_{m\in D_b}\boldsymbol{k}_m$ yields the centroid score by linearity of the inner product:
\begin{equation}
\bar{z}_b = \frac{\boldsymbol{q}^\top\bar{\boldsymbol{k}}_b}{\sqrt{d_h}} = \frac{\boldsymbol{q}^\top}{\sqrt{d_h}}\!\left(\frac{1}{|D_b|}\sum_{m\in D_b}\boldsymbol{k}_m\right) = \frac{1}{|D_b|}\sum_{m\in D_b}z_m.
\end{equation}
Because $x\mapsto\exp(x)$ is strictly convex on $\mathbb{R}$, applying Jensen's inequality to the uniform average $\bar{z}_b$ and using the strict positivity $\exp(z_m)>0$ gives, for every bucket $b\in\{1,\dots,n_t\}$,
\begin{equation}
\begin{aligned}
\exp(\bar{z}_b) &= \exp\!\left(\frac{1}{|D_b|}\sum_{m\in D_b}z_m\right) \\
&\le \frac{1}{|D_b|}\sum_{m\in D_b}\exp(z_m) \le \sum_{m\in D_b}\exp(z_m).
\end{aligned}
\label{eq:supp_jensen_bucket}
\end{equation}
In Eq.~\eqref{eq:supp_jensen_bucket}, the first inequality holds with equality if and only if all scores $\{z_m\}_{m\in D_b}$ within $D_b$ are identical, whereas the second inequality holds with equality when $|D_b|=1$ and is strictly less whenever $|D_b|\ge 2$ since $\sum_{m\in D_b}\exp(z_m)>0$ and $1/|D_b|\le 1/2<1$. Consequently, $\exp(\bar{z}_b)\le\sum_{m\in D_b}\exp(z_m)$ for all $b$, with strict inequality $\exp(\bar{z}_b)<\sum_{m\in D_b}\exp(z_m)$ if and only if $|D_b|\ge 2$.

Since $\{D_b\}_{b=1}^{n_t}$ partitions $\mathcal{P}_t=\bigcup_{b=1}^{n_t}D_b$ into disjoint nonempty subsets, summing Eq.~\eqref{eq:supp_jensen_bucket} over $b=1,\dots,n_t$ decomposes the reduction in the post-visual text denominator $A_{\mathcal{P}} - A'_{\mathcal{P}}$ into two nonnegative terms:
\begin{equation}
\begin{aligned}
A_{\mathcal{P}} - A'_{\mathcal{P}} &= \sum_{b=1}^{n_t}\!\underbrace{\Big(1 - \frac{1}{|D_b|}\Big)\!\sum_{m\in D_b}\!e^{z_m}}_{\text{count compression ($\ge 0$)}} \\
&\quad + \sum_{b=1}^{n_t}\!\underbrace{\Big(\frac{1}{|D_b|}\!\sum_{m\in D_b}\!e^{z_m} - e^{\bar{z}_b}\Big)}_{\text{Jensen convexity gap ($\ge 0$)}} \ge 0,
\end{aligned}
\label{eq:supp_mass_sum}
\end{equation}
where the count-compression sum is strictly positive whenever at least one bucket satisfies $|D_b|\ge 2$ (even if all scores $z_m$ in $D_b$ are equal), and the Jensen gap provides an additional reduction whenever scores vary within a bucket. Because the visual prefix keys $\{\boldsymbol{k}_j\}_{j\in\mathcal{I}}$ and the unmerged text entries remain intact, $A_{\mathcal{I}}>0$ and $B\ge 0$ are unchanged in the merged cache, so the merged visual attention share is $S'=A_{\mathcal{I}}/(A_{\mathcal{I}}+B+A'_{\mathcal{P}})$. Since $x\mapsto A_{\mathcal{I}}/(A_{\mathcal{I}}+B+x)$ is strictly decreasing on $[0,\infty)$ for $A_{\mathcal{I}}>0$, Eq.~\eqref{eq:supp_mass_sum} immediately yields $S'\ge S$, with strict inequality $S'>S$ whenever at least one bucket has $|D_b|\ge 2$.

Finally, if $\min_{1\le b\le n_t}|D_b|\ge b_0$, then $1/|D_b|\le 1/b_0$ for all $b=1,\dots,n_t$, so Eq.~\eqref{eq:supp_jensen_bucket} implies $A'_{\mathcal{P}}\le\sum_{b=1}^{n_t}\frac{1}{|D_b|}\sum_{m\in D_b}e^{z_m}\le\frac{1}{b_0}\sum_{b=1}^{n_t}\sum_{m\in D_b}e^{z_m}=A_{\mathcal{P}}/b_0$. Substituting $A'_{\mathcal{P}}\le A_{\mathcal{P}}/b_0$ into $S'=A_{\mathcal{I}}/(A_{\mathcal{I}}+B+A'_{\mathcal{P}})$ yields $S'\ge A_{\mathcal{I}}/(A_{\mathcal{I}}+B+A_{\mathcal{P}}/b_0)$, completing the proof.
\end{proof}

\noindent\textbf{Effect on Value Aggregation.}
Let $\boldsymbol{o}_{\mathcal{I}}=\sum_{j\in\mathcal{I}}(e^{z_j}/A_{\mathcal{I}})\boldsymbol{v}_j$ denote the normalized visual context vector. Under bucket merging in Eq.~(4) of the main paper, the attention head output on the merged cache is $\boldsymbol{o}' = S'\boldsymbol{o}_{\mathcal{I}} + \boldsymbol{o}'_{\mathcal{U}} + \sum_{b=1}^{n_t}\alpha'_b\bar{\boldsymbol{v}}_b$, where $\alpha'_b = e^{\bar{z}_b}/(A_{\mathcal{I}}+B+A'_{\mathcal{P}})$. Thus, keeping the visual prefix intact preserves the within-image relative attention distribution $\boldsymbol{o}_{\mathcal{I}}$ without distortion while amplifying its overall weight from $S$ to $S'>S$, and averaging the bucket values $\bar{\boldsymbol{v}}_b=\frac{1}{|D_b|}\sum_{m\in D_b}\boldsymbol{v}_m$ retains the first-order local semantic summary of every post-visual token $m\in D_b$ rather than discarding non-anchor values as in hard token pruning.

\begin{table*}[!t]
\centering
\caption{Dataset-wide closed-loop mechanistic verification across eight LVLMs on CHAIR at $\lambda{=}0.4$ (numerical breakdown for Fig.~6 of the main paper): \textbf{(a)} step-level visual attention restoration, online gate response, and object-level intervention; \textbf{(b)} eight-LVLM mean bucket composition, granularity, and representation fidelity across KV merging rules.}
\label{tab:smech8b}
\small
\setlength{\tabcolsep}{1.445pt}
\begin{tabular*}{\textwidth}{@{\extracolsep{\fill}}l cccc cc cc cc@{}}
\toprule
\multicolumn{11}{@{}l}{\textit{(a) Step-level visual attention restoration ($S_t{\to}S'_t$, \%), online gate response ($\alpha_t$), and object intervention}} \\
\midrule
Model & Early $S_t{\to}S'_t$ & Late $S_t{\to}S'_t$ & GT $S_t{\to}S'_t$ & Hal $S_t{\to}S'_t$ & $\Pr(S'_t{>}S_t)$ & Hal@Late & $\alpha_t$ Early${\to}$Late & $\Delta p(y^{\mathrm{hal}})$ & Hal Rev.\ & GT Keep \\
\midrule
LLaVA-1.5 & 11.1${\to}$19.2 & 6.5${\to}$11.4 & 13.5${\to}$15.3 & 11.6${\to}$13.4 & 99.2/72.3 & 75.6\% & 0.63${\to}$0.83 & $-$69.2\% & 71.9/54.6 & 82.1/91.4 \\
LLaVA-NeXT & 9.1${\to}$14.8 & 5.9${\to}$13.8 & 11.0${\to}$13.9 & 9.5${\to}$13.4 & 100/90.5 & 42.1\% & 0.67${\to}$0.83 & $-$66.1\% & 68.4/40.3 & 85.8/93.5 \\
LLaVA-OV & 12.7${\to}$20.7 & 7.6${\to}$15.4 & 15.6${\to}$21.0 & 12.4${\to}$19.6 & 96.7/92.3 & 43.9\% & 0.71${\to}$0.82 & $-$66.9\% & 69.9/36.7 & 84.4/92.7 \\
Qwen2-VL & 13.2${\to}$17.4 & 8.3${\to}$13.1 & 14.7${\to}$19.2 & 12.5${\to}$17.1 & 100/100 & 61.7\% & 0.58${\to}$0.83 & $-$66.8\% & 71.8/27.7 & 87.7/94.1 \\
Qwen2.5-VL & 11.4${\to}$14.9 & 7.6${\to}$11.5 & 13.5${\to}$17.5 & 11.7${\to}$16.0 & 98.3/99.5 & 50.9\% & 0.53${\to}$0.66 & $-$65.1\% & 68.4/29.3 & 90.4/94.5 \\
Qwen3-VL & 9.8${\to}$12.2 & 7.8${\to}$10.8 & 12.6${\to}$15.0 & 10.3${\to}$13.2 & 97.5/91.8 & 60.4\% & 0.53${\to}$0.86 & $-$56.3\% & 64.2/14.5 & 92.5/94.7 \\
InternVL3 & 16.5${\to}$19.1 & 13.3${\to}$18.5 & 18.6${\to}$18.7 & 16.8${\to}$19.8 & 89.2/76.8 & 56.7\% & 0.52${\to}$0.76 & $-$67.8\% & 77.3/41.2 & 90.4/88.7 \\
InstructBLIP & 59.1${\to}$81.8 & 53.3${\to}$77.5 & 60.7${\to}$82.1 & 57.5${\to}$79.1 & 100/100 & 72.7\% & 0.60${\to}$0.63 & $-$72.1\% & 80.5/46.0 & 90.2/89.5 \\
\midrule
\textbf{Average} & \textbf{17.9${\to}$25.0} & \textbf{13.8${\to}$21.5} & \textbf{20.0${\to}$25.3} & \textbf{17.8${\to}$24.0} & \textbf{97.6/90.4} & \textbf{58.0\%} & \textbf{0.60${\to}$0.78} & \textbf{$-$66.3\%} & \textbf{71.6/36.3} & \textbf{87.9/92.4} \\
\end{tabular*}\\[1pt]
\begin{tabular*}{\textwidth}{@{\extracolsep{\fill}}l ccccccccc@{}}
\toprule
\multicolumn{10}{@{}l}{\textit{(b) Bucket lexical composition, late-span granularity, and value fidelity across KV merging rules ($\lambda{=}0.4$)}} \\
\midrule
Strategy & Func.\ Anchors & GT Anchors & Late Bucket & $\Delta S_t^{\mathrm{late}}{\uparrow}$ & $\Delta S_t^{\mathrm{hal}}{\uparrow}$ & Val $\mathrm{Cos}{\uparrow}$ & Out $\mathrm{Cos}{\uparrow}$ & GT Keep & $C_I{\downarrow}$ / F1${\uparrow}$ \\
\midrule
High-attn & 53.2\% & 6.05\% & 10.18 tok & $+$4.01\% & $+$2.44\% & 79.3\% & 90.6\% & 87.7/92.4 & 8.4 / 76.9 \\
Uniform & 55.5\% & 5.20\% & 2.59 tok & $+$7.60\% & $+$5.68\% & \textbf{85.0\%} & \textbf{92.6\%} & \textbf{88.3}/\textbf{92.4} & 8.4 / \textbf{77.7} \\
Prune-low & 55.6\% & 4.23\% & 1.00 tok & $+$6.91\% & $+$5.38\% & 68.1\% & 91.3\% & 87.3/91.9 & 9.2 / 77.1 \\
\textbf{IKOD} & \textbf{55.6\%} & \textbf{4.23\%} & \textbf{3.70 tok} & \textbf{$+$7.69\%} & \textbf{$+$6.17\%} & 76.7\% & 91.2\% & 87.9/\textbf{92.4} & \textbf{8.0} / 77.3 \\
\bottomrule
\end{tabular*}
\end{table*}

\noindent\textbf{Dataset-Wide Mechanistic Verification and Bucket Anatomy.}
Table~\ref{tab:smech8b} provides the full eight-backbone numerical breakdown supporting Figure~6 and Sections~IV and~V of the main paper across 3,419 grounded-object onset steps, 1,694 hallucination-onset steps, and 801,118 total decoding steps on CHAIR. In Table~\ref{tab:smech8b}(a), base visual attention is lower on hallucination-onset steps than on grounded-object onset steps across every single backbone, dropping from an eight-model mean of 20.0\% to 17.8\%, while 58.0\% of hallucinated objects occur in the second half of generation ($t \ge T/2$). Executing the merged-branch forward pass ($\lambda{=}0.4$) confirms Theorem~1 ($S'_t > S_t$) on 97.6\% of late decoding steps and 90.4\% of hallucination-onset steps, boosting mean late-stage visual attention by 67.5\% relative from 13.8\% to 21.5\% and restoring visual attention on hallucination-onset steps by 34.0\% relative from 17.8\% to 24.0\%. Simultaneously, IKOD-A's online decay gate $\alpha_t$ rises from 0.60 in early decoding to 0.78 in late decoding and reaches 0.94 at the 90th percentile, cutting the probability of hallucinated nouns by 66.3\% relative on the merged branch and reversing 36.3\% of hallucinated predictions on the single step alone under collaborative fusion while preserving 87.9\% and 92.4\% of grounded object tokens under IKOD and IKOD-A. In Table~\ref{tab:smech8b}(b), analyzing the internal bucket composition across merging rules reveals why lowest-$s_m$ centroid averaging outperforms alternative strategies: 55.6\% of selected lowest-$s_m$ anchors are function words and punctuation whereas only 4.23\% are grounded object nouns, placing anchors directly inside late attention-decayed spans at 3.70 tokens per bucket and yielding the strongest visual attention gain at hallucination-onset steps of $+$6.17 percentage points. Conversely, highest-$s_m$ anchors cluster in early grounded clauses and leave late attention-decayed spans in coarse 10.18-token buckets, cutting hallucination-step visual attention gain to $+$2.44 percentage points, whereas hard pruning (\textit{Prune low-attn}) discards non-anchor grounded object tokens from the KV cache, dropping grounded value-representation cosine similarity from 76.7\% to 68.1\% and worsening CHAIR $C_I$ from 8.0\% to 9.2\%.

\begin{table}[!t]
\centering
\caption{Zero-shot POPE-COCO F1 ($\uparrow$, \%) across four 13B--32B model families without retuning. Bold and underline mark the best and second-best results within each block.}
\label{tab:scale13b_pope}
\small
\setlength{\tabcolsep}{1.654pt}
\begin{tabular}{ll cccc}
\toprule
Model & Method & Random & Popular & Adversarial & Average \\
\midrule
\multirow{3}{*}{LLaVA-1.5-13B} & Greedy & 86.28 & 85.58 & 83.97 & 85.28 \\
 & \textbf{IKOD} & \textbf{87.85} & \textbf{86.68} & \textbf{84.87} & \textbf{86.47} \\
 & \textbf{IKOD-A} & \underline{87.18} & \underline{86.38} & \underline{84.63} & \underline{86.06} \\
\midrule
\multirow{3}{*}{InstructBLIP-13B} & Greedy & \underline{84.79} & \underline{82.92} & \underline{80.89} & \underline{82.87} \\
 & \textbf{IKOD} & \textbf{85.64} & \textbf{83.26} & \textbf{81.27} & \textbf{83.39} \\
 & \textbf{IKOD-A} & \textbf{85.64} & \textbf{83.26} & \textbf{81.27} & \textbf{83.39} \\
\midrule
\multirow{3}{*}{InternVL3-14B} & Greedy & 94.22 & \textbf{90.16} & \textbf{86.96} & \textbf{90.45} \\
 & \textbf{IKOD} & \underline{94.34} & \underline{89.85} & \underline{86.79} & \underline{90.33} \\
 & \textbf{IKOD-A} & \textbf{94.52} & 89.20 & 86.11 & 89.94 \\
\midrule
\multirow{3}{*}{Qwen2.5-VL-32B} & Greedy & 87.58 & 86.65 & 85.30 & 86.51 \\
 & \textbf{IKOD} & \textbf{89.29} & \textbf{87.72} & \textbf{86.45} & \textbf{87.82} \\
 & \textbf{IKOD-A} & \underline{88.55} & \underline{87.33} & \underline{86.08} & \underline{87.32} \\
\midrule
\multirow{3}{*}{\textbf{Average}} & Greedy & 88.22 & 86.33 & 84.28 & 86.28 \\
 & \textbf{IKOD} & \textbf{89.28} & \textbf{86.88} & \textbf{84.85} & \textbf{87.00} \\
 & \textbf{IKOD-A} & \underline{88.97} & \underline{86.54} & \underline{84.52} & \underline{86.68} \\
\bottomrule
\end{tabular}
\end{table}

\begin{figure*}[!t]
\centering
\includegraphics[width=0.94\textwidth]{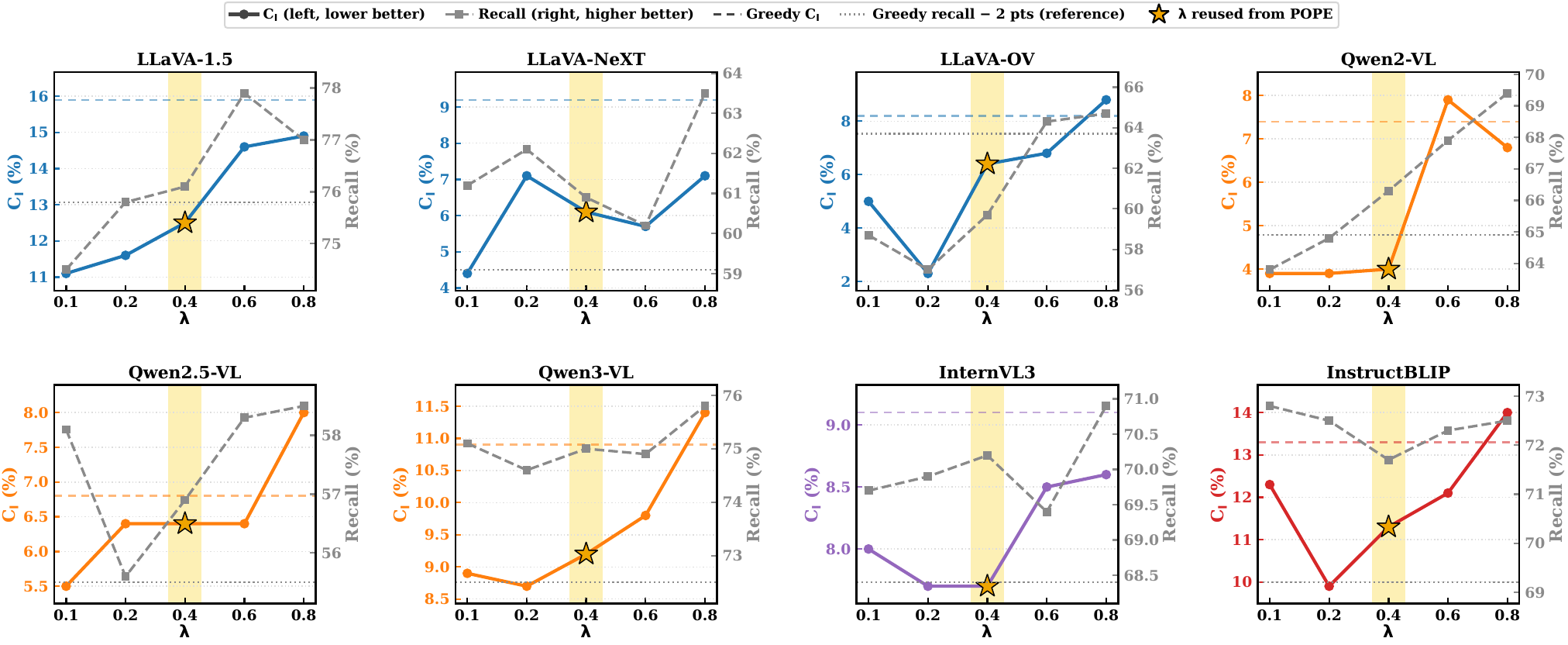}
\caption{Per-backbone sensitivity to the retained-anchor ratio $\lambda$ on CHAIR (512-token budget), showing $C_I$ (left axis) and recall (right axis) at the main-table family $\alpha$. Dashed line: Greedy $C_I$; dotted line: Greedy recall minus two points; star and band: universal $\lambda=0.4$.}
\label{fig:slambdasupp}
\end{figure*}

\begin{figure*}[!t]
\centering
\includegraphics[width=0.94\textwidth]{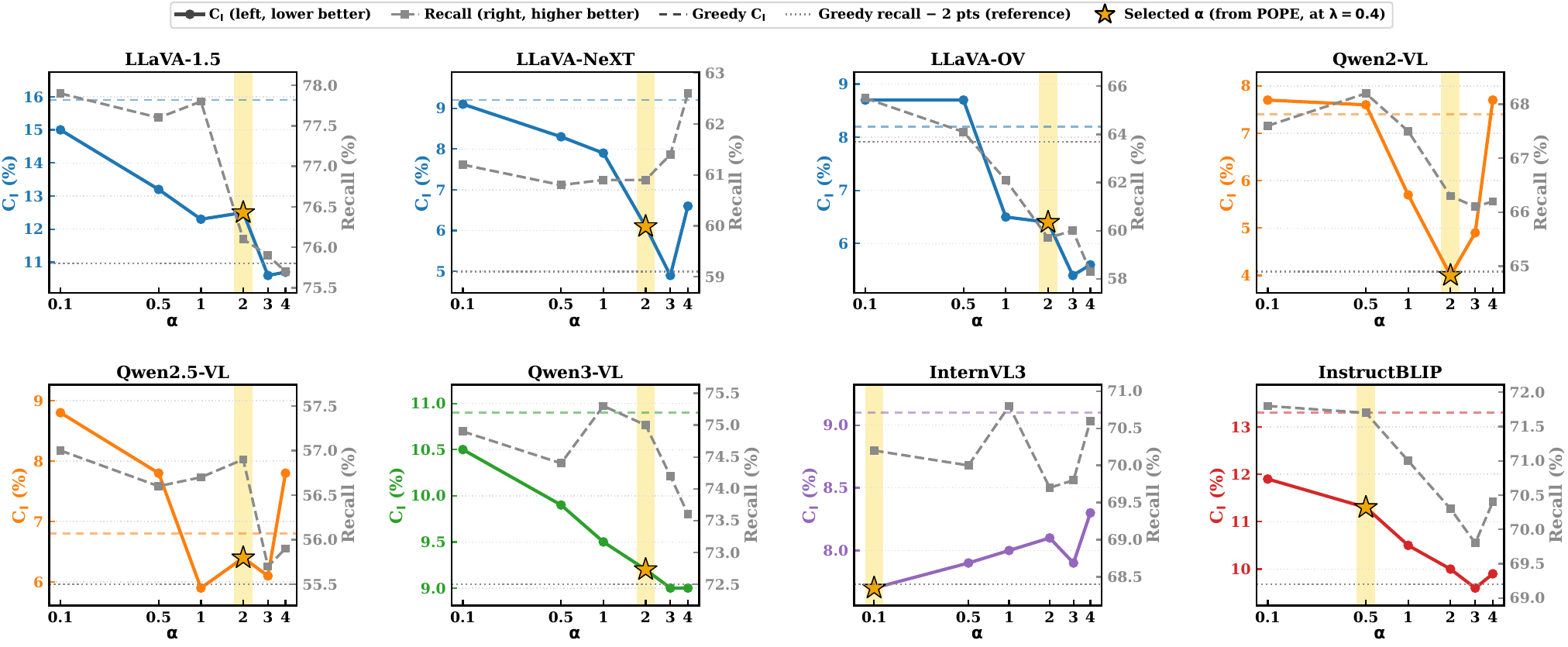}
\caption{Per-backbone sensitivity to the collaboration weight $\alpha$ on CHAIR (512-token budget), showing $C_I$ (left axis) and recall (right axis) at the universal anchor ratio $\lambda=0.4$. Dashed line: Greedy $C_I$; dotted line: Greedy recall minus two points; star and band: family $\alpha$ transferred from POPE.}
\label{fig:salphasupp}
\end{figure*}

\section{Hyperparameter Sensitivity on CHAIR}
Figures~\ref{fig:slambdasupp} and~\ref{fig:salphasupp} evaluate per-backbone sensitivity to the retained-anchor ratio $\lambda$ and collaboration weight $\alpha$ on 512-token CHAIR captioning across all eight LVLMs. In Figure~\ref{fig:slambdasupp}, overly aggressive merging ($\lambda\le 0.2$) blurs distinct entity mentions and lowers recall, whereas overly mild compression ($\lambda\ge 0.6$) leaves the post-visual text denominator large and allows $C_I$ to rebound; the universal ratio $\lambda=0.4$ consistently sits at the knee of both curves across all eight models. In Figure~\ref{fig:salphasupp}, the POPE-selected family weights transfer directly to long-form captioning at $\lambda=0.4$: ViT-MLP and M-RoPE backbones (LLaVA and Qwen-VL) benefit from stronger collaboration ($\alpha=2.0$), whereas dense-prefix InternVL3 and Q-Former-based InstructBLIP prefer lighter weights ($\alpha=0.1$ and $0.5$), directly motivating the online decay gate in IKOD-A.

\begin{table*}[!t]
\centering
\caption{Per-backbone three-split mean POPE-COCO F1 ($\uparrow$, \%) across five stochastic sampling strategies and temperatures on all eight LVLMs (eight-LVLM means in Table~VIII of the main paper). Bold and underline mark the best and second-best results within each block.}
\label{tab:spope5strat}
\small
\setlength{\tabcolsep}{1.767pt}
\begin{tabular}{llccccccccc}
\toprule
Strategy & Method & LLaVA-1.5 & LLaVA-NeXT & LLaVA-OV & Qwen2-VL & Qwen2.5-VL & Qwen3-VL & InternVL3 & InstructBLIP & \textbf{Average} \\
\midrule
\multirow{3}{*}{\begin{tabular}[c]{@{}l@{}}Top-$P$\\($p{=}0.9, T{=}1.0$)\end{tabular}} & Base & 82.44 & 82.77 & 87.62 & 87.01 & 85.49 & 88.33 & \textbf{90.09} & \textbf{80.93} & 85.59 \\
 & \textbf{IKOD} & \textbf{86.03} & \textbf{87.21} & \textbf{89.91} & \textbf{87.56} & \underline{86.36} & \textbf{89.50} & \underline{90.01} & 80.64 & \textbf{87.15} \\
 & \textbf{IKOD-A} & \underline{83.71} & \underline{84.37} & \underline{88.68} & \underline{87.49} & \textbf{86.68} & \underline{88.90} & 89.68 & \underline{80.72} & \underline{86.28} \\
\midrule
\multirow{3}{*}{\begin{tabular}[c]{@{}l@{}}Top-$K$\\($k{=}50, T{=}1.0$)\end{tabular}} & Base & 80.09 & 80.71 & 86.78 & 86.20 & 84.90 & 88.19 & 89.51 & \underline{80.13} & 84.56 \\
 & \textbf{IKOD} & \textbf{85.62} & \textbf{86.81} & \textbf{89.82} & \textbf{87.31} & \underline{86.26} & \textbf{89.46} & \textbf{89.89} & \textbf{80.33} & \textbf{86.94} \\
 & \textbf{IKOD-A} & \underline{83.33} & \underline{83.86} & \underline{88.34} & \underline{87.29} & \textbf{86.51} & \underline{88.88} & \underline{89.66} & \textbf{80.33} & \underline{86.03} \\
\midrule
\multirow{3}{*}{\begin{tabular}[c]{@{}l@{}}Multinomial\\($p{=}1.0, T{=}1.0$)\end{tabular}} & Base & 80.09 & 80.71 & 86.78 & 86.20 & 84.90 & 88.31 & 89.51 & 78.27 & 84.35 \\
 & \textbf{IKOD} & \textbf{85.62} & \textbf{86.81} & \textbf{89.82} & \textbf{87.31} & \underline{86.13} & \textbf{89.46} & \textbf{89.89} & \underline{80.33} & \textbf{86.92} \\
 & \textbf{IKOD-A} & \underline{83.33} & \underline{83.86} & \underline{88.34} & \underline{87.29} & \textbf{86.51} & \underline{88.74} & \underline{89.66} & \textbf{80.45} & \underline{86.02} \\
\midrule
\multirow{3}{*}{\begin{tabular}[c]{@{}l@{}}Top-$K$\\($k{=}50, T{=}0.7$)\end{tabular}} & Base & 82.68 & 83.55 & 87.25 & 86.68 & 85.69 & 88.52 & 89.82 & 81.60 & 85.72 \\
 & \textbf{IKOD} & \textbf{86.17} & \textbf{87.40} & \textbf{89.98} & \textbf{87.49} & \underline{86.11} & \textbf{89.55} & \textbf{90.25} & \underline{81.84} & \textbf{87.35} \\
 & \textbf{IKOD-A} & \underline{84.75} & \underline{84.80} & \underline{89.07} & \underline{87.41} & \textbf{86.49} & \underline{89.06} & \underline{89.89} & \textbf{81.96} & \underline{86.68} \\
\midrule
\multirow{3}{*}{\begin{tabular}[c]{@{}l@{}}Top-$K$\\($k{=}50, T{=}1.5$)\end{tabular}} & Base & 74.04 & 76.44 & 84.29 & 84.09 & 84.14 & 87.71 & 88.09 & 74.40 & 81.65 \\
 & \textbf{IKOD} & \textbf{84.22} & \textbf{85.93} & \textbf{89.60} & \underline{87.19} & \textbf{85.83} & \textbf{89.29} & \textbf{89.39} & \textbf{79.03} & \textbf{86.31} \\
 & \textbf{IKOD-A} & \underline{81.92} & \underline{82.67} & \underline{88.37} & \textbf{87.31} & \underline{85.64} & \underline{88.57} & \underline{89.25} & \underline{78.40} & \underline{85.27} \\
\bottomrule
\end{tabular}
\end{table*}

\section{Cross-Scale POPE Results on 13B--32B Backbones}
Alongside the CHAIR cross-scale evaluation in Table~VII of the main paper, Table~\ref{tab:scale13b_pope} reports zero-shot POPE-COCO F1 across all three splits on the four scaled 13B--32B backbones using the 7B/8B settings ($\lambda=0.4$) without retuning. Without retuning, IKOD and tuning-free IKOD-A raise the four-family average F1 from 86.28\% to 87.00\% and 86.68\%, lifting three-split mean F1 by $1.31$ and $0.81$ points on Qwen2.5-VL-32B, by $1.19$ and $0.78$ points on LLaVA-1.5-13B, and by $0.52$ points on InstructBLIP-13B.

\begin{table*}[!t]
\centering
\caption{Per-backbone CHAIR-500 instance hallucination $C_I\downarrow$ (\%) under Top-$P$ ($p{=}0.9, T{=}1.0$) sampling across three independent random seeds on all eight LVLMs (eight-LVLM means in Table~IX of the main paper). Bold and underline mark the best and second-best results within each block.}
\label{tab:seedsweep}
\small
\setlength{\tabcolsep}{1.835pt}
\begin{tabular}{llccccccccc}
\toprule
Random Seed & Method & LLaVA-1.5 & LLaVA-NeXT & LLaVA-OV & Qwen2-VL & Qwen2.5-VL & Qwen3-VL & InternVL3 & InstructBLIP & \textbf{Average} \\
\midrule
\multirow{3}{*}{Seed 42} & Base & 19.2 & 11.2 & 10.0 & 10.7 & 9.1 & 10.8 & \underline{8.9} & 17.2 & 12.1 \\
 & \textbf{IKOD} & \textbf{13.6} & \textbf{7.5} & \underline{6.1} & \underline{7.3} & \underline{8.7} & \textbf{9.2} & 9.3 & \underline{13.8} & \textbf{9.4} \\
 & \textbf{IKOD-A} & \underline{17.3} & \underline{7.9} & \textbf{5.8} & \textbf{6.4} & \textbf{8.4} & \underline{9.9} & \textbf{8.1} & \textbf{12.5} & \underline{9.5} \\
\midrule
\multirow{3}{*}{Seed 1234} & Base & 19.2 & 10.6 & 10.0 & 10.9 & 10.1 & 11.1 & \underline{8.7} & 17.1 & 12.2 \\
 & \textbf{IKOD} & \textbf{14.3} & \textbf{7.2} & \textbf{3.8} & \textbf{5.6} & \textbf{6.0} & \textbf{9.4} & 9.2 & \textbf{12.6} & \textbf{8.5} \\
 & \textbf{IKOD-A} & \underline{15.1} & \underline{8.4} & \underline{7.5} & \underline{7.6} & \underline{7.5} & \underline{10.3} & \textbf{8.0} & \underline{14.0} & \underline{9.8} \\
\midrule
\multirow{3}{*}{Seed 2026} & Base & 17.8 & 10.9 & 10.4 & 7.6 & 11.4 & 11.3 & 9.0 & 17.6 & 12.0 \\
 & \textbf{IKOD} & \textbf{12.8} & \textbf{6.7} & \textbf{6.3} & \textbf{5.9} & \underline{8.1} & \textbf{9.0} & \textbf{8.4} & \underline{14.1} & \textbf{8.9} \\
 & \textbf{IKOD-A} & \underline{16.9} & \underline{8.4} & \underline{6.7} & \underline{6.8} & \textbf{7.8} & \underline{10.4} & \underline{8.7} & \textbf{13.5} & \underline{9.9} \\
\midrule
\multirow{3}{*}{\textbf{Average $\pm$ Std}} & Base & 18.7$\pm$0.8 & 10.9$\pm$0.3 & 10.1$\pm$0.2 & 9.7$\pm$1.9 & 10.2$\pm$1.2 & 11.1$\pm$0.3 & \underline{8.9}$\pm$0.2 & 17.3$\pm$0.3 & 12.1$\pm$0.1 \\
 & \textbf{IKOD} & \textbf{13.6}$\pm$0.8 & \textbf{7.1}$\pm$0.4 & \textbf{5.4}$\pm$1.4 & \textbf{6.3}$\pm$0.9 & \textbf{7.6}$\pm$1.4 & \textbf{9.2}$\pm$0.2 & 9.0$\pm$0.5 & \underline{13.5}$\pm$0.8 & \textbf{9.0}$\pm$0.5 \\
 & \textbf{IKOD-A} & \underline{16.4}$\pm$1.2 & \underline{8.2}$\pm$0.3 & \underline{6.7}$\pm$0.9 & \underline{6.9}$\pm$0.6 & \underline{7.9}$\pm$0.5 & \underline{10.2}$\pm$0.3 & \textbf{8.3}$\pm$0.4 & \textbf{13.3}$\pm$0.8 & \underline{9.7}$\pm$0.2 \\
\bottomrule
\end{tabular}
\end{table*}

\begin{table*}[!t]
\centering
\caption{Per-backbone CHAIR-500 metrics (\%) and mean caption length (Words) across three alternative prompt phrasings under unified $\lambda{=}0.4$ (eight-LVLM means in Table~IX of the main paper). Bold and underline mark the best and second-best eight-backbone averages per prompt.}
\label{tab:spromptlen}
\small
\setlength{\tabcolsep}{2.729pt}
\setlength{\cmidrulekern}{2.729pt}
\begin{tabular}{ll ccccc ccccc ccccc}
\toprule
& & \multicolumn{5}{c}{Greedy} & \multicolumn{5}{c}{\textbf{IKOD}} & \multicolumn{5}{c}{\textbf{IKOD-A}} \\
\cmidrule(lr){3-7}\cmidrule(lr){8-12}\cmidrule(l){13-17}
Prompt & Model & C$_S{\downarrow}$ & C$_I{\downarrow}$ & Recall${\uparrow}$ & F1${\uparrow}$ & Words & C$_S{\downarrow}$ & C$_I{\downarrow}$ & Recall${\uparrow}$ & F1${\uparrow}$ & Words & C$_S{\downarrow}$ & C$_I{\downarrow}$ & Recall${\uparrow}$ & F1${\uparrow}$ & Words \\
\midrule
\multirow{9}{*}{\begin{tabular}[c]{@{}l@{}}\textit{Describe}\\\textit{this image.}\end{tabular}} & LLaVA-1.5 & 49.2 & 15.0 & 78.3 & 81.5 & 82.6 & 40.2 & 11.6 & 76.1 & 81.8 & 87.9 & 51.0 & 15.5 & 78.3 & 81.3 & 84.4 \\
 & LLaVA-NeXT & 27.0 & 8.3 & 62.1 & 74.1 & 150.0 & 26.2 & 6.2 & 62.0 & 74.7 & 156.1 & 26.8 & 7.5 & 62.8 & 74.8 & 148.7 \\
 & LLaVA-OV & 44.0 & 8.6 & 72.5 & 80.9 & 309.1 & 26.0 & 4.5 & 61.2 & 74.6 & 390.1 & 34.0 & 4.7 & 68.7 & 79.8 & 356.3 \\
 & Qwen2-VL & 31.8 & 8.8 & 68.1 & 78.0 & 159.0 & 27.2 & 11.1 & 66.5 & 76.1 & 165.3 & 28.0 & 7.4 & 69.8 & 79.6 & 149.9 \\
 & Qwen2.5-VL & 21.8 & 7.9 & 57.3 & 70.6 & 87.7 & 18.6 & 5.3 & 54.9 & 69.5 & 97.0 & 17.6 & 6.1 & 55.5 & 69.8 & 89.9 \\
 & Qwen3-VL & 45.6 & 9.8 & 72.9 & 80.6 & 249.2 & 41.2 & 8.5 & 73.0 & 81.2 & 241.3 & 43.0 & 9.1 & 72.3 & 80.5 & 240.7 \\
 & InternVL3 & 51.0 & 9.0 & 75.9 & 82.8 & 405.1 & 47.4 & 8.6 & 75.4 & 82.6 & 405.7 & 43.4 & 7.8 & 74.2 & 82.2 & 410.7 \\
 & InstructBLIP & 35.8 & 10.1 & 69.8 & 78.6 & 87.8 & 33.0 & 8.3 & 68.3 & 78.3 & 95.6 & 33.6 & 8.1 & 68.9 & 78.8 & 95.2 \\
\cmidrule(l){2-17}
 & \textbf{Average} & 38.3 & 9.7 & \textbf{69.6} & \textbf{78.4} & 191.3 & \textbf{32.5} & \textbf{8.0} & 67.2 & \underline{77.4} & 204.9 & \underline{34.7} & \underline{8.3} & \underline{68.8} & \textbf{78.4} & 197.0 \\
\midrule
\multirow{9}{*}{\begin{tabular}[c]{@{}l@{}}\textit{Please provide a}\\\textit{detailed description}\\\textit{of this image.}\end{tabular}} & LLaVA-1.5 & 52.6 & 16.1 & 77.4 & 80.5 & 83.4 & 39.8 & 12.2 & 76.7 & 81.9 & 90.0 & 50.4 & 14.9 & 77.8 & 81.3 & 85.2 \\
 & LLaVA-NeXT & 31.8 & 9.4 & 61.5 & 73.3 & 166.1 & 28.4 & 6.9 & 59.2 & 72.4 & 174.1 & 28.8 & 7.8 & 61.4 & 73.7 & 165.7 \\
 & LLaVA-OV & 38.8 & 8.9 & 67.5 & 77.5 & 239.2 & 22.6 & 4.6 & 58.8 & 72.8 & 365.7 & 32.8 & 6.4 & 66.4 & 77.7 & 298.0 \\
 & Qwen2-VL & 26.8 & 8.4 & 67.6 & 77.8 & 111.9 & 22.2 & 4.8 & 66.6 & 78.4 & 140.5 & 26.4 & 6.7 & 66.9 & 77.9 & 120.2 \\
 & Qwen2.5-VL & 25.6 & 7.7 & 58.0 & 71.2 & 137.2 & 20.8 & 5.2 & 54.5 & 69.2 & 141.4 & 22.8 & 9.7 & 54.3 & 67.8 & 128.0 \\
 & Qwen3-VL & 49.8 & 10.0 & 74.9 & 81.8 & 312.0 & 48.2 & 9.3 & 74.2 & 81.6 & 304.4 & 49.6 & 9.4 & 74.0 & 81.5 & 304.0 \\
 & InternVL3 & 26.4 & 8.8 & 67.9 & 77.8 & 94.1 & 26.0 & 8.1 & 68.3 & 78.4 & 96.1 & 26.0 & 7.4 & 68.4 & 78.7 & 102.2 \\
 & InstructBLIP & 43.8 & 12.2 & 71.0 & 78.5 & 87.8 & 40.8 & 11.1 & 71.1 & 79.0 & 96.9 & 42.2 & 11.5 & 70.7 & 78.6 & 96.7 \\
\cmidrule(l){2-17}
 & \textbf{Average} & 37.0 & 10.2 & \textbf{68.2} & \textbf{77.3} & 154.0 & \textbf{31.1} & \textbf{7.8} & 66.2 & 76.7 & 176.1 & \underline{34.9} & \underline{9.2} & \underline{67.5} & \underline{77.2} & 162.5 \\
\midrule
\multirow{9}{*}{\begin{tabular}[c]{@{}l@{}}\textit{What is showing}\\\textit{in this image?}\end{tabular}} & LLaVA-1.5 & 6.6 & 3.9 & 47.3 & 63.4 & 16.2 & 7.4 & 4.4 & 47.6 & 63.6 & 18.2 & 6.8 & 3.9 & 48.6 & 64.6 & 17.1 \\
 & LLaVA-NeXT & 17.6 & 5.9 & 63.1 & 75.5 & 81.0 & 18.0 & 5.2 & 61.2 & 74.4 & 87.9 & 16.6 & 4.8 & 62.4 & 75.4 & 82.8 \\
 & LLaVA-OV & 1.2 & 1.5 & 22.7 & 36.9 & 3.8 & 2.0 & 1.5 & 26.5 & 41.8 & 13.4 & 1.8 & 1.7 & 25.8 & 40.9 & 9.4 \\
 & Qwen2-VL & 14.2 & 5.6 & 60.1 & 73.4 & 43.4 & 13.0 & 4.3 & 60.2 & 73.9 & 48.3 & 12.2 & 6.4 & 60.6 & 73.6 & 48.3 \\
 & Qwen2.5-VL & 20.6 & 7.5 & 56.8 & 70.4 & 76.3 & 14.0 & 4.1 & 53.6 & 68.8 & 79.8 & 14.4 & 4.3 & 53.9 & 69.0 & 81.3 \\
 & Qwen3-VL & 37.8 & 8.5 & 70.2 & 79.4 & 192.7 & 39.6 & 9.1 & 70.8 & 79.6 & 197.0 & 39.4 & 7.9 & 71.5 & 80.5 & 192.5 \\
 & InternVL3 & 17.2 & 6.2 & 64.3 & 76.3 & 59.4 & 17.0 & 6.4 & 65.2 & 76.9 & 60.6 & 15.8 & 5.3 & 65.6 & 77.5 & 64.4 \\
 & InstructBLIP & 26.8 & 9.5 & 65.0 & 75.7 & 41.8 & 18.8 & 7.7 & 62.6 & 74.6 & 33.1 & 18.2 & 7.5 & 62.6 & 74.7 & 32.7 \\
\cmidrule(l){2-17}
 & \textbf{Average} & 17.8 & 6.1 & \underline{56.2} & 68.9 & 64.3 & \underline{16.2} & \underline{5.3} & 56.0 & \underline{69.2} & 67.3 & \textbf{15.7} & \textbf{5.2} & \textbf{56.4} & \textbf{69.5} & 66.1 \\
\bottomrule
\end{tabular}
\end{table*}

\begin{table*}[!t]
\centering
\caption{Per-backbone CHAIR-500 metrics (\%) and mean caption length (Words) under the 128-token generation budget at unified $\lambda{=}0.4$ (eight-LVLM means in Table~IX of the main paper). Bold and underline mark the best and second-best eight-backbone averages.}
\label{tab:lengthcontrol}
\small
\setlength{\tabcolsep}{4.770pt}
\setlength{\cmidrulekern}{4.770pt}
\begin{tabular}{l ccccc ccccc ccccc}
\toprule
& \multicolumn{5}{c}{Greedy} & \multicolumn{5}{c}{\textbf{IKOD}} & \multicolumn{5}{c}{\textbf{IKOD-A}} \\
\cmidrule(lr){2-6}\cmidrule(lr){7-11}\cmidrule(l){12-16}
Model & C$_S\downarrow$ & C$_I\downarrow$ & Recall$\uparrow$ & F1$\uparrow$ & Words & C$_S\downarrow$ & C$_I\downarrow$ & Recall$\uparrow$ & F1$\uparrow$ & Words & C$_S\downarrow$ & C$_I\downarrow$ & Recall$\uparrow$ & F1$\uparrow$ & Words \\
\midrule
LLaVA-1.5 & 50.8 & 15.8 & 77.3 & 80.6 & 82.7 & 40.2 & 12.7 & 75.6 & 81.0 & 84.9 & 47.2 & 14.1 & 79.0 & 82.3 & 83.7 \\
LLaVA-NeXT & 22.4 & 8.6 & 54.5 & 68.3 & 90.2 & 20.0 & 6.1 & 56.8 & 70.8 & 89.8 & 22.2 & 7.2 & 55.2 & 69.2 & 89.4 \\
LLaVA-OV & 23.8 & 7.7 & 55.9 & 69.6 & 106.4 & 20.2 & 6.1 & 52.5 & 67.3 & 106.6 & 22.6 & 6.2 & 55.6 & 69.8 & 106.7 \\
Qwen2-VL & 24.2 & 7.4 & 66.2 & 77.2 & 75.1 & 21.0 & 5.0 & 66.1 & 78.0 & 77.3 & 23.0 & 6.8 & 68.1 & 78.7 & 75.4 \\
Qwen2.5-VL & 23.6 & 8.7 & 54.4 & 68.2 & 94.4 & 22.4 & 6.7 & 54.3 & 68.7 & 95.5 & 23.4 & 8.0 & 53.5 & 67.8 & 93.8 \\
Qwen3-VL & 23.0 & 8.1 & 47.3 & 62.5 & 100.9 & 22.6 & 7.1 & 47.6 & 62.9 & 101.7 & 22.8 & 7.4 & 48.1 & 63.3 & 101.9 \\
InternVL3 & 29.4 & 8.9 & 67.4 & 77.5 & 94.5 & 24.6 & 7.1 & 66.9 & 77.8 & 95.4 & 24.8 & 6.8 & 66.3 & 77.5 & 98.1 \\
InstructBLIP & 45.6 & 12.8 & 71.1 & 78.3 & 89.6 & 41.8 & 11.0 & 71.3 & 79.1 & 94.0 & 42.2 & 11.5 & 71.8 & 79.1 & 93.6 \\
\midrule
\textbf{Average} & 30.4 & 9.8 & \underline{61.8} & 72.8 & 91.7 & \textbf{26.6} & \textbf{7.7} & 61.4 & \underline{73.2} & 93.1 & \underline{28.5} & \underline{8.5} & \textbf{62.2} & \textbf{73.5} & 92.8 \\
\bottomrule
\end{tabular}
\end{table*}

\begin{table*}[!t]
\centering
\caption{Per-backbone 512-token decoding latency (ms/tok$\downarrow$) and eight-backbone relative slowdown ($\text{Time}\times\downarrow$) across all ten decoding methods under unified $\lambda{=}0.4$ on NVIDIA A100 80GB GPUs. $^{\ast}$: five-beam search.}
\label{tab:seff}
\small
\begin{tabular}{l cccccccccc}
\toprule
Model & Greedy & VCD & M3ID & OPERA$^{\ast}$ & ICD & SID & AGLA & ReVisiT & \textbf{IKOD} & \textbf{IKOD-A} \\
\midrule
LLaVA-1.5 & 21.6 & 42.4 & 39.9 & 321.4 & 43.5 & 43.1 & 44.8 & 22.1 & 25.0 & 25.0 \\
LLaVA-NeXT & 22.3 & 44.1 & 42.9 & 331.0 & 45.4 & 46.2 & 47.1 & 24.3 & 25.4 & 25.7 \\
LLaVA-OV & 23.5 & 46.7 & 45.1 & 274.4 & 46.1 & 48.6 & 47.0 & 26.2 & 26.6 & 27.0 \\
Qwen2-VL & 24.1 & 46.8 & 47.1 & 161.1 & 47.1 & 49.8 & 49.2 & 24.8 & 26.5 & 26.5 \\
Qwen2.5-VL & 23.8 & 45.7 & 46.9 & 246.4 & 45.9 & 49.0 & 47.5 & 24.0 & 26.2 & 26.4 \\
Qwen3-VL & 31.0 & 60.8 & 61.7 & 218.1 & 60.9 & 63.5 & 62.2 & 31.3 & 35.6 & 35.6 \\
InternVL3 & 25.2 & 50.2 & 46.9 & 227.4 & 50.0 & 50.7 & 53.2 & 27.9 & 27.8 & 28.0 \\
InstructBLIP & 22.8 & 45.2 & 43.9 & 248.9 & 45.2 & 45.7 & 46.5 & 23.5 & 27.9 & 28.1 \\
\midrule
\textbf{Average (ms/tok)} & 24.3 & 47.7 & 46.8 & 253.6 & 48.0 & 49.6 & 49.7 & 25.5 & 27.6 & 27.8 \\
\textbf{Slowdown (Time$\times$)} & 1.00$\times$ & 1.97$\times$ & 1.92$\times$ & 10.67$\times$ & 1.98$\times$ & 2.04$\times$ & 2.05$\times$ & 1.05$\times$ & 1.14$\times$ & 1.15$\times$ \\
\bottomrule
\end{tabular}
\end{table*}

\section{Per-Backbone Sampling-Strategy and Random-Seed Results}
Tables~\ref{tab:spope5strat} and~\ref{tab:seedsweep} break down the stochastic decoding results in Tables~VIII and~IX of the main paper by individual backbone. In Table~\ref{tab:spope5strat}, both variants consistently outperform Base across all five sampling regimes, with the largest margins emerging at high temperature ($T{=}1.5$) where LLaVA-1.5, LLaVA-NeXT, and InstructBLIP otherwise suffer severe prior drift. In Table~\ref{tab:seedsweep}, across all 24 backbone--seed combinations under Top-$P$ sampling, tuning-free IKOD-A lowers $C_I$ in every single run and IKOD improves 22 of 24 runs, with per-backbone standard deviations remaining within $0.1$--$1.9$ percentage points and confirming that hallucination suppression is insensitive to random seed variation.

\section{Per-Backbone Prompt-Phrasing and Length-Controlled Results}
\label{sec:supp_prompt_len}
Table~\ref{tab:spromptlen} reports the per-backbone CHAIR-500 metrics across all eight LVLMs under the three alternative prompt phrasings evaluated in Table~IX of the main paper, where both IKOD and IKOD-A consistently reduce $C_S$ and $C_I$ across concise imperatives, detailed requests, and short interrogatives while preserving or improving object F1 across individual backbones. Table~\ref{tab:lengthcontrol} provides the corresponding per-backbone breakdown under the 128-token generation budget summarized in Table~IX of the main paper: within 128 tokens where every arm finishes within 4.4 words of Greedy, IKOD and IKOD-A lower $C_I$ across all eight backbones by $1.0$--$3.1$ and $0.6$--$2.1$ points while raising object F1 on seven of eight models, confirming that hallucination reduction stems from restored visual grounding rather than briefer outputs.

\section{Complexity Accounting and Strictly Lossless Acceleration}
\label{sec:supp_efficiency}
Following Section~IV of the main paper, let $U$ denote the visual prefix length (including system tokens prior to the last visual token), $P=|\mathcal{P}_t|$ the mergeable post-visual text length at decoding step $t$, and $n_t=\operatorname{round}(\lambda P)$ ($1\le n_t\le P$) the number of merged text buckets. For an $I$-layer Transformer decoder with hidden dimension $d$ and per-layer projection/FFN cost $C_{\mathrm{proj}}(d)$, the single-step arithmetic of the base branch (attending over $U+P+1$ entries) and the auxiliary merged branch (attending over $U+n_t+1$ entries) is
\begin{equation}
\begin{aligned}
F_t^{\mathrm{base}} &= I\big[C_{\mathrm{proj}}(d)+4(U+P+1)d\big],\\
F_t^{\mathrm{merged}} &= I\big[C_{\mathrm{proj}}(d)+4(U+n_t+1)d\big].
\end{aligned}
\label{eq:supp_flops}
\end{equation}
Below we formalize the two strictly lossless acceleration mechanisms introduced in Section~IV-D and Algorithm~1 of the main paper, proving that they preserve the exact token-by-token decoding trajectory across both deterministic and stochastic decoding while substantially reducing runtime overhead.

\subsection{Plausibility-Singleton Short-Circuit and Trajectory Invariance}
At decoding step $t$, the base forward pass produces logits $\ell_t^{\mathrm{base}} \in \mathbb{R}^{|\mathcal{V}|}$ and per-layer visual attention sums $\{s_t^i\}_{i=1}^I$ from the full KV cache $\mathcal{C}_t$. If the greedy base prediction is already an end-of-sequence token ($\arg\max_{w \in \mathcal{V}} \ell_t^{\mathrm{base}}(w) \in \mathcal{E}$), generation terminates immediately without constructing the merged cache. Otherwise, let $y_t^{(1)}$ and $y_t^{(2)}$ denote the highest- and second-highest-scoring non-EOS tokens under the base branch:
\begin{equation}
\begin{aligned}
y_t^{(1)} &= \arg\max_{w \in \mathcal{V} \setminus \mathcal{E}} \ell_t^{\mathrm{base}}(w), \\
y_t^{(2)} &= \arg\max_{w \in \mathcal{V} \setminus (\mathcal{E} \cup \{y_t^{(1)}\})} \ell_t^{\mathrm{base}}(w).
\end{aligned}
\end{equation}
By the definition of the relative plausibility constraint in Eq.~(7) of the main paper, $\mathcal{V}_t = \{y \in \mathcal{V} : \ell_t^{\mathrm{base}}(y) \ge \max_{w \in \mathcal{V}} \ell_t^{\mathrm{base}}(w) + \log\beta\}$ with plausibility threshold $\beta \in (0,1]$ (producing a log-margin cutoff of $-\log\beta \approx 2.3026$ at our default $\beta{=}0.1$). Since the greedy token is non-EOS ($\arg\max_{w \in \mathcal{V}} \ell_t^{\mathrm{base}}(w) \notin \mathcal{E}$), the global maximum over $\mathcal{V}$ is attained at $y_t^{(1)}$, i.e., $\max_{w \in \mathcal{V}} \ell_t^{\mathrm{base}}(w) = \ell_t^{\mathrm{base}}(y_t^{(1)})$, and every remaining non-EOS token $w \in \mathcal{V} \setminus (\mathcal{E} \cup \{y_t^{(1)}\})$ satisfies $\ell_t^{\mathrm{base}}(w) \le \ell_t^{\mathrm{base}}(y_t^{(2)})$. Consequently, the non-EOS plausible candidate set $\mathcal{V}_t \setminus \mathcal{E}$ reduces to the singleton $\{y_t^{(1)}\}$ if and only if the top-2 non-EOS base logit margin strictly exceeds $-\log\beta$:
\begin{equation}
\begin{aligned}
\Delta\ell_t^{\mathrm{base}} &\triangleq \ell_t^{\mathrm{base}}\big(y_t^{(1)}\big) - \ell_t^{\mathrm{base}}\big(y_t^{(2)}\big) > -\log\beta \\
&\iff \mathcal{V}_t \setminus \mathcal{E} = \big\{y_t^{(1)}\big\}.
\end{aligned}
\label{eq:supp_margin}
\end{equation}
Whenever Eq.~\eqref{eq:supp_margin} holds (or $\alpha_t = 0$), optimizing the collaborative objective in Eqs.~(6) and~(8) of the main paper over $y \in \mathcal{V}_t \setminus \mathcal{E}$ is taken over a single element and therefore evaluates identically to $y_t^{(1)}$ for any finite auxiliary logit vector $\ell_t^{\mathrm{merged}} \in \mathbb{R}^{|\mathcal{V}|}$:
\begin{equation}
\begin{aligned}
y_t &= \arg\max_{y \in \{y_t^{(1)}\}}\Big[\ell_t^{\mathrm{base}}(y) + \alpha_t\,\ell_t^{\mathrm{merged}}(y)\Big] \\
&\equiv y_t^{(1)}, \qquad \forall\,\ell_t^{\mathrm{merged}}\big(y_t^{(1)}\big) \in \mathbb{R}.
\end{aligned}
\label{eq:supp_singleton_id}
\end{equation}
Furthermore, skipping KV merging and the auxiliary forward pass on step $t$ preserves the exact state trajectory for all subsequent steps $t'\! > t$: (i) the full KV cache $\mathcal{C}_{t+1}$ is updated solely by appending the base-branch key--value pairs of $y_t = y_t^{(1)}$ (the merged cache $\tilde{\mathcal{C}}_t$ is transient and discarded after step $t$); (ii) the accumulated visual attention table $\{s_m^i\}_{m \in \mathcal{P}_t}$ depends only on base-branch attention matrices; and (iii) in IKOD-A, $S_t = \frac{1}{I}\sum_{i=1}^I s_t^i$, the relative decay $D_t = \max(0, (\bar{S}_0 - S_t)/\bar{S}_0)$, and the running median $\tilde{D}_t = \operatorname{median}_{k\le t}(D_k)$ in Eq.~(10) are computed from the base pass \emph{prior} to the short-circuit check, ensuring bit-for-bit identical $\tilde{D}_{t'}$ and $\alpha_{t'}$ across all future steps.

\begin{figure*}[!t]
    \centering
    \includegraphics[width=\textwidth]{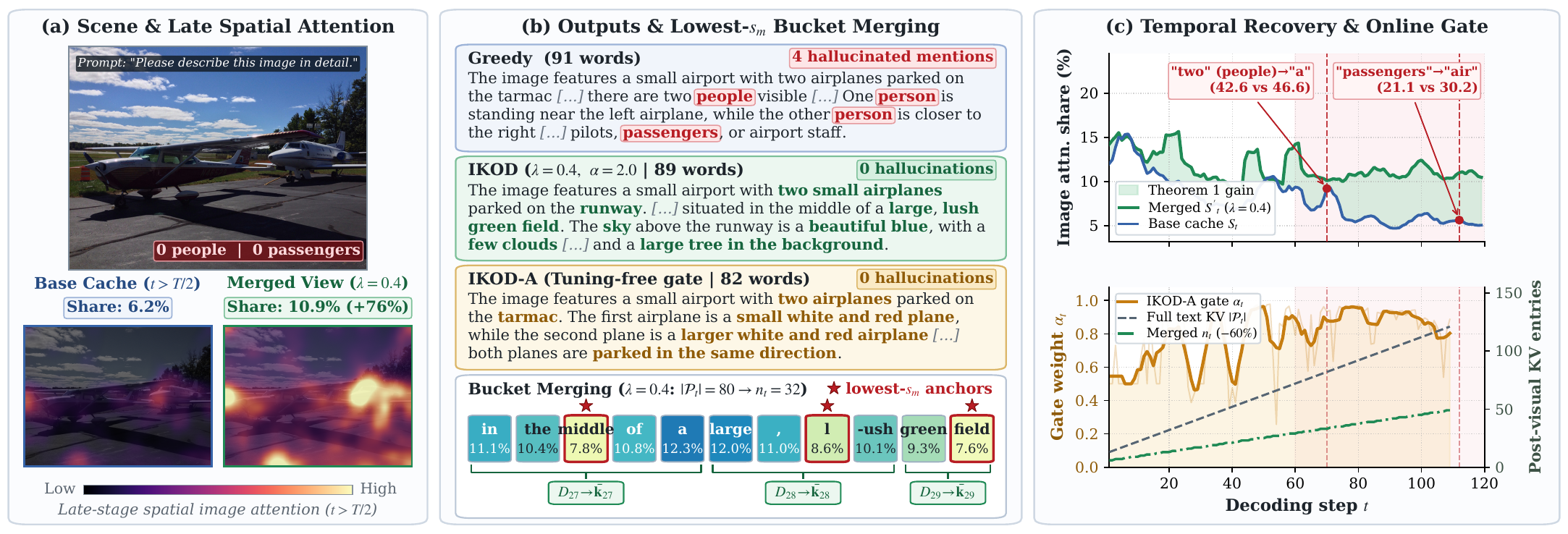}
    \caption{Additional mechanistic case study on LLaVA-1.5 (COCO val2014, twin parked airplanes under unified $\lambda{=}0.4$). (a) Input airport scene containing zero people or passengers and late-stage ($t > T/2$) spatial attention overlays. (b) Generated captions from Greedy, IKOD, and IKOD-A alongside lowest-$s_m$ bucket merging ($D_b \to \tilde{\boldsymbol{k}}_b$). (c) Per-step visual attention recovery ($S'_t > S_t$) with exact logit-reversal steps and IKOD-A's online gate trajectory $\alpha_t$.}
    \label{fig:smech268944}
\end{figure*}

\begin{figure*}[!t]
    \centering
    \includegraphics[width=\textwidth]{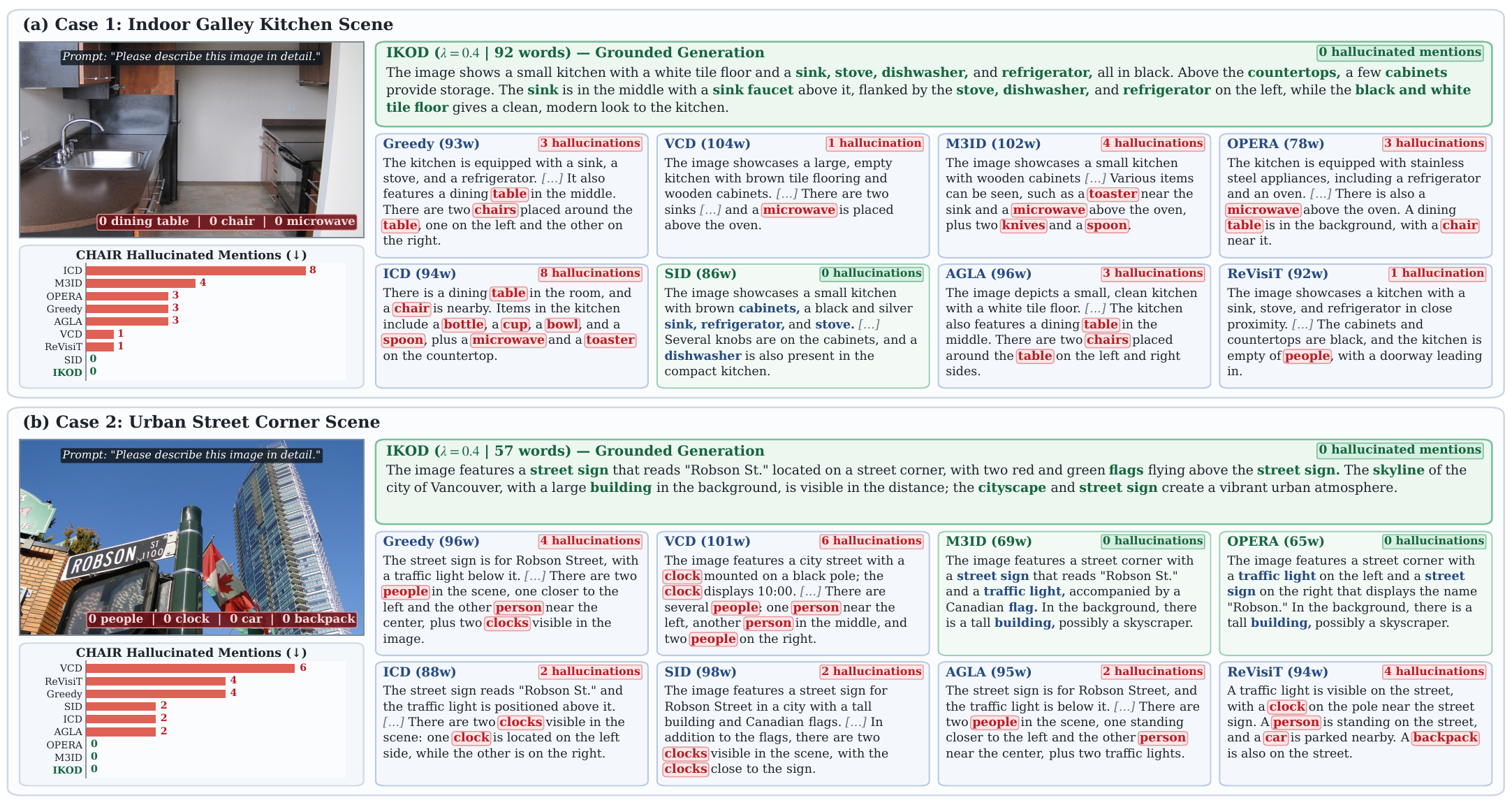}
    \caption{Qualitative comparison on LLaVA-1.5 against Greedy and the seven baselines of Table~II. Hallucinated objects identified by CHAIR are highlighted in red.}
    \label{fig:scasestudy}
\end{figure*}

\subsection{Persistent Visual-Prefix Buffer and Cross-Layer Vectorized Bucket Assignment}
On multi-candidate steps where $|\mathcal{V}_t \setminus \mathcal{E}| \ge 2$ and $\alpha_t > 0$, the auxiliary merged cache at layer $i \in \{1,\dots,I\}$ has the block structure
\begin{equation}
\begin{aligned}
\tilde{\boldsymbol{K}}_t^i &= \Big[\underbrace{\boldsymbol{K}_{[1:U]}^i}_{\text{visual prefix}} \;\Big\|\; \underbrace{\bar{\boldsymbol{K}}_{\mathcal{B}_t}^i}_{\text{$n_t$ centroids}}\Big] \in \mathbb{R}^{H_{\mathrm{kv}} \times (U+n_t) \times d_h}, \\
\tilde{\boldsymbol{V}}_t^i &= \Big[\boldsymbol{V}_{[1:U]}^i \;\Big\|\; \bar{\boldsymbol{V}}_{\mathcal{B}_t}^i\Big] \in \mathbb{R}^{H_{\mathrm{kv}} \times (U+n_t) \times d_h},
\end{aligned}
\label{eq:supp_block_kv}
\end{equation}
where $H_{\mathrm{kv}}$ is the number of key--value heads and $d_h$ is the head dimension. Because $\boldsymbol{K}_{[1:U]}^i$ and $\boldsymbol{V}_{[1:U]}^i$ remain fixed throughout generation, we allocate a persistent auxiliary buffer $\mathcal{B}_{\mathrm{aux}}^i \in \mathbb{R}^{H_{\mathrm{kv}} \times (U+P_{\max}) \times d_h}$ once after prompt prefill and copy the $U$ visual prefix entries a single time. At step $t$, rather than iterating over layers and buckets in Python, we stack the post-visual attention scores $\boldsymbol{S}_{\mathcal{P}_t} \in \mathbb{R}^{I \times P}$ across all $I$ layers simultaneously, extract the sorted lowest-attention anchor offsets $\boldsymbol{A}_t = \operatorname{sort}\!\big(\operatorname{argsort}(\boldsymbol{S}_{\mathcal{P}_t})[:, :n_t]\big) \in \{0,\dots,P-1\}^{I \times n_t}$, and compute the contiguous midpoint boundaries $\boldsymbol{M}_t = \lfloor(\boldsymbol{A}_{t,:,:-1} + \boldsymbol{A}_{t,:,1:})/2\rfloor \in \{0,\dots,P-1\}^{I \times (n_t-1)}$. Batched binary search ($\operatorname{searchsorted}$) assigns every post-visual position $m \in \mathcal{P}_t$ to its bucket $D_b^i$ ($b \in \{1,\dots,n_t\}$) across all layers in parallel, forming the row-normalized sparse assignment operator $\boldsymbol{W}_t^i \in \mathbb{R}^{n_t \times P}$ applied along the sequence dimension $P$ for each KV head:
\begin{equation}
\begin{aligned}
\big(\boldsymbol{W}_t^i\big)_{b,m} &= \frac{\mathbb{I}(m \in D_b^i)}{|D_b^i|}, \\
\bar{\boldsymbol{K}}_{\mathcal{B}_t}^i &= \boldsymbol{W}_t^i\,\boldsymbol{K}_{\mathcal{P}_t}^i, \qquad
\bar{\boldsymbol{V}}_{\mathcal{B}_t}^i = \boldsymbol{W}_t^i\,\boldsymbol{V}_{\mathcal{P}_t}^i,
\end{aligned}
\label{eq:supp_vec_merge}
\end{equation}
which overwrites only slice $[:, U:U+n_t, :]$ of $\mathcal{B}_{\mathrm{aux}}^i$ in-place and exposes the view $[:, 0:U+n_t, :]$ to the auxiliary forward pass. This reduces per-step KV memory movement from $O\big(I(U+n_t)d\big)$ to $O(In_t d)$ (eliminating $>90\%$ of copy traffic since $U \gg n_t$, e.g., $U{=}576$ vs.\ $n_t \le 40$). Denoting the empirical short-circuit rate over a $T$-step generation by $\rho_{\mathrm{sc}} = \frac{1}{T}\sum_{t=1}^T \mathbb{I}\big(|\mathcal{V}_t \setminus \mathcal{E}|{=}1 \lor \alpha_t{=}0\big)$ and the vectorized bucket-aggregation cost by $C_{\mathrm{vec}}(I,P,n_t,d) = O(IPd + IP\log P)$, the expected per-step complexity of IKOD is
\begin{equation}
\bar{F}_t^{\mathrm{IKOD}} = F_t^{\mathrm{base}} + \big(1 - \rho_{\mathrm{sc}}\big)\Big(F_t^{\mathrm{merged}} + C_{\mathrm{vec}}(I,P,n_t,d)\Big).
\label{eq:supp_expected_cost}
\end{equation}
On 512-token captioning, where $\rho_{\mathrm{sc}} \approx 38\%$ of steps are short-circuited and the remaining steps reuse the persistent visual prefix, Table~\ref{tab:seff} reports the per-backbone decoding latency across all eight LVLMs and all ten methods summarized in Table~X of the main paper. Across the eight backbones, IKOD and IKOD-A average $27.6$ and $27.8$\,ms/tok ($1.14\times$ and $1.15\times$ Greedy), whereas two-pass contrastive baselines require $46.8$--$49.7$\,ms/tok and five-beam OPERA reaches $253.6$\,ms/tok.

\section{Additional Qualitative Examples}
Figure~\ref{fig:smech268944} presents a second mechanistic case study on LLaVA-1.5 ($\lambda{=}0.4$) on an airport scene with two parked airplanes and zero people or passengers. Under Greedy decoding, late-stage ($t > T/2$) image attention drops to $6.2\%$ and induces four hallucinations (\textit{people}, \textit{person}, \textit{passengers}), whereas merging contiguous post-visual intervals around the lowest-$s_m$ anchors into $n_t{=}32$ centroids ($D_b \to \tilde{\boldsymbol{k}}_b$, 60\% KV reduction) restores late attention to $10.9\%$ ($+76\%$), reverses hallucinatory preferences at the divergence steps (\texttt{"two"\,(people)$\to$"a"}, \texttt{"passengers"$\to$"air"}), and drives IKOD-A's gate $\alpha_t$ above $0.9$ as $S_t$ decays. Figure~\ref{fig:scasestudy} compares full captions on two scenes against Greedy and all seven baselines from Table~II: baselines hallucinate up to 8 absent objects in the indoor kitchen (\textit{dining table}, \textit{chair}, \textit{microwave}, \textit{toaster}, \textit{bottle}, \textit{cup}, \textit{bowl}, \textit{spoon}) and up to 6 in the urban street corner (\textit{clock}, \textit{people}, \textit{car}, \textit{backpack}), whereas IKOD accurately grounds every visible appliance, countertop, street sign, flag, and building across both scenes with zero hallucinated mentions. Notably, whereas contrastive baselines still drift into co-occurring kitchenware or street clutter during the second half of the caption, IKOD sustains fine-grained spatial and relational descriptions throughout the entire response.

\end{document}